\documentclass[letterpaper]{article} 
\usepackage{aaai23}  
\usepackage{times}  
\usepackage{helvet}  
\usepackage{courier}  
\usepackage[hyphens]{url}  
\usepackage{graphicx} 
\usepackage{natbib}  
\usepackage{caption} 
\usepackage{algorithm}
\usepackage{algorithmic}
\usepackage{xcolor}

\usepackage{newfloat}
\usepackage{listings}
\DeclareCaptionStyle{ruled}{labelfont=normalfont,labelsep=colon,strut=off} 
\floatstyle{ruled}
\newfloat{listing}{tb}{lst}{}
\floatname{listing}{Listing}
\usepackage{wrapfig}

\usepackage{amsfonts}
\usepackage{caption}
\usepackage{subcaption}
\usepackage{enumitem}
 
\newcommand{\note}[1]{\textcolor{blue!50}{note:~~#1}} 

\usepackage{xspace} 
\newcommand{\ie}{\emph{i.e.,}\xspace}
\newcommand{\eg}{\emph{e.g.,}\xspace}

\newcommand{\aka}{\emph{a.k.a.,}\xspace}
\newcommand{\etal}{\emph{et~al.}\xspace} 
\usepackage{amsmath}
\usepackage{amssymb}
\usepackage{mathtools}
\usepackage{amsthm}

\usepackage[capitalize,noabbrev]{cleveref}

\theoremstyle{plain}
\newtheorem{theorem}{Theorem}

\theoremstyle{definition}
\newtheorem{definition}[theorem]{Definition}

\theoremstyle{remark}

\title{Evidential Conditional Neural Processes}

\author {
    Deep Shankar Pandey, $\;$
    Qi Yu 
}
\affiliations {
    Rochester Institute of Technology\\
   \{dp7972, qi.yu\}@rit.edu
}

\usepackage{bibentry}

\begin{document}

\maketitle

\begin{abstract}
The Conditional Neural Process (CNP) family of models offer a promising direction to tackle few-shot problems by achieving better scalability and competitive predictive performance. However, the current CNP models only capture the overall uncertainty for the prediction made on a target data point. They lack a systematic fine-grained quantification on the distinct sources of uncertainty that are essential for model training and decision-making under the few-shot setting. We propose Evidential Conditional Neural Processes (ECNP), which replace the standard Gaussian distribution used by CNP with a much richer hierarchical Bayesian structure through evidential learning to achieve epistemic-aleatoric uncertainty decomposition. The evidential hierarchical structure also leads to a theoretically justified robustness over noisy training tasks. Theoretical analysis on the proposed ECNP establishes the relationship with CNP while offering deeper insights on the roles of the evidential parameters. Extensive experiments conducted on both synthetic and real-world data demonstrate the effectiveness of our proposed model in various few-shot settings.
\end{abstract}

\section{Introduction}
\label{introduction}
Meta-learning \cite{finn2017model} offers a powerful vehicle to tackle the challenges of learning from limited data. 
It formulates learning into two phases: meta-training that learns the global (meta) knowledge shared across tasks and meta-testing that adapts the global knowledge to the limited data from few-shot testing tasks. While meta-learning achieves improved generalization capability by leveraging the meta-knowledge obtained from the meta-training tasks, few-shot tasks arising in the testing phase may deviate significantly from the training tasks. Furthermore, data in many real-world applications may be highly noisy, incomplete, or corrupted. These, when coupled with the weakly supervised signal from limited training data, make few-shot learning inherently uncertain and challenging.

Among existing meta-learning models, metric-based approaches \cite{vinyals2016matching,snell2017prototypical,chen2021meta} have achieved high predictive accuracy for few-shot classification problems. However,  most metric-based models are not designed to output uncertainty, limiting their applicability to many real-world problems. Meanwhile, gradient-based approaches, such as MAML \cite{finn2017model}, have been extended to achieve uncertainty-aware meta-learning through Bayesian modeling. MAML formulates meta-learning as a bi-level optimization problem that requires expensive Hessian-gradient products during meta-learning along with other challenges such as training stability. First order approximations and alternatives of MAML, such as Reptile~\cite{nichol2018first}, require time consuming gradient based adaptation during inference limiting their applications. Extending such models for uncertainty quantification \cite{yoon2018bayesian}
may further increase the computational costs. 

Different from deep learning (DL) models, Gaussian Processes (GPs)~\cite{williams2006gaussian} offer a principled way to quantify uncertainty. By combining Bayesian modeling and kernel methods, a GP outputs a distribution over functions, where the kernel serves as a fixed prior that determines the smoothness of the functions as a specific form of meta-knowledge. 
However, GPs, in their original form, suffer from a high computational cost for inference. Their generalization capability may also be limited due to the restricted priors induced from the fixed kernel functions, lacking the flexibility to adapt to the training data. This also significantly hinders GPs from being used as an effective meta-learning model, which needs to encode the meta-knowledge learned from other tasks in support of few-shot learning from new tasks.   

The recently developed conditional neural processes (CNPs) \cite{garnelo2018conditional}, neural processes (NPs) \cite{garnelo2018neural}, and their extensions provide a suite of effective meta-learning models, which bring together the benefits of GP's uncertainty capabilities and the DL models' flexibility of adapting to the data. Besides offering better scalability, rapid inference, and competitive predictive performance \cite{kim2018attentive,Gordon2020Convolutional}, these models also naturally quantify uncertainty by simulating a stochastic process like a GP. However, the current NP models are sensitive to the outliers in the training tasks and lack competitive performance. Alternatively, CNP family of models achieve strong performance but only capture the overall uncertainty for the prediction made on a target data point. They lack a systematic fine-grained quantification of the different sources of uncertainty. Simply, CNP based models approximate the predictive distribution on a target data point by predicting both the mean and variance of a Gaussian. However, the variance term itself does not offer deeper insight on the two distinct sources of uncertainty: (i) lack of knowledge by the model (epistemic) or (ii) noise inherent in the data (aleatoric). Identifying the source of uncertainty can offer effective means to improve the model training (\eg by collecting more training data or constructing more informative sets) and facilitate critical decision-making (\eg whether to include humans in the loop).


In this paper, we propose Evidential Conditional Neural Processes (ECNPs), which provide novel and nontrivial extensions to CNP family of models with principled uncertainty quantification and decomposition. Being an CNP, an ECNP inherits all attractive model behaviors from the CNP family, including competitive predictive performance and scalability. By integrating evidential learning, an ECNP replaces a simple Gaussian distribution of CNP models with a much richer hierarchical Bayesian structure that leads to a robust neural process model with accurate epistemic-aleatoric uncertainty decomposition capabilities without any additional computational overhead. Such decomposition allows us to separate uncertainty caused by the noise in the data and the model's lack of knowledge on the target data point when making a prediction. 
Our main contributions are:
\begin{itemize}[noitemsep,topsep=2pt,leftmargin=*]
    \item The integration of evidential learning with CNPs results in a novel family of evidential conditional neural processes that are robust to outliers in meta-training and provides fine-grained uncertainty decomposition, both of which are essential for few-shot learning.
    \item A thorough theoretical analysis on the proposed ECNPs establishes the relationship with CNPs while offering deeper insights on the roles of the evidential parameters and why ECNPs are more suitable for few-shot learning.
\end{itemize}

\section{Related Works}
We discuss existing works that are most relevant to the proposed evidential neural processes in this section. Some additional related works are covered in the Appendix~\cite{appendixdpaaai2023}. 
\paragraph{Uncertainty-aware Meta-Learning.} There have been increasing efforts ~\cite{yoon2018bayesian,finn2018probabilistic,gordon2018meta,grant2018recasting,ravi2018amortized} to develop meta-learning models that can quantify uncertainty.
Uncertainty information can be achieved through an ensemble of a diverse set of meta-learning models as in Bayesian MAML \cite{yoon2018bayesian}. Uncertainty can also be estimated by considering a hierarchical model for meta-learning and carrying out Bayesian inference. To this end, ABML \cite{ravi2018amortized} considers a hierarchical Bayesian model and uses amortized variational inference across tasks to obtain the uncertainty information. LLAMA \cite{grant2018recasting} shows MAML as inference in a hierarchical Bayesian model with empirical Bayes and uses Laplace approximation to obtain Gaussian distribution for the posterior distribution that effectively captures the uncertainty. PLATIPUS \cite{finn2018probabilistic} extends MAML using amortized variational inference to learn a distribution over prior model parameters that captures the uncertainty. These meta-learning approaches are computationally expensive and may lack rapid inference capabilities.
\paragraph{Neural Process Family.} Neural Process (NP)-based models \cite{garnelo2018conditional,garnelo2018neural,kim2018attentive,Gordon2020Convolutional} offer computationally efficient alternatives to existing uncertainty-aware meta-learning approaches as inference in NP is a computationally cheap forward pass through an encoder-decoder architecture. 
Generative Query Networks (GQN) \cite{eslami2018neural} can be seen as one of the earliest NP models that use a generation network and a query network to tackle scene representation and autonomous scene understanding problems. Conditional Neural Processes (CNP) \cite{garnelo2018conditional} generalize GQN using an encoder-aggregator-decoder architecture. 
Neural processes~\cite{garnelo2018neural} further generalize CNPs by introducing a latent variable in the encoder-decoder architecture. 
Attentive Neural processes (ANP)~\cite{kim2018attentive} replace the mean aggregation in CNP with multi-headed attention that learns to attend to the most relevant context points
leading to significantly better target embedding and improved results at the cost of increased computational cost from the attention mechanism. Convolutional Conditional Neural Processes (ConvCNPs) \cite{Gordon2020Convolutional} achieve translation equivariance using a functional space representation for the context set. 
CNAPS \cite{requeima2019fast} and Simple CNAPS \cite{bateni2020improved} extend the NP models to handle few-shot classification tasks. Various evaluation metrics such as Inclusion@K and Uncertainty Increase were introduced in \cite{groverprobing} to better analyze the uncertainty capabilities of neural process models. Le \etal  \cite{le2018empirical} and Naderiparizi  \etal \cite{naderiparizi2020uncertainty} studied the impact of architecture choices and different optimization objective choices for NP and CNP models. GNP \cite{bruinsma2021gaussian} and  FullConvGN \cite{DBLP:journals/corr/abs-2108-09676} extended the CNP models to handle predictive correlations, i.e., the dependencies in output.

As discussed earlier, CNPs and their variants can only capture the overall predictive uncertainty on the target points. NPs recover the posterior distribution of the model after being exposed to the context points by introducing a global latent variable. However, NPs require approximation procedures and usually resort to computationally expensive sampling schemes for model training/inference.
The proposed ECNPs address these critical gaps by integrating evidential learning with CNP models through an evidential hierarchical Bayesian prior with a much richer representation power to support fine-grained uncertainty decomposition while achieving robust predictions and being computationally efficient in few-shot settings.

\vspace{-1mm}\section{Evidential Neural Processes}
\paragraph{Problem Setup:} Consider a meta-dataset $\mathcal{M} = \{\mathcal{D}^i\}_{i=1}^{M}$, which consists of a collection of datasets/tasks. Each task $\mathcal{D} = (\mathcal{C}, \mathcal{T}) = \{ ( x_{n}, y_{n} ) \}_{n=1}^{N_c + N_t}$ consists of a context set (\aka support set) $\mathcal{C} = \{X_c, Y_c\} = \{ ( x_{n}, y_{n} ) \}_{n=1}^{N_c}$, a collection of $N_c$ input-output pairs, and the target set (\aka the query set) $\mathcal{T} = \{X_t, Y_t\} = \{ ( x_{t}, y_{t} ) \}_{t=1}^{N_t}$ a collection of $N_t$ input-output pairs. Meta-learning occurs in two phases: 1) meta-training where both the context set and target set information is available to the model, and 2) meta-testing where the model is provided with context set information and evaluated based on the performance over target set inputs. 

\begin{figure}[t!]
        \includegraphics[width=0.48\textwidth]{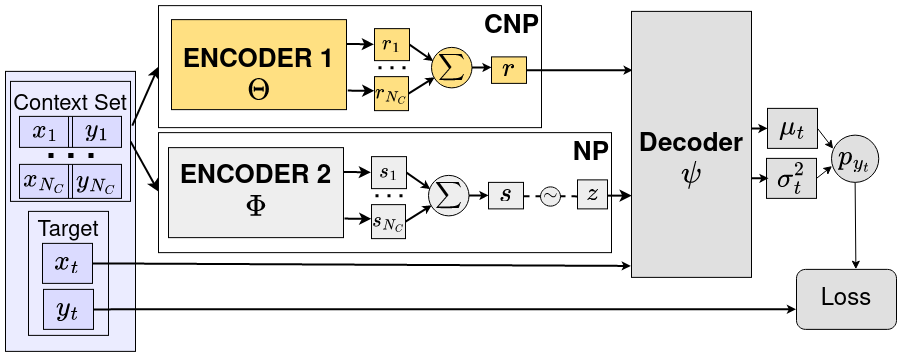}
        \centering
        \caption{CNP and NP Models
        }
        \label{fig:npcnpModel} 
\vspace{-1mm}
\end{figure}

\begin{figure}[h]
        \includegraphics[width=0.48\textwidth]{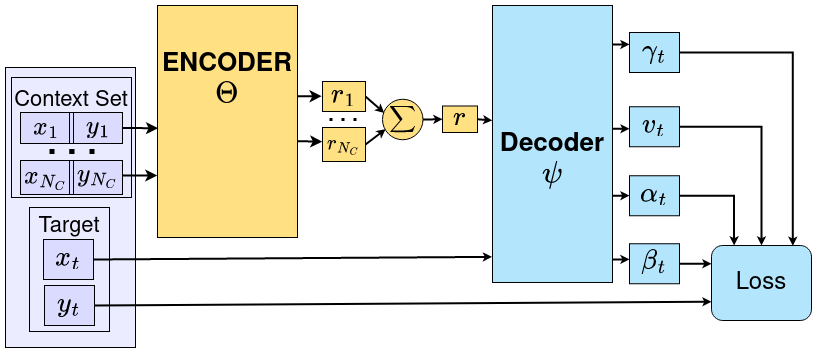}
        \centering
        \caption{ECNP Model
        }
        \label{fig:ecnpModel} 
\vspace{-1mm}
\end{figure}

\vspace{-1mm}\subsection{Uncertainty Analysis via CNP and NP}\vspace{-1mm}
A CNP, as shown by the top branch of Figure \ref{fig:npcnpModel} , has a deterministic mapping from a context set $\mathcal{C}$ and a target input $x_t$ to its prediction. 
Specifically, each context point is embedded by encoder $\Theta$ to representations $r_1,...,r_{N_C}$, aggregated to a representation $r$, and passed through the decoder to obtain the parameters of the predictive distribution. 
The predictive distribution is assumed to be a Gaussian, where the decoder outputs the mean $\mu_t$ and the variance $\sigma_t^2$: $P_\theta(y_t | x_t ; \mathcal{C})= \mathcal{N}(y_t | \mu_t, \sigma_t^2)$. The variance term $\sigma_t^2$ captures the overall uncertainty. A NP model, as shown by the bottom branch of Figure \ref{fig:npcnpModel}, introduces a latent variable $z$ to produce a distribution over functions given the same context set $\mathcal{C}$. 
Specifically, each context point is embedded by encoder $\Phi$ to representations $s_1,...,s_{N_C}$, aggregated to a representation $s$, and passed through a NN to obtain the parameters for the latent distribution. 
The latent variable induced distribution over functions allows NPs to model both epistemic and aleatoric uncertainty for each target point prediction. Assume that the latent variable follows a distribution $q(z)$ and by sampling from this distribution, we obtain $z_{1,..,L} \sim q(z)$. For each sampled $z_l$ and a target point $x_t$, the decoder outputs a predictive distribution $p(y_t|x_t, z_l)=\mathcal{N}(\mu_l,\sigma_l^2)$. As a result, we can obtain the epistemic uncertainty as the variance of the mean outputs $\text{Var}[\mu_l]$ and the aleatoric uncertainty as the expected variance $\mathbb{E}[\sigma_l^2]$. 

However, there are two key limitations for NP-based uncertainty decomposition. First, it requires sampling of the latent variable, which may make the overall inference computationally expensive, especially when learning from a large number of tasks. Second, the fine-grained uncertainties are obtained indirectly (\eg via MC sampling) and thus it becomes challenging to guide the model to correct its inherent mistakes regarding fine-grained uncertainties during training. Moreover, both CNP and NP lack robustness to outliers in the training tasks.
The proposed ECNP addresses the key limitations of both CNP and NPs. We present an evidential extension of CNP that outputs the aleatoric and epistemic uncertainty directly from the deterministic path while ensuring a robust prediction given a noisy context set. The introduced hierarchical structure explicitly captures fine-grained uncertainty enabling the model to correct its mistakes in the fine-grained uncertainties.

\subsection{Evidential Conditional Neural Process (ECNP)}\label{sec:ECNP}
We extend the CNP model to an evidential neural process. To this end, as in \cite{NEURIPS2020_aab08546}, we assume that the likelihood function is a Gaussian with an unknown mean and variance. We place an evidential prior over the mean and variance and train the neural process to output the hyperparameters of the evidential distribution using the limited information of the context set and target input. Moreover, we train the evidential model to be confident for the correct prediction and output low evidence (\ie confidence) when the model's predictions are incorrect. Our evidential conditional model introduces insignificant computational overhead and is deterministic while being expressive in uncertainty quantification. In particular, it can quantify both aleatoric and epistemic uncertainty with a single forward pass through the network without any sampling as in the NP. 

\vspace{-1mm}\paragraph{Uncertainty and evidence quantification by ECNP.} In the ECNP model, we consider a hierarchical Bayesian structure in which each target observation $y_t$ is a sample from a Gaussian $\mathcal{N}(y_t|\mu,\sigma)$, whose mean and variance are governed by a higher-order Normal-Inverse-Gamma prior \cite{bishop2006pattern}:
\begin{align}
\label{eq:nig}    \text{NIG}(\mu, \sigma^2|\mathbf{p}_t) = \mathcal{N}(\mu|\gamma_t,\frac{\sigma^2}{v_t})
    \Gamma^{-1}(\sigma^2|\alpha_t, \beta_t)
\end{align}
where $\mathbf{p}_t= (\gamma_t, v_t, \alpha_t, \beta_t)$ and $\Gamma^{-1}$ is an inverse-gamma distribution. 
Intuitively, the context set $\mathcal{C}$ interacts with the meta knowledge in the meta-learning model to output the prior NIG parameters $\mathbf{p}_t= (\gamma_t, v_t, \alpha_t, \beta_t)$. When being exposed a new target point $x_t$, this prior will interact with the Gaussian likelihood 
$p(y_t| x_t) = \mathcal{N}(\mu, \sigma^2)$ to produce a Student-t predictive distribution given by 
\begin{align}
 \nonumber   &p(y_t|x_t, \mathbf{p}_t) = \int_\mu \int_{\sigma^2} p(y_t| x_t, \mu, \sigma^2) \text{NIG}(\mu, \sigma^2 |\mathbf{p}_{t}) \text{d}\mu \text{d}\sigma^2 \\
     =& \frac{\Gamma(\alpha_t + \frac{1}{2})}{\Gamma(\alpha_t)} \sqrt{\frac{v_t}{2\pi\beta_t(1+v_t)}}  \Big(1+\frac{v_t (y_t - \gamma_t)^2}{ 2\beta_t (1 +v_t)}\Big)^{- ( \alpha_t + \frac{1}{2})} \nonumber \\
    =& \text{St} \Big(y_t; \gamma_t, \frac{\beta_t(1+v_t)}{v_t\alpha_t}, 2\alpha_t\Big)
\end{align}
As a result, the prediction for a target point $x_t$ is 
\begin{align}
    \hat{y}_t &= \mathbb{E}_{p(y_t|x_t, \mathbf{p}_t)}{[y_t]} = \int y_t p(y_t|x_t, \mathbf{p}_t) \text{d} y_t = \gamma_t 
\end{align}
Given the predicted evidential parameters, the NIG distribution is fully characterized, which allows us to evaluate $\text{Var}[\mu]$ and $\mathbb{E}[\sigma^2]$ that can be used to quantify the aleatoric (AL) and epistemic (EP) uncertainty,  respectively: 
\begin{align}
\label{eq:AL} \texttt{AL}  = \mathbb{E}[\sigma^2] = \frac{\beta_t}{\alpha_t - 1}, \quad
\texttt{EP} = \text{Var}[\mu] = \frac{\beta_t}{v_t(\alpha_t - 1)}
\end{align}
By leveraging the conjugacy between the NIG prior and the Gaussian likelihood, it can be shown that after interacting with $N$ i.i.d. data samples, the posterior is still a $\text{NIG}(\mu, \sigma^2|\mathbf{p}_N)$, where $\mathbf{p}_N=(\gamma_N, v_N, \alpha_N, \beta_N)$ with 
\begin{align}
    v_N=v+N, \quad \alpha_N=\alpha+ \frac{N}{2}
\end{align}
Thus, both $v$ and $\alpha$ can be naturally interpreted as the evidence (in the form of pseudo counts) to quantify the confidence on the prior mean and the prediction of a target data sample, respectively. Furthermore, $\beta$ denotes the initial variance of the model and \eqref{eq:AL} shows that a large  $\beta$ leads to a low confidence in the model's prediction, which implies lack of evidence. By aggregating all evidence related parameters, ECNP is able to quantify the overall model confidence as
\begin{align}
    \mathcal{E}_t = v_t + \alpha_t + \frac{1}{\beta_t}   
\end{align}
A more detailed posterior analysis on the hierarchical model for evidence quantification is provided in the Appendix.
 
\vspace{-1mm}\paragraph{Training ECNP.} In this evidential framework, learning is formulated as an evidence acquisition process and the model is trained to maximize the likelihood of model evidence. Equivalently, we train the model to minimize the negative log-likelihood of the model given by  
\begin{multline}
    L_t^\text{NLL} =\log\frac{\Gamma(\alpha_t)\sqrt{\frac{\pi}{v_t}}}{\Gamma(\alpha_t + \frac{1}{2})} - \alpha_t \log(2\beta_t(1+v_t)) +\\
    \big(\alpha_t + \frac{1}{2}\big) \log\left((y_t - \gamma_t)^2v_t + 2\beta_t(1+v_t)\right) 
\end{multline}

We further introduce an evidence regularization term to encourage the model to output low evidence/confidence when the predictions are incorrect:
\begin{align}\label{eq:evidence_regularization}
    \mathcal{L}_t^\text{R} = |y_t - \gamma_t |\times \mathcal{E}_t
\end{align}
The regularization term $L_t^\text{R}$ penalizes the evidence of highly confident wrong predictions. In other words, the model is trained to output a low value of $v_t$ and $\alpha_t$ and high $\beta_t$ values when the prediction is wrong leading to high uncertainty in the predictions. Finally, the model is expected to output high epistemic uncertainty at regions far from the observed context points as only the meta-knowledge is available for predictions at those points. To this end, a novel kernel-based regularization term is introduced as
    \begin{align}\label{kernel_regularization}
    \mathcal{L}_t^\text{KER} = v_t \times D(x_t, \mathcal{C})
\end{align}
where $D(x_t, \mathcal{C})$ is a distance function that measures the minimum Euclidean distance between the target point input $x_t$ and the context set $\mathcal{C}$. When the target input is far away from the context set, this kernel loss dominates the overall loss leading to small $v_t$ values and equivalently high epistemic uncertainty (EP).

The overall loss in the evidential model is the regularized sum of the model evidence loss, evidence regularization loss, and the kernel regularization loss: 
\begin{equation}
\label{eq:cnpLoss}
    \mathcal{L} =  \sum_{t=1}^{N_t} \mathcal{L}_t^\text{NLL} + \lambda_1 \mathcal{L}_t^\text{R} + \lambda_2 L_t^\text{KER} 
\end{equation}
where $\lambda_1$ and $\lambda_2$ are regularization terms.

\vspace{-1mm}\subsection{Theoretical Analysis}\vspace{-1mm}
\label{sec:toNP}

In this section, we present our theoretical results that show the superiority of the ECNP and reveal the deeper connection between the proposed ECNPs and the NP models. These theoretical results help to justify why ECNPs provide a more principled way to conduct meta-learning over few-shot tasks than the CNP family of models.

\begin{theorem}
\label{OutlierTheorem}
The ECNP model with a hierarchical Bayesian structure in the decoder is  guaranteed to be more robust to outliers in the training tasks as compared to the CNP models that use a Gaussian structure.
\end{theorem}
The detailed proof is provided in the Appendix. Intuitively, when the model evidence is finite (\ie $\alpha_t < \infty$), the outliers will be assigned a lower weight than normal data samples when evaluating the gradient for model update. When $\alpha_t \rightarrow \infty$, the model will behave similarly to the CNP model and be less robust to outliers. In our model, the hierarchical bayesian structure leads to the heavy tailed t predictive distribution enabling outlier robustness. Similar outlier robustness can, in theory, be introduced in the CNP models by modifying the CNP decoder to directly parameterize the heavy distributions (e.g. Student t distribution) and training to minimize the log likelihood under the new distribution. Empirical studies of such robust distributions for CNP models can be an interesting future work. However, such modeling would lack efficient and fine-grained uncertainty quantification capabilities, a major focus of our work.  

\begin{theorem}
\label{mainTheorem}
The conditional neural process is one instance of an evidential neural process when two of the evidential hyperparameters meet the following conditions: (i) $\alpha_t \rightarrow \infty$; (ii) $\alpha_t v_t = \text{const}$. 
\end{theorem}
A detailed proof is provided in the Appendix.

\vspace{-1mm}\paragraph{Interpretation of evidential parameters.} The theoretical results given above not only establish the important relationship between ECNPs and CNPs but also unveil some key insights on why ECNPs along with an evidential Bayesian hierarchical prior is fundamentally more suitable for meta-learning based few-shot learning. As discussed in Section ~\ref{sec:ECNP}above, both evidential parameters $v_t$ and $\alpha_t$ can be interpreted as evidence of the model (in the form of pseudo counts). Meanwhile, the hierarchical structure of the NIG prior as defined in \eqref{eq:nig} indicates that $v_t$ and $\alpha_t$ capture the evidence at different levels, where $v_t$ corresponds to the evidence collected for the global knowledge in the form of the prior mean (\ie $\gamma$) whereas $\alpha_t$ provides evidence on the local knowledge in the form of the variance (\ie $\sigma$) on the per data sample level. Theorem~\ref{mainTheorem} shows that CNP primarily focus on improving the local knowledge by allowing $\alpha_t$ to grow while keeping $v_t$ very small (due to $\alpha_t v_t = \text{const}$). While this has the effect of using an uninformative prior (by assigning minimal evidence $v_t$ to the prior mean), it misses the opportunity to incorporate useful global knowledge that can be obtained through meta-learning from other relevant tasks. While using an uninformative prior is encouraged in a regular learning setting with sufficient training data, it is inherently inadequate for the few-shot setting, where there is not enough labeled data to support model training. 
\begin{figure}
    \begin{subfigure}[b]{0.22\textwidth}
\includegraphics[width=\linewidth]{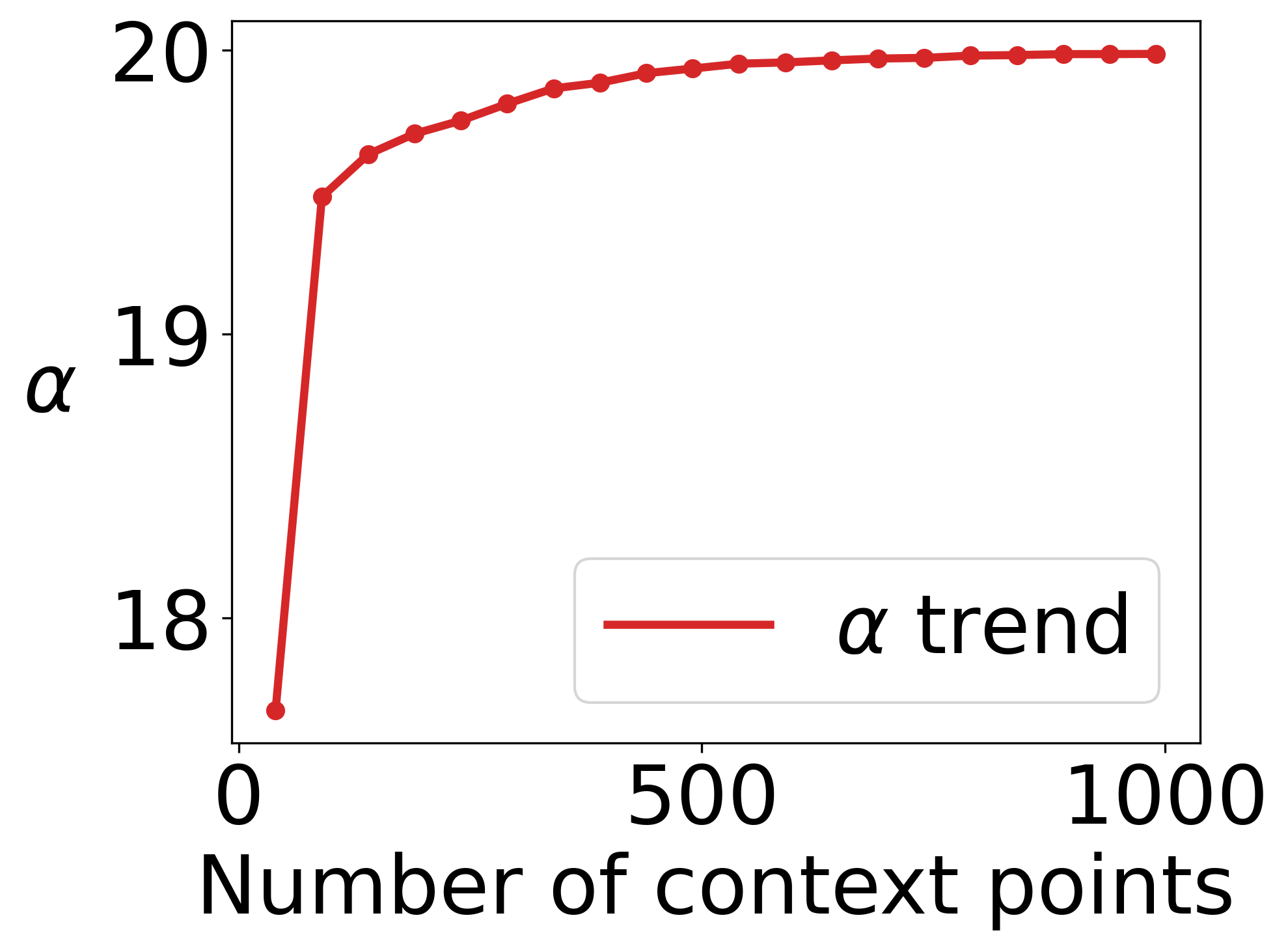}
\caption{Change of $\alpha$}
\end{subfigure}
\begin{subfigure}[b]{0.22\textwidth}

\includegraphics[width=\linewidth]{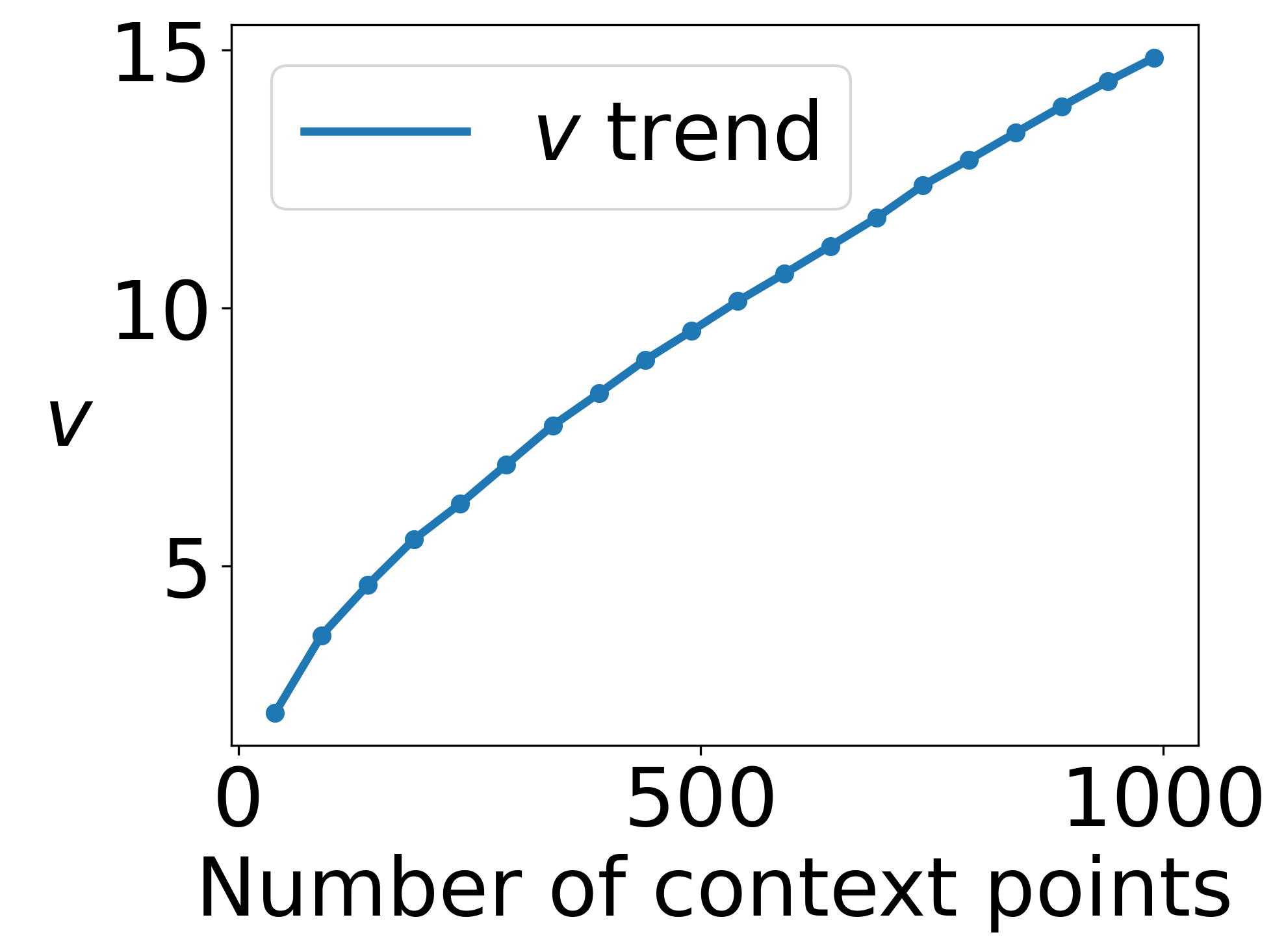}
\caption{Change of $v$}
\end{subfigure}
    \caption{Evidence change on a complex task}
    \label{fig:evipar}
\end{figure}

\begin{table*}[h]
\centering
\small
    \caption{\label{tab:5shotRegressionBig1}Comparison on 5-Shot Regression Problem 
    }    
    \vspace{-1mm}
\begin{tabular}{|p{0.07\textwidth}|p{0.24\textwidth}p{0.02\textwidth}p{0.10\textwidth}|p{0.24\textwidth}p{0.02\textwidth}p{0.10\textwidth}|}
\hline
Dataset: & Sinusoidal Regression &&& GP Regression&&\\
\hline
\end{tabular}
\begin{tabular}{|p{0.07\textwidth}|p{0.12\textwidth}p{0.12\textwidth}p{0.12\textwidth}|p{0.12\textwidth}p{0.12\textwidth}p{0.12\textwidth}|}
\hline
{\bf Model}&MSE($\downarrow$)&Inclusion@K($\uparrow$)&Unc. Increase($\uparrow$)&MSE($\downarrow$)&Inclusion@K($\uparrow$)&Unc. Increase($\uparrow$)\\
\hline \hline
NP              &0.1050$\pm$0.0200&0.192 $\pm$ 0.040&0.563 $\pm$ 0.005&0.348$\pm$0.0116&0.205 $\pm$ 0.019&0.686 $\pm$ 0.007\\
CNP             &0.0458$\pm$0.0074&0.144 $\pm$ 0.025&0.590 $\pm$ 0.006&0.3158$\pm$0.0038& 0.362 $\pm$ 0.027&0.783 $\pm$ 0.002\\
ANP             &0.3561$\pm$0.1084&0.351 $\pm$ 0.046&0.785 $\pm$ 0.048&0.3219$\pm$0.0124&0.318 $\pm$ 0.014&\bf{0.875 $\pm$ 0.026}\\
\textbf{ECNP}   &\bf{0.0391$\pm$0.0078}&0.205 $\pm$ 0.018&0.608 $\pm$ 0.013&\bf{0.3084$\pm$0.0014}&0.435 $\pm$ 0.02&0.798 $\pm$ 0.003\\
\textbf{ECNP-A} &0.2932$\pm$0.0956&\bf{0.437 $\pm$ 0.044}&\bf{0.814 $\pm$ 0.030}&0.3258$\pm$0.0162&\bf{0.505 $\pm$ 0.038}&\bf{0.875 $\pm$ 0.042}\\
\hline
\end{tabular}
\vspace{-1mm}
\end{table*} 

By leveraging a more expressive Bayesian hierarchical structure, ECNPs effectively address the key limitations of the CNPs as outlined above. In particular, they allow the evidence $v_t$ to grow with the global knowledge, which is particularly important for more complex few-shot tasks where the meta-knowledge could play a more critical role. Figure~\ref{fig:evipar} shows the change in local (\ie $\alpha$) and global (\ie $v$) evidence for different number of context points in a complex few-shot task (\ie image completion and details are provided in the experiment section). It is interesting to see that $\alpha$ grows fast and then shows a much slower increasing trend, which implies that the local knowledge may already reach the limit. On the other hand, $v$ continues to grow, which indicates that adding new context points can help retrieve more relevant global knowledge acquired through meta-learning. Meanwhile, the prediction error also continues to decrease (see Figure~\ref{fig:ANPEvidHyp} in the Appendix), which demonstrates effective knowledge transfer achieved by the ECNP model. 

\vspace{-1mm}
\section{Experiments}

\vspace{-1mm}\paragraph{Datasets.} 
For function regression experiments, we consider two synthetic datasets i) sinusoidal function regression \cite{gondal2021function}, and ii) regression on sample functions from a Gaussian process \cite{garnelo2018conditional}. The sinusoidal regression function is of the form $y = A \sin (x + \phi), A \in [0.1,5.0], \phi \in [0, \pi]$ and $x\in[-5,5]$ and  the GP is defined by a squared-exponential kernel with length scale of $0.6$, variance of $1.0$ and $x \in [-2,2]$. Each function regression task is defined by a $K$-shot context set with $K + u$ data points in the target set where $u \sim U(3,K)$, and $U(a,b)$ represents a uniform distribution in range $(a,b)$. Moreover, the function regression models are trained for 30,000 meta-iterations using a batch of 8 tasks and evaluated on 2,000 test tasks. For Image completion experiments, we consider three benchmark datasets: MNIST,
\cite{lecun1998mnist}
CelebA,
\cite{liu2015faceattributes}, 
and Cifar10
\cite{krizhevsky2009learning}.
The details of the benchmark datasets are summarized in Appendix Table \ref{tab:1dataset_details}. Image completion task is created by randomly selecting a subset of the set points (input-output pairs) from an image. Specifically, each position in the image grid is the input and the pixel value (\eg the RGB value) is the output. We randomly select 50 points to make the context set, use the remaining points in the image to make the target set, and train the models for 50 epochs using a batch of 8 tasks, and evaluate the model on the test set. 

\vspace{-1mm}\paragraph{Baselines.}  
We consider three baseline models: Neural Processes (NP)~\cite{garnelo2018neural}, Conditional Neural Processes (CNP)~\cite{garnelo2018conditional}, and the Attentive Neural Process (ANP)~\cite{kim2018attentive}. For a fair comparison to the baselines, we consider the evidential equivalent of the baselines with the same encoder and decoder architectures. Specifically, for our evidential models, we consider two variants: i) ECNP: evidential model with deterministic path similar to CNP, and ii) ECNP-A: the evidential model with multi-head attention mechanism in encoder similar to ANP. Additional details of the model architecture and training are presented in the Appendix.

\vspace{-1mm}\subsection{Performance Evaluation}
\begin{figure}
\vspace{-1mm}
\centering
\begin{subfigure}[b]{0.22\textwidth}
\centering
\includegraphics[width=\linewidth]{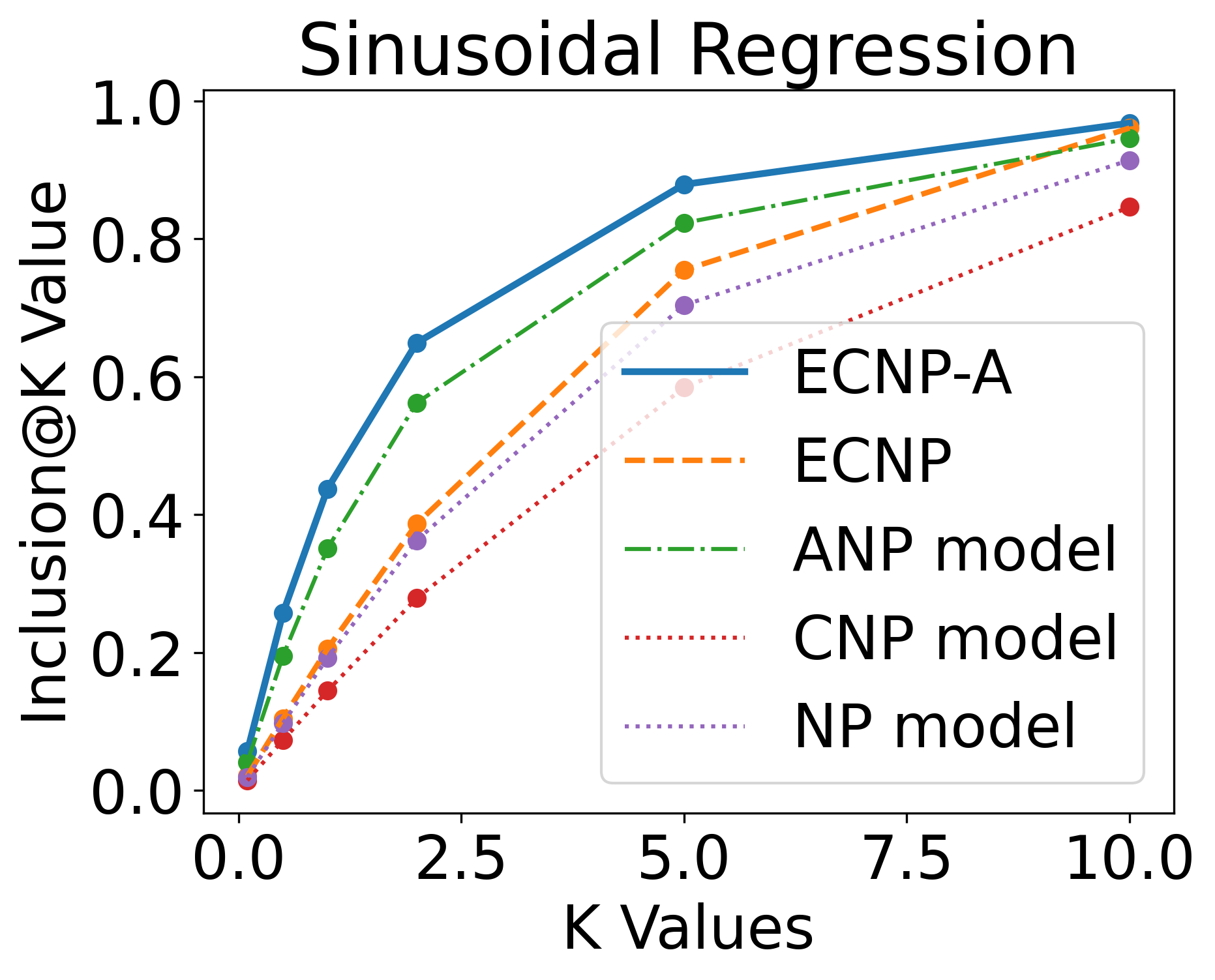}
\vspace{-1mm}
\caption{Sinusoidal Regression}
\end{subfigure}
\begin{subfigure}[b]{0.22\textwidth}
\centering
\includegraphics[width=\linewidth]{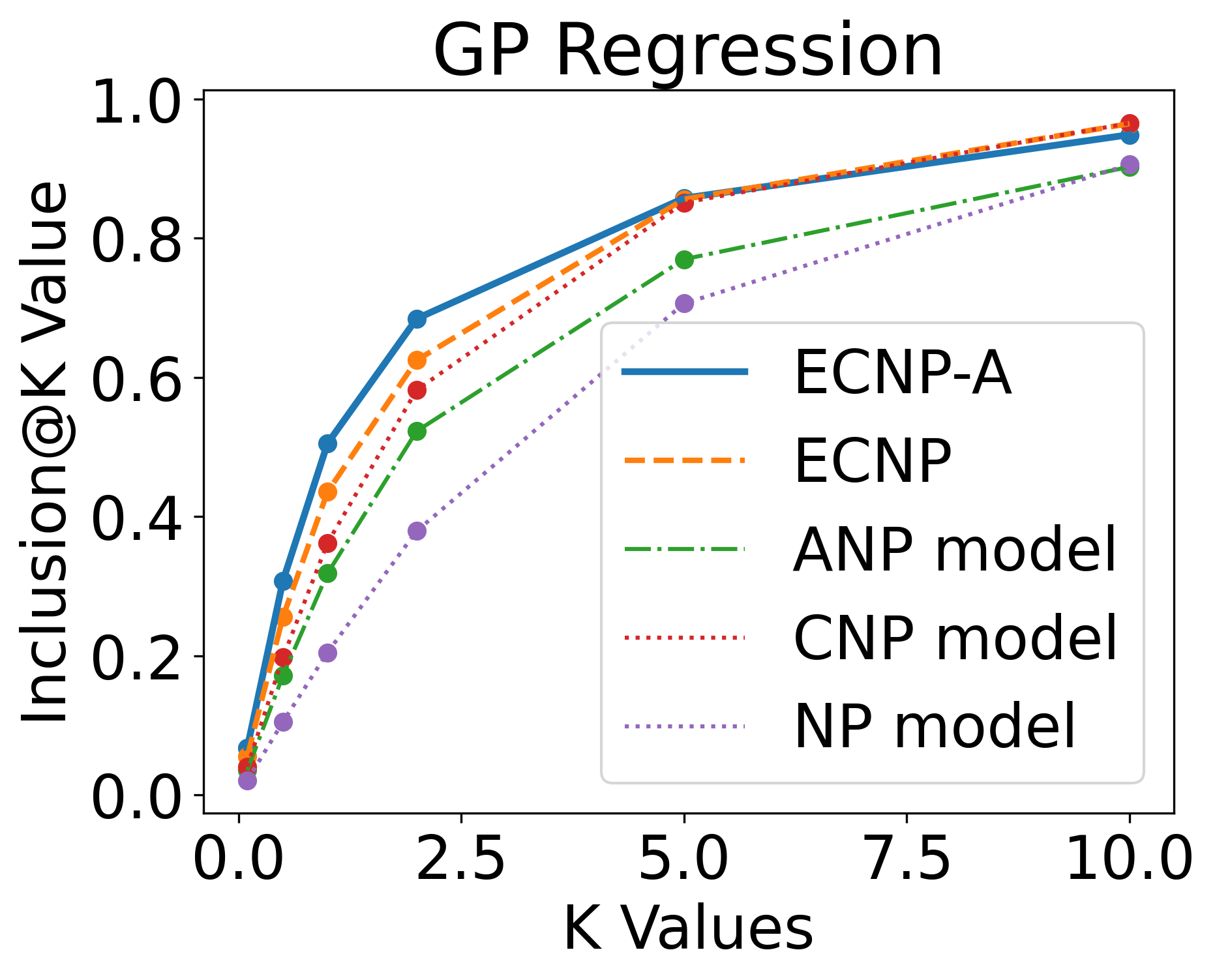}
\vspace{-1mm}
\caption{GP Regression}
\end{subfigure}
\vspace{-1mm}
\caption{
Impact of K to Inclusion@K }
\label{fig:inclusionKAblation}\vspace{-1mm}
\end{figure}

\vspace{-1mm}
In this set of experiments, we report the generalization performance in terms of Mean Squared Error (MSE) along with three uncertainty based evaluation metrics: Log Likelihood (LL), Inclusion @K, and Uncertainty-Increase \cite{groverprobing} for all the models on function regression and image completion tasks. We consider Inclusion@K 
with $K=1$ 
in Table~\ref{tab:5shotRegressionBig1} and Table \ref{tab:overallResultsGen50Shot}. Inclusion and Uncertainty-Increase have been developed to analyze and compare the uncertainty estimates of NP based models. Additional details along with comparisons are presented in the Appendix. We also empirically verify their robustness to outliers for function regression and image completion tasks. Limited by space, we present ablation studies in the Appendix. 
\vspace{-1mm}\paragraph{Function regression.}
In the function regression problem, the model has to learn the underlying function based on the limited information of the context set and the meta-knowledge. Table~\ref{tab:5shotRegressionBig1} shows the results for 5-shot regression experiments. Our model improves  the generalization performance compared to the the corresponding baseline model across almost all the datasets. Moreover, when considering the uncertainty metrics, as shown in Table~\ref{tab:5shotRegressionBig1} and Figure \ref{fig:inclusionKAblation}, our model considerably improves over the baselines.

\vspace{-1mm}\paragraph{2D image completion.}
We consider image completion experiments similar to \cite{eslami2018neural}, where the model needs to infer the underlying function $f: [0,1]^2 \rightarrow [0,1]^{ch}$ ($ch - $ number of channels) to make prediction for each image pixel position in the target set given the context set. Table \ref{tab:overallResultsGen50Shot} compares our model with the baselines for 50-shot experiments. As can be seen, our model leads to comparable to improved performance than corresponding baselines in terms of MSE and log likelihood. We compare the uncertainty behavior (using Inclusion@K ($K = 1$) and uncertainty increase) of the representative CNP and the corresponding ECNP models in Table \ref{tab:uncInclusionRepTable}. Additional results are presented in the Appendix. 

As can be seen in function regression and 2D image completion experiments, our model has better uncertainty characteristics than the baselines which is mainly due to the fine-grained and accurate uncertainty guidance capabilities in our hierarchical model. Our model explicitly captures the aleatoric and epistemic uncertainties through the evidence parameters ($\beta, \alpha, v$). Furthermore, the model is guided during training to have accurate overall uncertainty via the evidence regularization in \eqref{eq:evidence_regularization}, and accurate epistemic uncertainty from the kernel regularization in \eqref{kernel_regularization}. Such uncertainty guidance leads to more accurate uncertainty performance in our model. These results empirically validate our model's generalization performance and superiority over other comparison baselines. 
\begin{table*}[h!]
\centering
\small
    \caption{Comparison on 50-Shot Image Completion Problems
    }
    \label{tab:overallResultsGen50Shot}
    \vspace{-1mm}
\begin{tabular}{|p{0.07\textwidth}|p{0.12\textwidth}p{0.12\textwidth}|p{0.12\textwidth}p{0.12\textwidth}|p{0.12\textwidth}p{0.12\textwidth}|}
\hline
Dataset & MNIST && Cifar10 && CelebA &\\
\hline
{\bf Model}&MSE($\downarrow$)&LL($\uparrow$)&MSE($\downarrow$)&LL($\uparrow$)&MSE($\downarrow$)&LL($\uparrow$)\\
\hline \hline
NP              &0.048$\pm0.001$&0.538$\pm$0.010&0.027$\pm$0.000&0.434$\pm$0.003&0.025$\pm$0.000&0.433$\pm$0.006\\
CNP             &0.044$\pm$0.001&0.710$\pm$0.009 & 0.023$\pm$0.000&0.576$\pm$0.005 & 0.021$\pm$0.001&0.660$\pm$0.004\\
ANP             &0.045$\pm$0.001&0.702$\pm$0.007 & 0.017$\pm$0.000&\textbf{0.765}$\pm$0.004 & \textbf{0.014}$\pm$0.000&0.850$\pm$0.002\\
\textbf{ECNP}  &\textbf{0.041}$\pm$0.002&\textbf{0.734}$\pm$0.014&0.022$\pm$0.001 &0.601$\pm$0.004&0.020$\pm$0.000&0.694$\pm$0.004\\
\textbf{ECNP-A}  &0.043$\pm$0.001&0.713$\pm$0.013&\textbf{0.016}$\pm$0.001&0.764$\pm$0.004&\textbf{0.014}$\pm$0.000&\textbf{0.852}$\pm$0.002\\
\hline
\end{tabular}
\vspace{-1mm}
\end{table*}

\begin{table}[h!]
\centering
\small
    \caption{Comparison of representative CNP and ECNP models on Inclusion and Uncertainty Increase Metrics for 50-Shot Image Completion Problem
    }
    \label{tab:uncInclusionRepTable}
\begin{tabular}{|p{0.41\textwidth}|}
\hline
Metric: \textbf{Inclusion@K ($\uparrow$)}\\
\hline
\end{tabular}
\begin{tabular}{|p{0.12\textwidth}||p{0.12\textwidth}|p{0.12\textwidth}|}
\hline
\textbf{Dataset} & CNP model&ECNP model\\
\hline
MNIST &0.622 $\pm$ 0.001&\textbf{0.828 $\pm$ 0.000}\\
Cifar10 & 0.129 $\pm$ 0.005&\textbf{0.144 $\pm$ 0.003}\\
CelebA &0.133 $\pm$ 0.004&\textbf{0.156 $\pm$ 0.003}\\
\hline
\end{tabular}
\begin{tabular}{|p{0.41\textwidth}|}
Metric: \textbf{Uncertainty Increase ($\uparrow$)}\\
\hline
\end{tabular}
\begin{tabular}{|p{0.12\textwidth}||p{0.12\textwidth}|p{0.12\textwidth}|}
\hline
\textbf{Dataset} & CNP model&ECNP model\\
\hline
MNIST &0.306 $\pm$ 0.000&\textbf{0.524 $\pm$ 0.000}\\
Cifar10 &0.505 $\pm$ 0.005&\textbf{0.531 $\pm$ 0.004}\\
CelebA &0.519 $\pm$ 0.003&\textbf{0.541 $\pm$ 0.003}\\
\hline
\end{tabular}
\end{table} 

\vspace{-1mm}\paragraph{Outlier robustness.}
Due to the hierarchical Bayesian structure leading to a heavy tailed predictive distribution, our model is theoretically guaranteed to be robust to outliers in the training tasks. Here, we empirically validate the claim by experimenting with 5-shot sinusoidal regression and 50-shot MNIST image completion (results on other datasets and settings are presented in the Appendix). 

\begin{figure}
\centering
\begin{subfigure}[b]{0.22\textwidth}
\centering
\includegraphics[width=\linewidth]{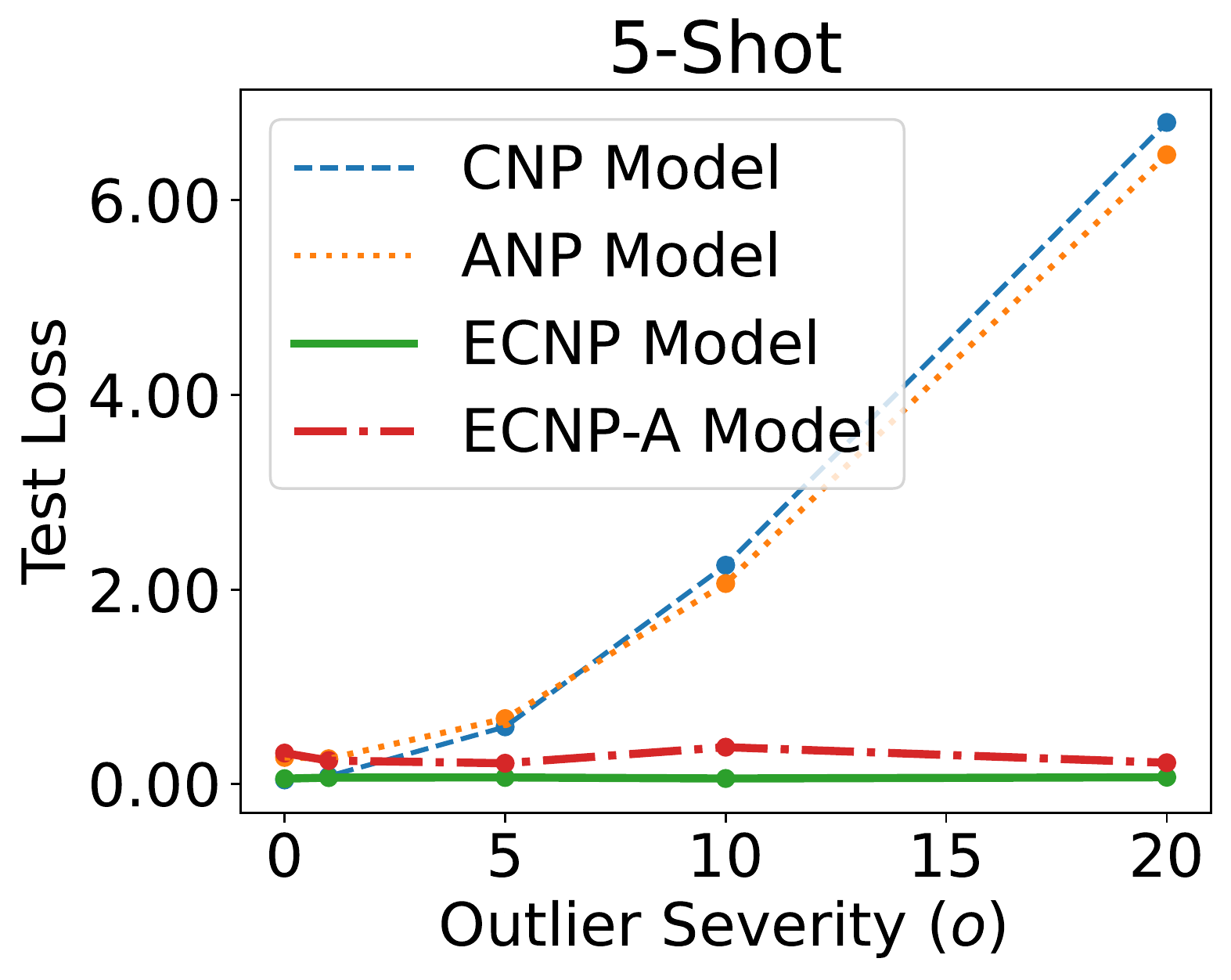}
\vspace{-1mm}
\caption{Sinusoid}
\end{subfigure}
\begin{subfigure}[b]{0.22\textwidth}
\centering
\includegraphics[width=\linewidth]{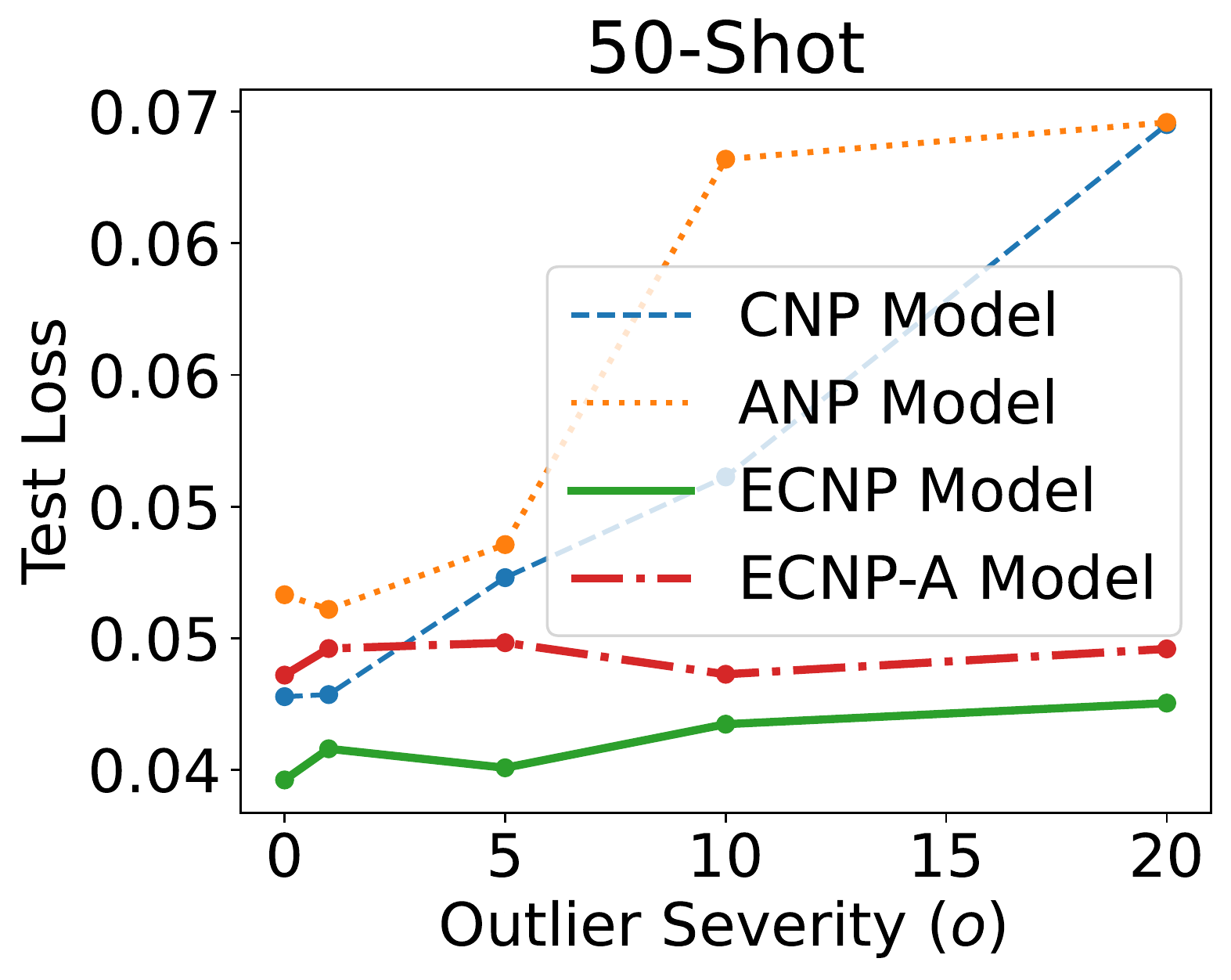}
\vspace{-1mm}
\caption{MNIST}
\end{subfigure}
\vspace{-1mm}
\caption{
Outlier robustness performance} 
\label{fig:outlierRobustness}\vspace{-1mm}
\end{figure}
To make the noisy training task, we randomly select one target point in all the training tasks and apply an additive transformation $y_t = y_t + o$ to make it an outlier (here $o$ determines the outlier severity). We train the models on the noisy tasks (\ie tasks with outlier), and after training, we evaluate on clean test set tasks. Figure \ref{fig:outlierRobustness} (a)-(b) shows the comparison results of the ECNP and ECNP-A models with their corresponding baselines of CNP and ANP models. Across both experiments, ECNP models remain robust to the outlier as their test set performance remains relatively unaffected even for severe outliers. Such outlier robustness in our model can be attributed to  the heavy tailed predictive distribution that is inherently introduced by the hierarchical structure. In comparison, the baseline models lack the required robustness characteristics and their performance degrades severely as the outlier becomes extreme. Such baseline models may require additional mechanisms to handle the outliers, something our model can automatically do. These results empirically validate the robustness superiority for the proposed ECNP model.

\vspace{-1mm}
\subsection{Effectiveness of Uncertainty Decomposition}\vspace{-1mm} 
In this set of experiments, we show that the proposed ECNP models can capture fine-grained uncertainty to best support few-shot learning through epistemic-aleatoric (EP-AL) uncertainty decomposition that can enable active context set construction and effective meta-knowledge transfer. 

\begin{figure}[h!]
\vspace{-1mm}
\centering
\begin{subfigure}[b]{0.22\textwidth}
\centering
\includegraphics[width=\linewidth]{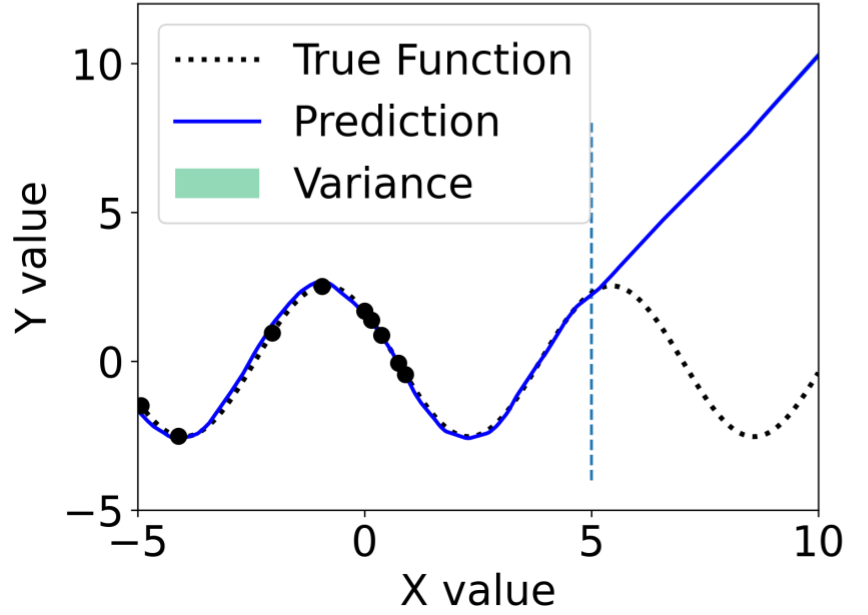}
\caption{CNP}
\end{subfigure}
\begin{subfigure}[b]{0.22\textwidth}
\centering
\includegraphics[width=\linewidth]{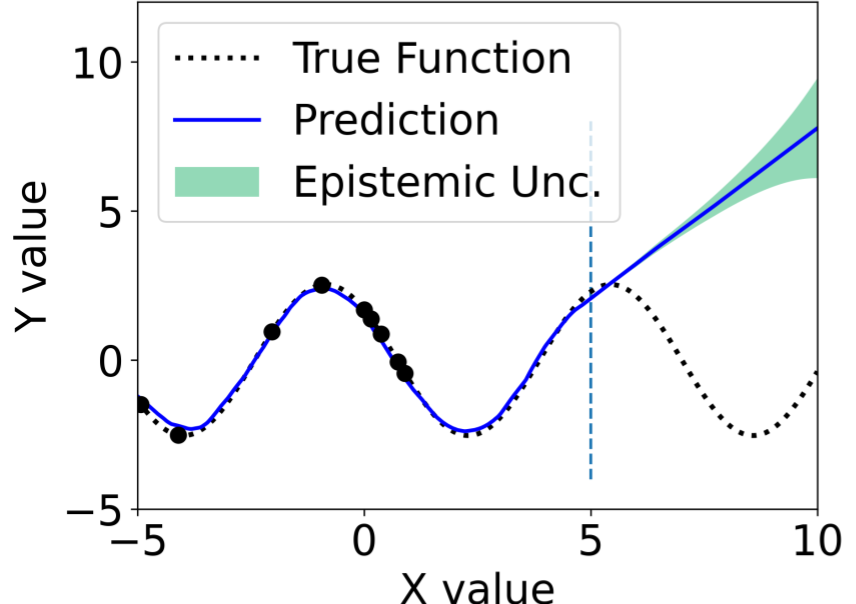}
\caption{ECNP}
\end{subfigure}
\begin{subfigure}[b]{0.22\textwidth}
\centering
\includegraphics[width=1.1\linewidth]{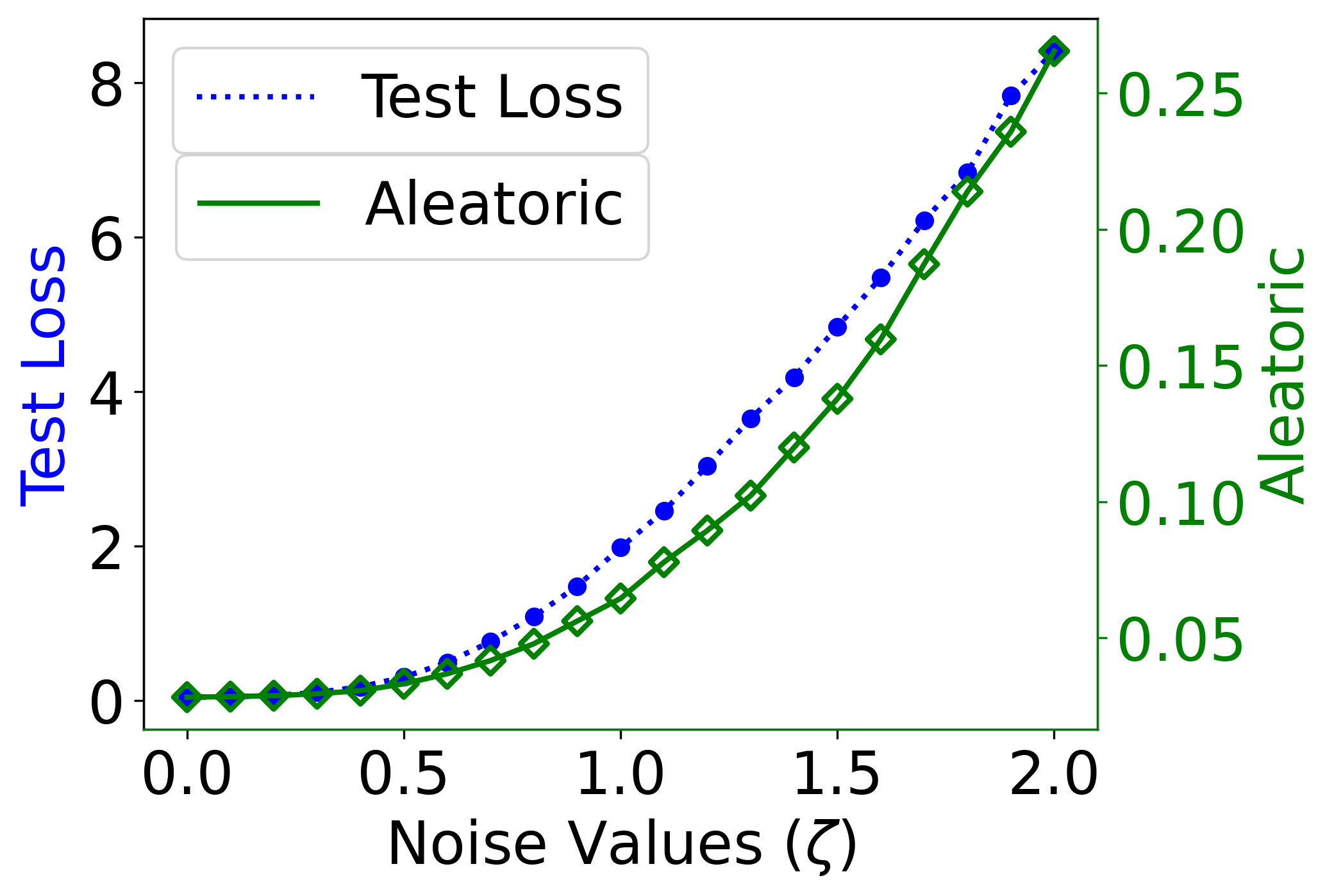}
\caption{Sinusoidal Regression}
\end{subfigure}
\begin{subfigure}[b]{0.22\textwidth}
\centering
\includegraphics[width=1.1\linewidth]{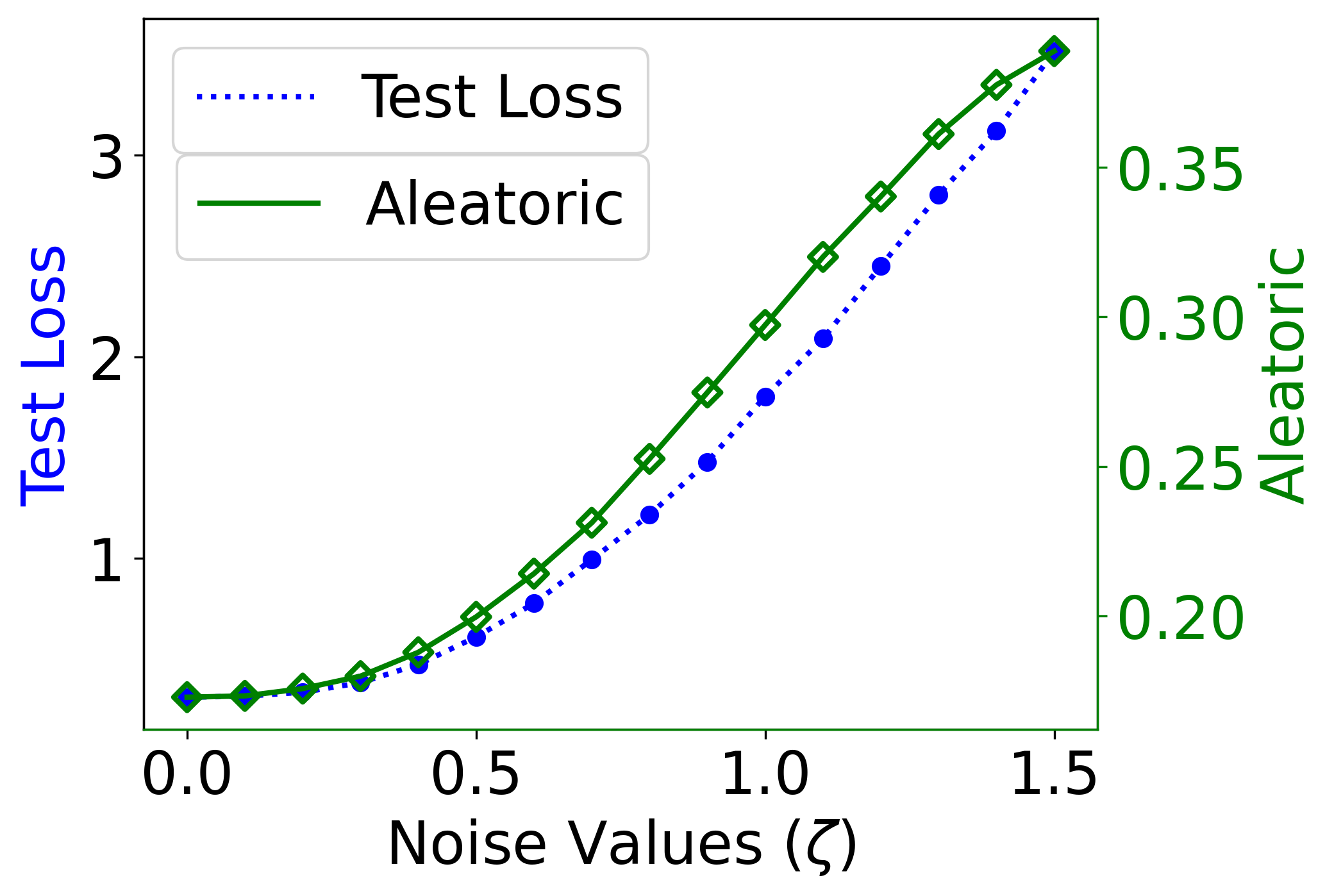}
\caption{GP Regression}
\end{subfigure}
\vspace{-0mm}
\caption{
(a)-(b) ECNP vs. CNP on a sinusoidal task; (c)-(d) ECNP performance for noisy test tasks} 
\label{fig:epistemicUsefulness}\vspace{-1mm}
\end{figure}

\vspace{-1mm}\paragraph{EP-AL decomposition.}
Our proposed model can perform Epistemic-Aleatoric uncertainty decomposition for any test task. Here, we compare the predicted uncertainty for the proposed ECNP model with the respective CNP baseline in sinusoidal regression task. Both models are trained for 20,000 iterations using training tasks with data in range $[-5,5]$. As shown in Figure \ref{fig:epistemicUsefulness} (c)-(d), outside the training range (\ie $x_t \in [5, 10]$), prediction from both CNP and ECNP is inaccurate as expected. The CNP model continues to remain confident in regions far from the data whereas our ECNP model correctly outputs high epistemic uncertainty in the regions far away from the observed data.  

Next, we experiment with noisy test tasks to analyze the aleatoric uncertainty of our proposed model. We consider a model trained on clean 5-shot regression tasks and evaluate on 5-shot noisy test tasks. Specifically, we add random Gaussian noise to the context set of the test tasks ($y_c = y_c + \zeta \epsilon, \epsilon \sim \mathcal{N}(0,1) )$ and vary the level of noise (\ie $\zeta$) to study the model behavior. Figure \ref{fig:epistemicUsefulness} visualizes the impact of the noise on the predicted performance (MSE) and the model's predicted aleatoric uncertainty for two datasets averaged across 2000 test tasks. As expected, the model's predictive accuracy decreases as tasks become more noisy. Our proposed model accurately identifies the noisy tasks and outputs more aleatoric uncertainty as tasks become more noisy showing the effectiveness of our model's predicted aleatoric uncertainty in identifying noisy tasks.

\vspace{-1mm}\paragraph{Active context set construction.}  
\begin{figure}[h!]
        \vspace{-0mm}
        \includegraphics[width=0.48\textwidth]{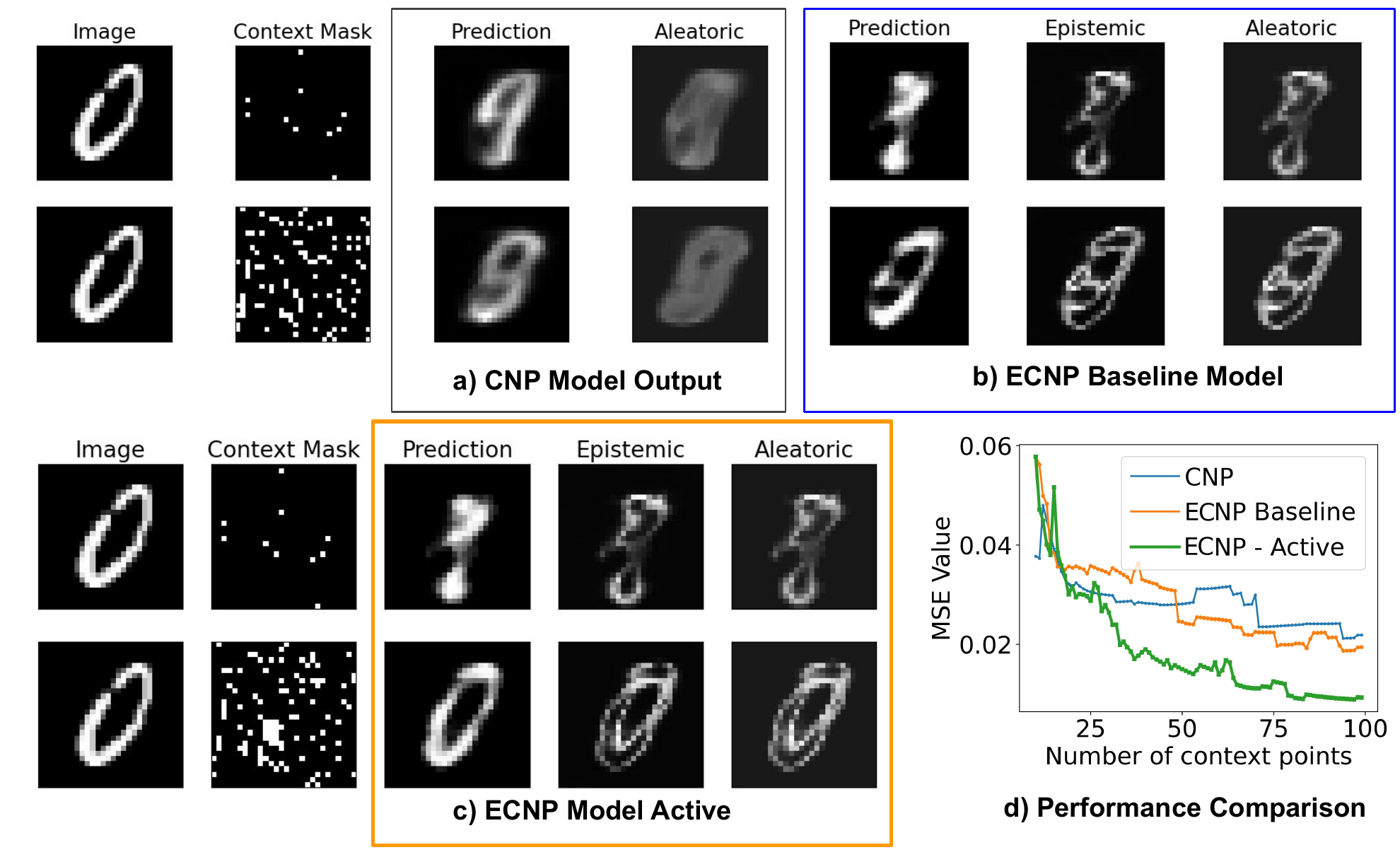}
        \centering
        \caption{Active context set construction
        }
        \label{fig:cnpRes} 
        \vspace{-2mm}
\end{figure}
The proposed ECNP model can capture both aleatoric and epistemic uncertainty in a single forward pass. We investigate the effectiveness of the captured epistemic uncertainty in a context point selection experiment (Figure \ref{fig:cnpRes}). We randomly select a test image and both models start with random 10 context points as visualized by the context mask (CM), which represents the pixel positions that are included in the context set. CNP model and ECNP model randomly select the next 100 context points. ECNP-Active model iteratively queries the epistemic uncertainty for different target positions and includes the queried data with the greatest epistemic uncertainty in the context set for the next iteration. By including the most informative points in the context set using the epistemic uncertainty information, ECNP-Active performs significantly better than the other models (Figure \ref{fig:cnpRes} (d)) illustrating the effectiveness of our proposed model's uncertainty.

\vspace{-1mm}\subsection{Ablations Study and Additional Experiments}
We carry out a detailed ablation study to investigate some key model parameters. The results along with some additional illustrative examples  are presented in the Appendix. 

\vspace{-1mm}\section{Conclusion}\vspace{-1mm}
We propose evidential conditional neural processes, that can conduct epistemic-aleatoric uncertainty decomposition in few-shot learning. ECNPs introduce a hierarchical Bayesian structure to replace the standard Gaussian distribution. The hierarchical bayesian structure enables the model to quantify fine-grained uncertainty in an efficient way. Moreover, our theoretical results reveal a deep connection with the CNP models and further justify why a richer hierarchical structure provides a more principled way to capture the meta-knowledge through higher-order priors, making it fundamentally more suitable for meta-learning over few-shot tasks. Experiments over various 1D regression and 2D image completion tasks demonstrate the superiority of our proposed model and its uncertainty capabilities. 

\section{Acknowledgement}
This research was supported in part by an NSF IIS award IIS-1814450 and an ONR award N00014-18-1-2875. The views and conclusions contained in this paper are
those of the authors and should not be interpreted as representing any funding agency.

\bibliography{aaai2023FinalDraft}

\appendix

\clearpage
\newpage
\appendix
\begin{center}
\large{\bf Appendix}    
\end{center}
\paragraph{Organization of Appendix.} We first present  the proof of the theoretical results presented in the main paper. We then present the analysis of the hierarchical Bayesian structure introduced in the ECNP model that extends evidential learning to meta-learning models and discuss additional relevant works in evidential learning. Next, we provide the details of the datasets, experimental settings, model architectures, and discuss complexity. We then carry out an ablation study to demonstrate the effect of different parameters. We then provide additional experimental results across different datasets and settings that demonstrate the effectiveness of the proposed evidential model. Finally, we discuss some limitations of the work that we aim to address in our future works. The codes for all the experiments in this paper can be found at this link\footnote{Source codes:https://github.com/pandeydeep9/ECNP}.
\section{Proofs of Theoretical Results}\label{app:proofs}
In this section, we show the proofs of the theoretical results presented in the main paper. 
\subsection{Theorem 1}
\textit{The ECNP model with a hierarchical Bayesian structure in the decoder is  guaranteed to be more robust to outliers in the training tasks as compared to the CNP models that use a Gaussian structure.}
\label{sec:outlierproof}
\begin{definition}

\textit{Outlier:} Consider a task defined by an underlying generating function $f(\cdot)$. A data point $(x_o,y_o)$ in the target set of the task is an outlier with severity $os$ if the value of the output $y_o$ deviates from the ground truth value $y_{true}$ with a margin of $os$. i.e $|y_{true}-y_{o}|>os$,  $y_{true} = f(x_o)$. 
\end{definition}
\begin{proof}
Consider we have a task with the context set $\mathcal{C}$, and $N_t$ input output pairs of the form ($x_t, y_t$) in the target set. The meta-learning model has to be able to correctly predict for each target set input $x_t$ after learning from the context set $\mathcal{C}$.

Consider a CNP model that outputs the mean $\gamma_t$ and variance $s^2_t$ for a target input $x_t$ given the context set $\mathcal{C}$. Consider the model parameters $\psi_{\gamma}$ output the prediction for the target set i.e.  $f_{\psi_{\gamma}}(x_t|\mathcal{C}) = \gamma_t$. Now, for a task with $N_t$ points in the target set, the Maximum Likelihood estimate of the model parameters $\psi_{\gamma}$ is given by
\begin{align}
    \frac{\partial \mathcal{L}}{\partial \psi_{\gamma}} &=  \frac{\partial \big(\sum_{t=1}^{N_t} L_t^{NLL}\big)}{\partial \psi_{\gamma}} \\
    &=  \frac{\partial \big(\sum_{t=1}^{N_t}-\log\mathcal{N}(y_t|\gamma_t, s_t^2)\big)}{\partial \psi_{\gamma}}= 0 \\
    or,& \sum_{t=1}^{N_t}\
    s_t^{-2}(y_t - \gamma_t) \frac{\partial \gamma_t}{\partial \psi_\gamma}  = 0 
\end{align}
Consider the ECNP model that outputs the prediction $\gamma_t$ along with the evidential parameters $v_t$, $\alpha_t$ and $\beta_t$ leading to scale parameter $s_t$ and $2\alpha_t$ degrees of freedom. Consider the model is trained without regularization and assume that the model parameters $\psi_{\gamma}$ output the prediction $\gamma_t$ are  i.e. $f_{\psi_{\gamma}}(x_t|\mathcal{C}) = \gamma_t$. Now, for a task with $N_t$ points in the target set, the Maximum Likelihood estimate of the model parameters $\psi_{\gamma}$ is given by
\begin{align}
    \frac{\partial \mathcal{L}}{\partial \psi_{\gamma}} 
    &=  \frac{\partial \big(\sum_{t=1}^{N_t} L_t^{NLL}\big)}{\partial \psi_{\gamma}}=0 \\
     or,&  \frac{\partial \big(\sum_{t=1}^{N_t} - \log St(y_t|\gamma_t, s_t, 2\alpha_t) \big)}{\partial \psi_{\gamma}} = 0 \\
    or,& \sum_{t=1}^{N_t} \frac{ \partial \Big((\alpha_t + \frac{1}{2}) \log ( 2 \alpha_t + s_t^{-2}(y_t - \gamma_t)^2\Big)}{\partial\gamma_t} \frac{\partial \gamma_t}{\partial \psi_\gamma}  = 0 \\
    or,& \sum_{t=1}^{N_t} \frac{(2\alpha_t + 1)}{(2 \alpha_t + \delta_t^2)} s_t^{-2}(y_t - \gamma_t) \frac{\partial \gamma_t}{\partial \psi_\gamma}  = 0 \\
    or,& \sum_{t=1}^{N_t} w_t s_t^{-2}(y_t - \gamma_t) \frac{\partial \gamma_t}{\partial \psi_\gamma}  = 0 
\end{align}
where $\delta_t^2 = s_t^{-2}(y_t - \gamma_t)^2$ is the Mahalanobis distance between the prediction and the ground truth, and $w_t = \frac{(2\alpha_t + 1)}{(2 \alpha_t + \delta_t^2)}$ is the outlier dependent scaling factor. As the outliers in the target set of training tasks become more extreme, $\delta_t^2$ increases, outlier scaling factor $w_t$ decreases proportionally for ECNP model to down-weight the impact of the outliers in estimation of the model parameters, effectively enabling the ECNP model to be robust to outliers.
\end{proof}

\paragraph{Remark 1.} The ECNP model is least robust to outliers when the ECNP model realizes the CNP model, \ie $2\alpha_t \rightarrow \infty, \ \& \ \alpha_t v_t = const$ $\implies$ $w_t = 1$.

\paragraph{Remark 2.} The robustness ECNP model and the maximum likelihood estimate of the parameters $\psi_{\gamma}$ are unaffected by the proposed kernel based regularization. 

\paragraph{Impact to generalization:} As shown by Theorem 1, ECNP is robust to outliers in training tasks. The robustness ensures that the model learns from true signals, avoiding outliers, which is expected to improve generalization compared to a less robust model. 

\subsection{Empirical Validation}
\begin{figure*}[h!]
\vspace{-1mm}
\centering
\begin{subfigure}{0.23\textwidth}
\centering
\includegraphics[width=0.9\linewidth]{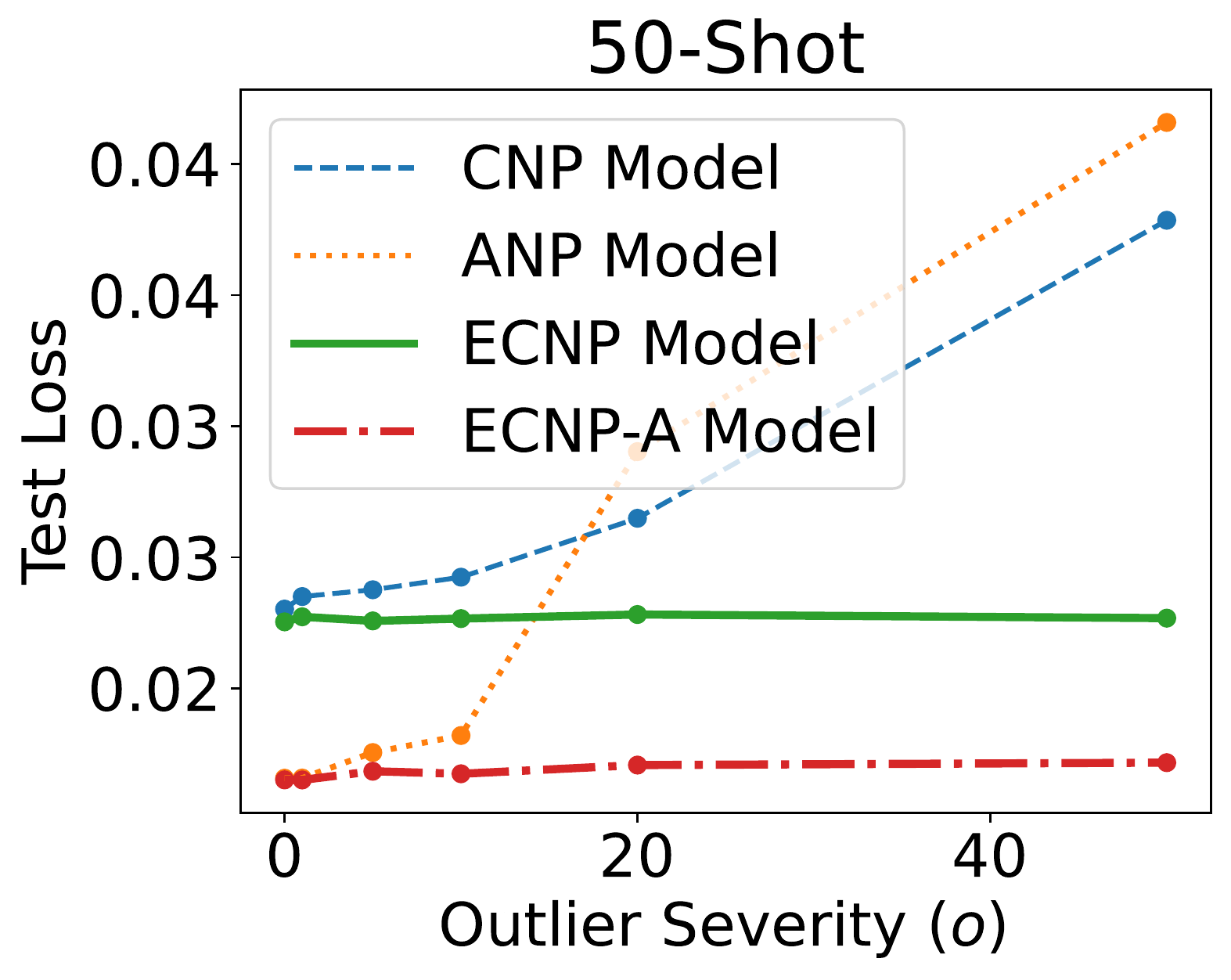}
\caption{Cifar10-MSE}
\end{subfigure}
\begin{subfigure}{0.23\textwidth}
\centering
\includegraphics[width=0.9\linewidth]{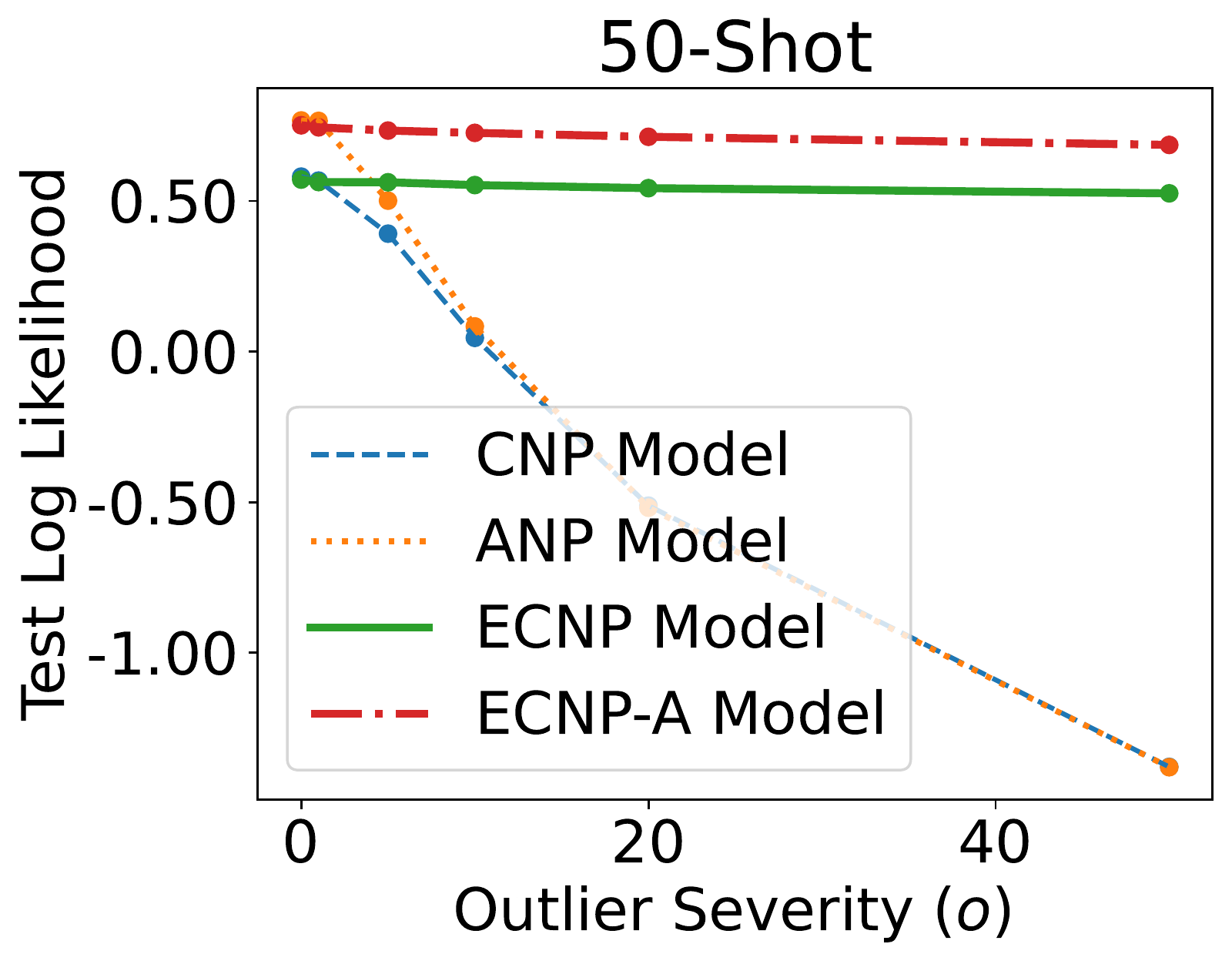}
\caption{Cifar10-LL}
\end{subfigure}
\begin{subfigure}{0.23\textwidth}
\centering
\includegraphics[width=0.9\linewidth]{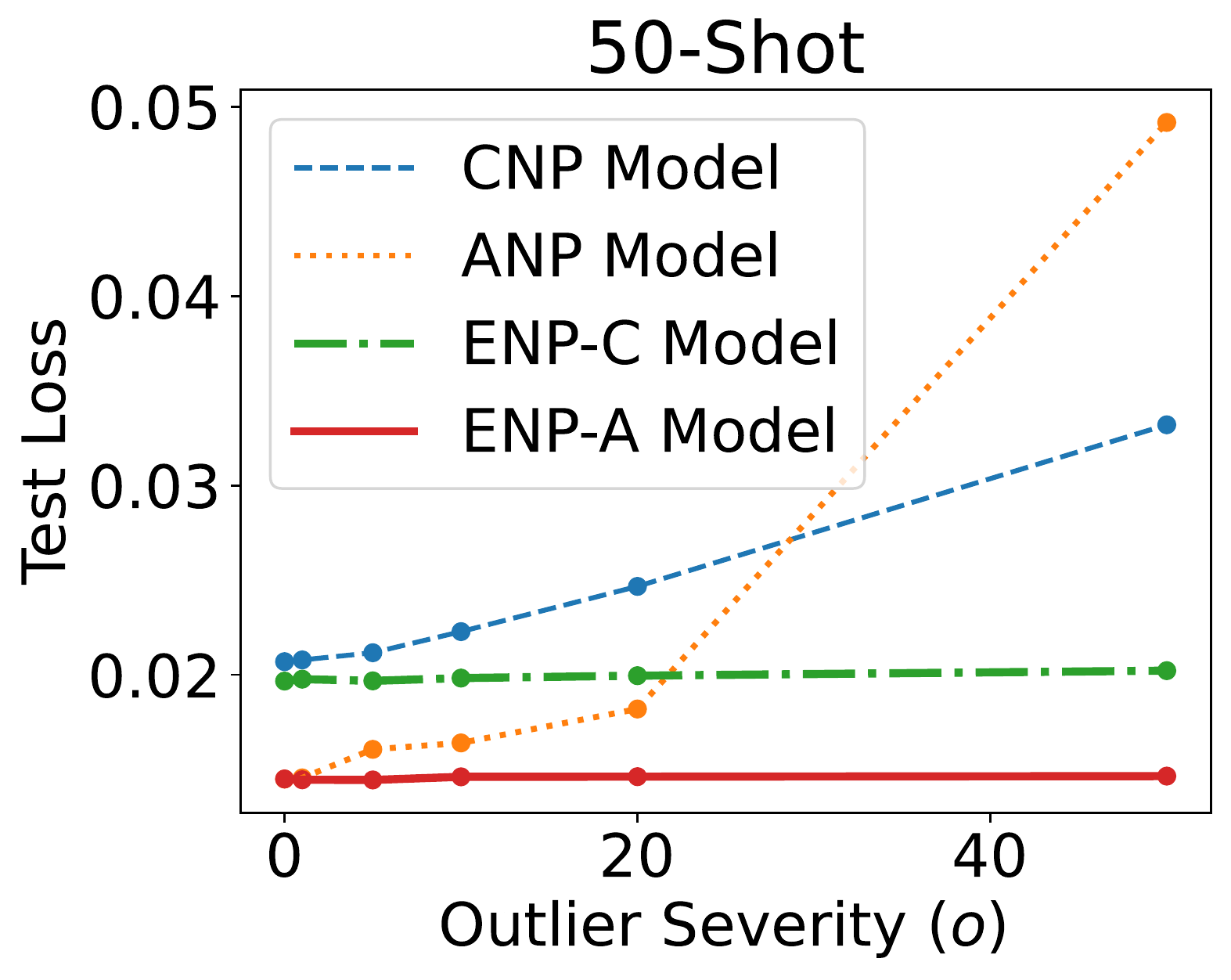}
\caption{CelebA-MSE}
\end{subfigure}
\begin{subfigure}{0.23\textwidth}
\centering
\includegraphics[width=0.9\linewidth]{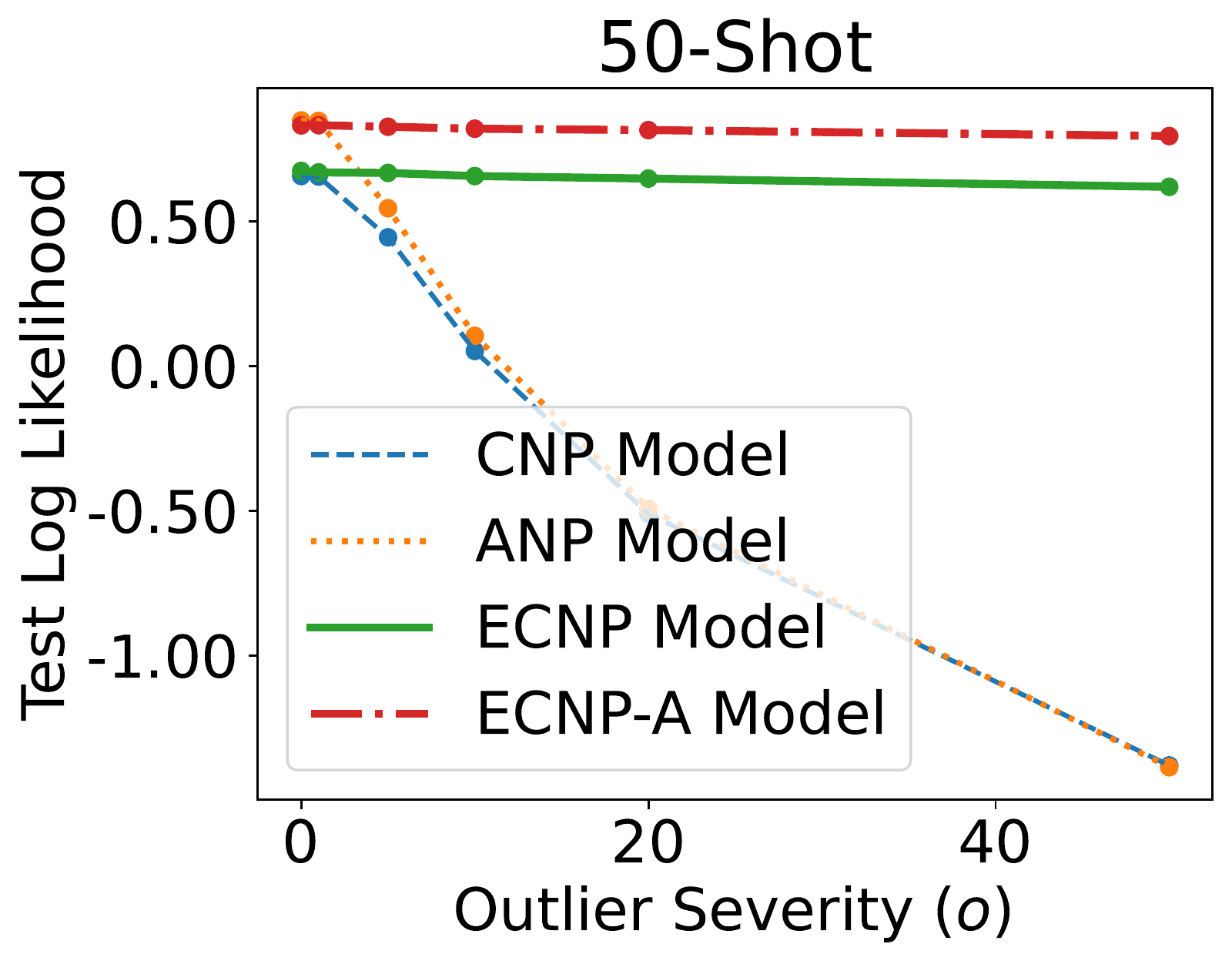}
\caption{CelebA-LL}
\end{subfigure}
\caption{Impact of outlier to NP based models for different 50-shot image completion tasks.}
\vspace{-1mm}
\label{fig:appendixOutlierAll50shot}

\end{figure*}

\begin{figure*}[h!]
\vspace{-1mm}
\centering
\begin{subfigure}{0.23\textwidth}
\centering
\includegraphics[width=0.9\linewidth]{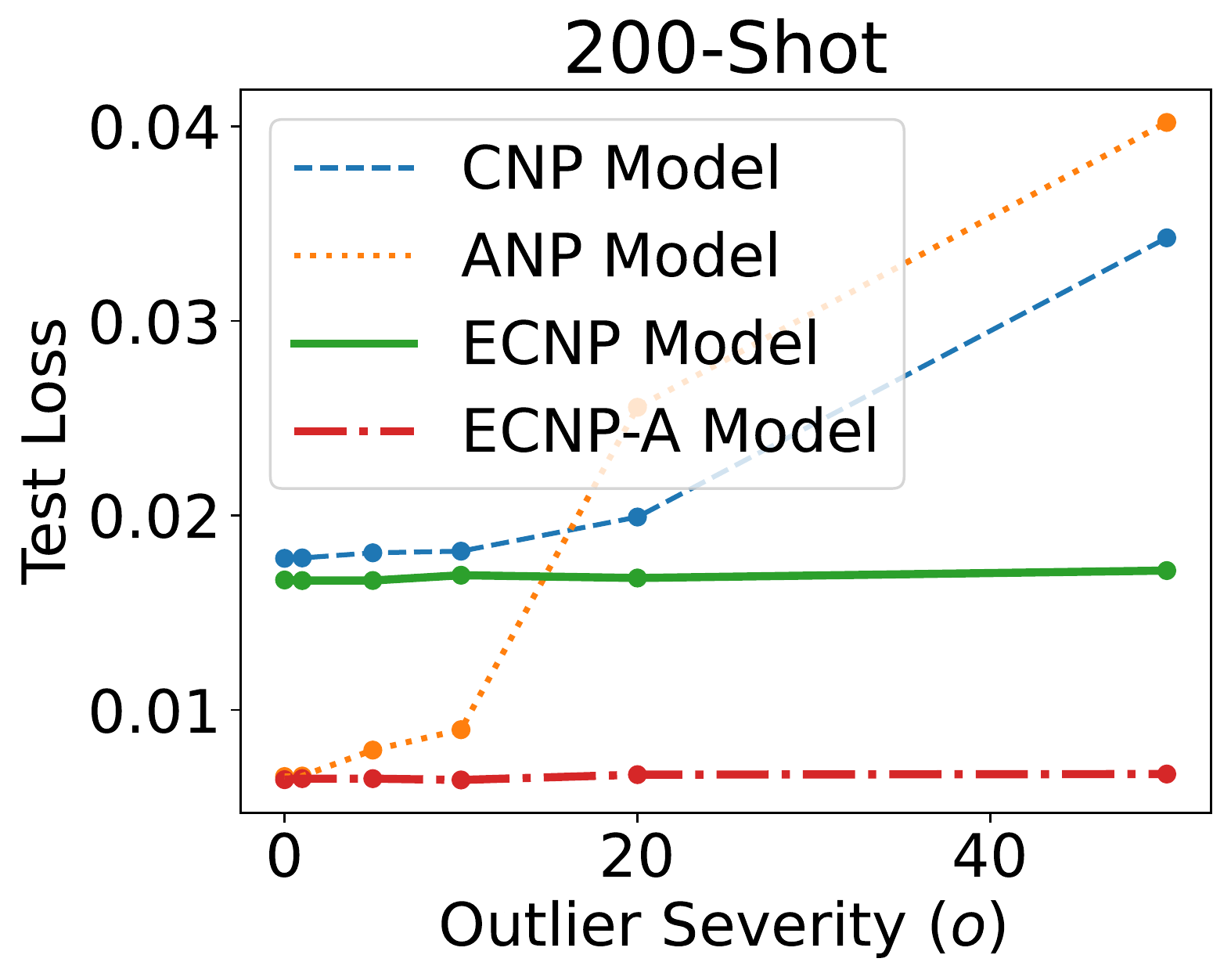}
\caption{Cifar10-MSE}
\end{subfigure}
\begin{subfigure}{0.23\textwidth}
\centering
\includegraphics[width=0.9\linewidth]{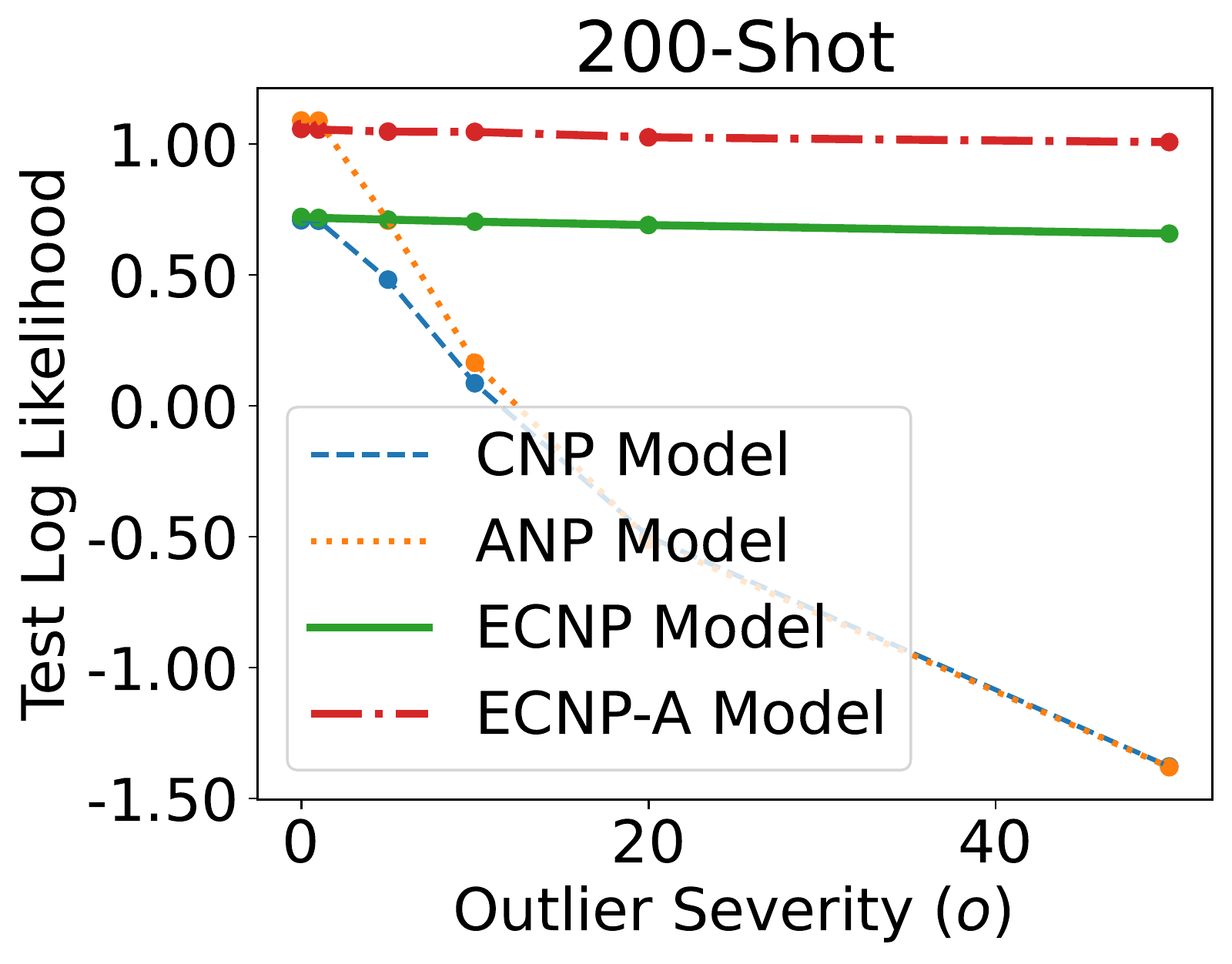}
\caption{Cifar10-LL}
\end{subfigure}
\begin{subfigure}{0.23\textwidth}
\centering
\includegraphics[width=0.9\linewidth]{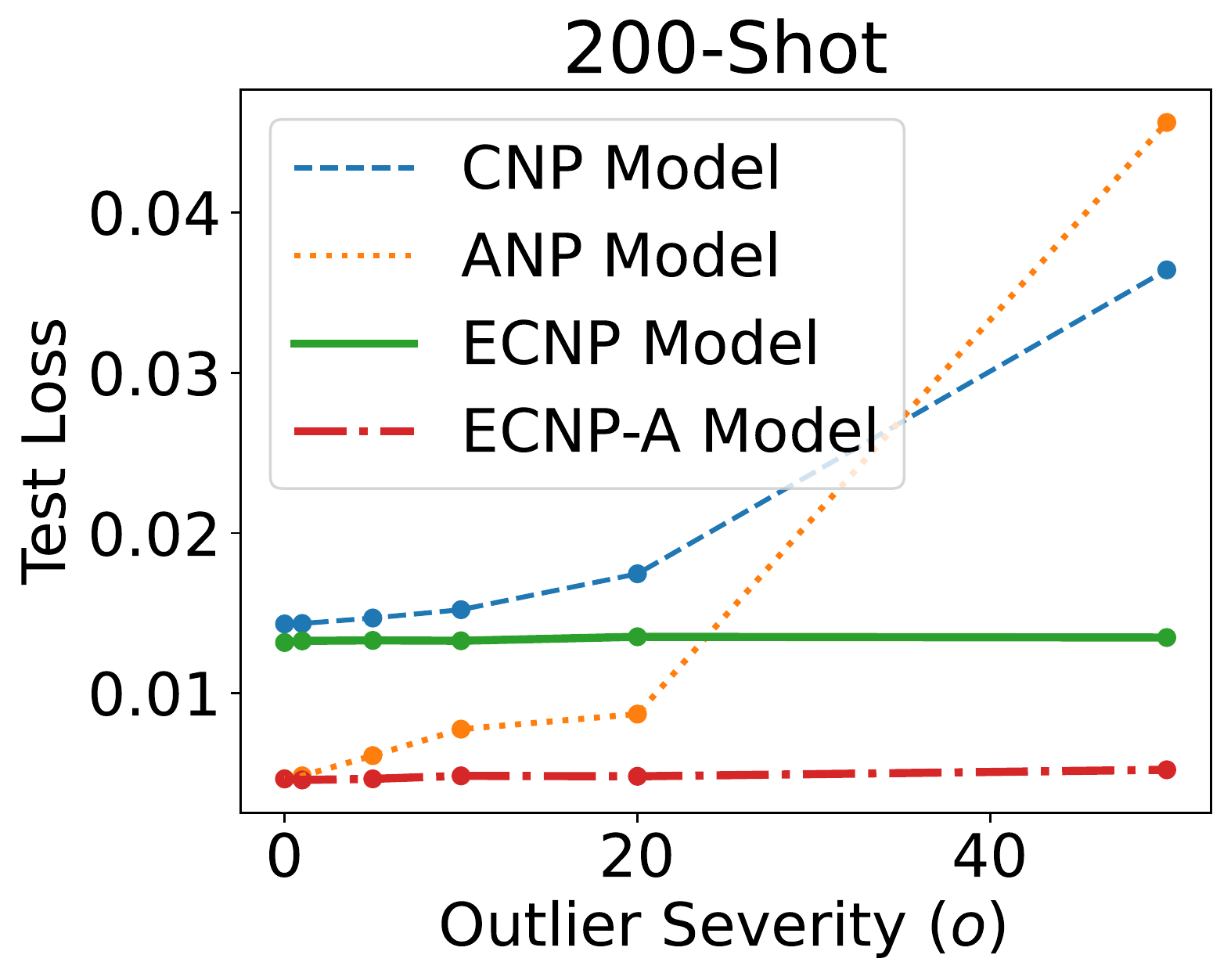}
\caption{CelebA-MSE}
\end{subfigure}
\begin{subfigure}{0.23\textwidth}
\centering
\includegraphics[width=0.9\linewidth]{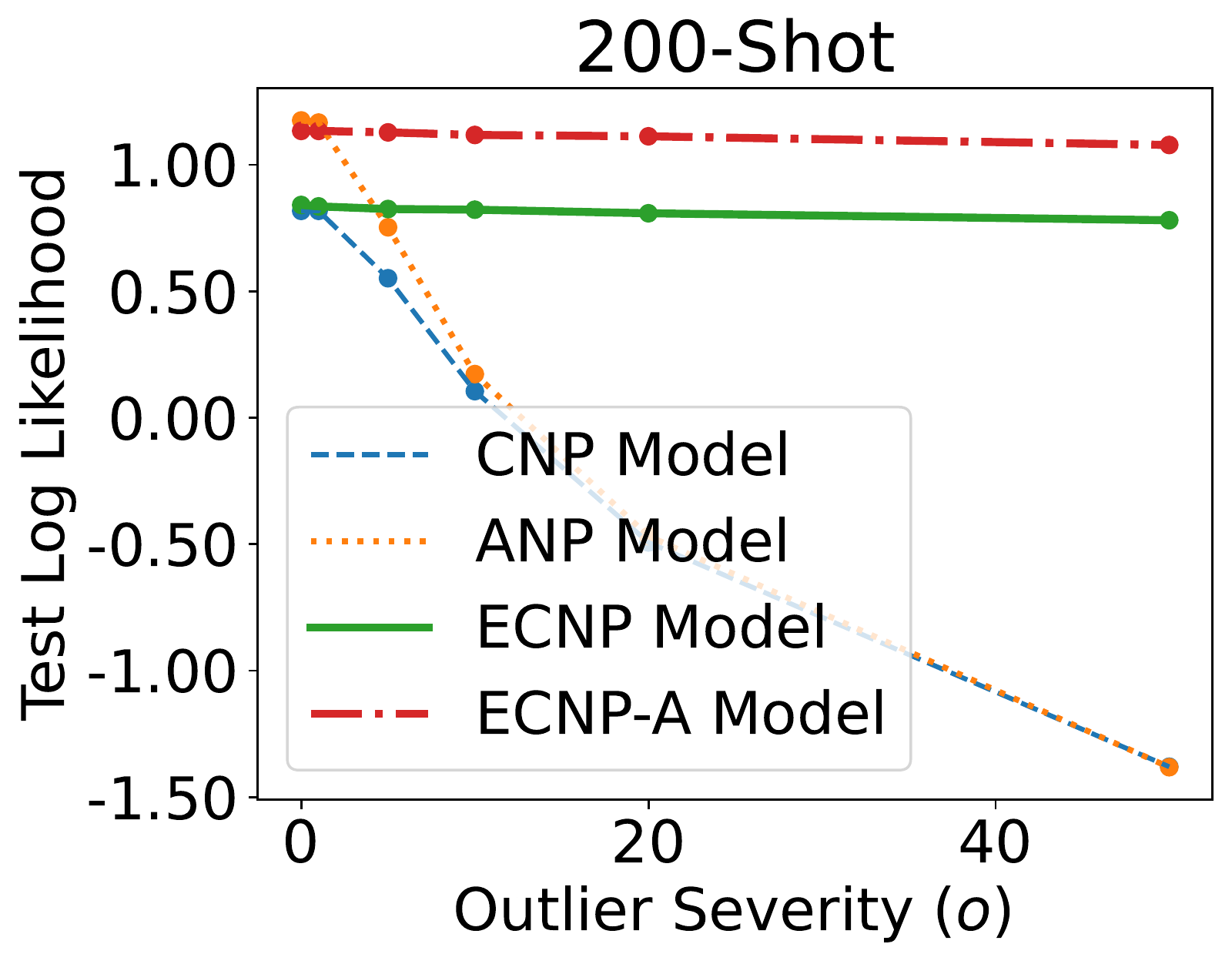}
\caption{CelebA-LL}
\end{subfigure}
\caption{Impact of outlier to NP based models for different 200-shot image completion tasks.}
\vspace{-1mm}
\label{fig:appendixOutlierAll200shot}
\end{figure*}

We carry out experiments with Cifar10 and CelebA datasets across 50-shot and 200-shot settings (Figure \ref{fig:appendixOutlierAll50shot} and Figure \ref{fig:appendixOutlierAll200shot}) to empirically validate above theoretical claims. We consider the CNP and ANP models as the baselines and compare with their evidential extensions: ECNP and ECNP-A. For all the evidential models, we set both $\lambda_1$ and $\lambda_2$ to 0.1. In absence of any outliers in the training tasks, our evidential model shows comparable to marginally better performance. As the outlier in training tasks become more extreme, the baseline models start to break down and their performance degrades significantly, In comparison, our model continues to remain robust to outliers for different severity level across all datasets and settings. Experiments clearly demonstrate superiority of our proposed model for outlier robustness. 

\subsection{Theorem 2}
\textit{The conditional neural process is one instance of an evidential neural process when two of the evidential hyperparameters meet the following conditions: (i) $\alpha_t \rightarrow \infty$; (ii) $\alpha_t v_t = \text{const}$. }
\label{app:proof2}
\begin{proof}
Let $\frac{v_t\alpha_t}{\beta_t(1+v_t)} = k$. Now, consider one instance of our model when $2\alpha_t \rightarrow \infty$ and $v_t\alpha_t$ is a constant. The condition $v_t\alpha_t = const$ can easily be satisfied \eg by setting $v_t = \frac{1}{\alpha_t}$. In this case, $k \rightarrow \frac{1}{\beta}$ and we get
\begin{align}
p(y_t|x_t,\mathbf{p}_t) &\propto \Big(1 + \frac{k (y_t - \gamma_t)^2}{ 2\alpha_t}\Big)^{- ( \alpha_t + 1/2)} \\
&= e^{-\frac{k(y_t - \gamma_t)^2}{2} + O(\frac{1}{2\alpha_t})}
\end{align}
The predictive distribution is an exponential quadratic function w.r.t. $y_t$, which gives rise to a Gaussian $y_t \sim \mathcal{N}(y_t|\gamma_t, \beta_t^{-1})$. This matches the predictive distribution output by a CNP model. 
\end{proof}

\section{Posterior Analysis of the Hierarchical Bayesian Model for Evidence Quantification\label{postanalysis}}

In the ECNP, we assume a hierarchical Bayesian model in which each observation $y_n$ is a sample from a Gaussian with unknown mean and unknown variance, with a higher-order Normal-Inverse-Gamma prior $\text{NIG}(\mu, \sigma^2|\mathbf{p})$ over the Gaussian likelihood function
\begin{align}
        &y_n \sim \mathcal{N} (\mu, \sigma^2)\\
    & \mu \sim \mathcal{N}(\mu|\gamma_p, \sigma^2 v_p^{-1}), \quad \quad \sigma^{2} \sim \Gamma^{-1}(\sigma^2|\alpha_p, \beta_p)\\
    &\text{NIG}(\mu, \sigma^2|\mathbf{p}) = \mathcal{N}(\mu|\gamma_p,\frac{\sigma^2}{v_p})
    \Gamma^{-1}(\sigma^2|\alpha_p, \beta_p)
\end{align}
where $\mathbf{p} = (\gamma_p, v_p, \alpha_p, \beta_p)$ represents the parameters of the NIG distribution and $\Gamma^{-1}$ represents the inverse-gamma distribution. The evidential hyperparameters are governed by data observations where each new observation contributes to the model's predictive behavior.

To illustrate the model behavior, let us assume that we observe $N$ i.i.d. data points $Y_N = \{y_1, y_2,...y_N \}$ and study the impact of the observations on our hierarchical evidential model. Due to the conjugacy between the prior and the likelihood, the posterior is also an NIG distribution:
\begin{align}
    p(\mu, \sigma^2|Y_N) \propto \text{NIG}(\mu, \sigma^2|\mathbf{p}) \prod_{n=1}^N \mathcal{N} (y_n|\mu, \sigma^2)
\end{align}
This posterior factorizes as $ p(\mu, \sigma^2|Y_N) = p(\mu|\sigma^2, Y_N)p(\sigma^2|Y_N)$. The conditional posterior of the mean  ($p(\mu|\sigma^2, Y_N)$) is 
\begin{align}
    p(\mu | \sigma^2, Y_N) &= \mathcal{N}(\mu|\gamma_N, \sigma^2 v_N^{-1} )
\end{align}
where the parameters of the conditional distribution are given by
\begin{align}
    v_N &= v_p + N \\
    \gamma_N &= \frac{v_p}{v_N} \gamma_p + \frac{N}{v_N} \bar{y}_N, \quad\bar{y}_N = \frac{1}{N} \sum_{n=1}^N y_n
\end{align}
Here, $v_N$ effectively serves as the evidence for the observations: as we collect more data, $v_N$ increases leading to reduced variance and more confident predictions.
The posterior over the variance ($p( \sigma^2| Y_N)$) is an inverse-gamma distribution of the form
\begin{align}
    &p( \sigma^2| Y_N) = \int p(\mu, \sigma^2|Y_N) \text{d}\mu \\
    &\begin{aligned}
    &\propto_{\sigma^2}\int \prod_{n=1}^N \mathcal{N} (y_n|\mu, \sigma^2) \mathcal{N}(\mu|\gamma_p, \frac{\sigma^2}{v_p} )\Gamma^{-1}(\sigma^2|\alpha_p, \beta_p) \text{d}\mu\\
    \end{aligned}\\
    &=\Gamma^{-1}(\sigma^2|\alpha_N, \beta_N )
\end{align}
where the parameters of the posterior are 
\begin{align}
    \alpha_N &= \alpha_p + \frac{N}{2} \\
    \beta_N &= \beta_p + \frac{1}{2}\sum_{n=1}^N (y_n - \bar{y}_N )^2 + \frac{Nv_p}{2(v_p +N)}(\bar{y}_N - \gamma_p)^2 
\end{align}
The parameters $\alpha_N$ and $\beta_N$ contribute to the model's confidence (\ie the model evidence) indirectly through the higher-order Inverse gamma (IG) distribution. The expected value of $\sigma^2$ is $\frac{\beta_N}{\alpha_N-1}$. When $\alpha_N$ is high and $\beta_N$ is small, the IG samples $\sigma^2 \sim \Gamma^{-1}(\alpha_N, \beta_N)$ are close to zero indicating low variance and high confidence in prediction. Conversely, when $\alpha_N$ is small and $\beta_N$ increases, the variance $\sigma^2$ increases indicating low confidence in the prediction. Based on the above analysis, we define the evidence for the prediction ($\mathcal{E})$ in the evidential hierarchical model as
\begin{align}
        \mathcal{E}_N &= v_N + \alpha_N + \frac{1}{\beta_N}  
\end{align}  
In this hierarchical Bayesian model, the $N$ training data observations interact with the prior distribution $\text{NIG}(\mu, \sigma^2)$ to output the hyperparameters for the NIG posterior. Equivalently, in the proposed evidential conditional neural processes as shown in Figure \ref{fig:ecnpModel} right, the context set $\mathcal{C}$ interacts with the meta knowledge in the meta-learning model to output the posterior NIG parameters $\mathbf{p}_t= (\gamma_t, v_t, \alpha_t, \beta_t)$ for a target input $x_t$. The parameter $\gamma_t$ corresponds to the prediction, and the remaining NIG parameters work together to quantify the aleatoric uncertainty, the epistemic uncertainty, and the evidence for the prediction.

\subsection{Related works on Evidential Deep Learning} In Evidential Deep Learning models, ideas from Subjective Logic \cite{josang2016subjective} are used to equip Deep Learning models with accurate uncertainty quantification capabilities. Evidential Deep learning has been extended to both classification and regression problems. 
EDL\cite{sensoy2018evidential} introduces higher-order evidential Dirichlet prior for the multinomial likelihood in classification problems that enables the deterministic neural network model to capture different uncertainty characteristics. Units-ML \cite{pandey2022multidimensional} extends EDL for few-shot classification.
ETP \cite{kandemir2021evidential}, an improvement on EDL, develops an uncertainty-aware classification model by integrating parametric Bayesian and evidential Bayesian model into a complete Bayesian model that addresses the issue of total calibration in classification. 
DER~\cite{NEURIPS2020_aab08546} extends evidential learning to regression problems by introducing a NIG prior for the Gaussian likelihood that leads to effective aleatoric-epistemic uncertainty quantification.
NatPN~\cite{charpentier2022natural} develops an unified evidential deep learning model for both classification and regression by introducing exponential family distributions as effective prior distributions. Compared to the above evidential works, our work can be seen as a novel extension of DER work~\cite{NEURIPS2020_aab08546} to the meta-learning setting that enables fine-grained uncertainty quantification in the few-shot regression tasks.

\section{Details of Datasets and Experimental Settings }\label{app:setting}
In this work, we consider two synthetic regression experiments (sinusoidal regression and GP) and three real-world benchmark datasets for image completion experiments: MNIST, Cifar10, and CelebA. For the synthetic regression experiments, we consider $K$-shot tasks with additional $u$ samples (i.e. effectively $K + u$ samples where $u \sim U(3,K)$, represents sampling from a uniform distribution in range $(3,K)$) in the target set of training tasks and 400 samples in the target set of test tasks. For image completion experiments, we consider $K$ random pixel positions ($K  = 50/200$) in the context set and all the pixel positions in the target set. The details of the image datasets are presented in Table \ref{tab:1dataset_details}.
\begin{table}[htpb]
    \centering
        \caption{Dataset Details}
    \label{tab:1dataset_details}
\begin{tabular}{||c| c c c||} 
 \hline
 Characteristic&MNIST&CelebA&Cifar10\\ [0.5ex] 
 \hline\hline
 Image Size&28$\times$28&32*32&32$\times$32\\
 \hline
 Channels ($ch$)&1&3&3\\
 \hline
 Training Images&60,000&162,770&50,000\\
  \hline
Test Images&10,000&19,962&10,000\\
\hline
\end{tabular}
\vspace{-5mm}
\end{table}

\subsection{Details of Uncertainty Metrics}\label{app:uncertainty Metrics}
In this work, we use Mean Squared Error (MSE) and Log Likelihood (LL) to compare the generalization performance; and consider Inclusion@K and Uncertainty-Increase \cite{groverprobing} to evaluate the uncertainty estimates of the models. 

\noindent\vspace{2mm}\textbf{Inclusion@K}(I(k)) is defined as:
\begin{align}
    I(k) = \mathbb{E}_{x\sim \text{Uniform}(\mathcal{X})}[\mathbb{I} (|f(x)-m(x, \mathcal{S})|<ks(x,\mathcal{S})]
\end{align}
where the model outputs the  prediction $m(x, \mathcal{S})$ and uncertainty $s(x, \mathcal{S})$ for the task $\mathcal{T}$ defined by the true function $f(.)$. The task has support set $\mathcal{S}$, query set $\mathcal{Q}$, and $\mathcal{X}$ represents the entire set of inputs in the task. To compute inclusion, we consider the variance of the predictive distribution as the uncertainty $s(x, \mathcal{S})$. Moreover, we consider query set inputs for $\mathcal{X}$.

\noindent\vspace{2mm}\textbf{Uncertainty-Increase}($UI$)  is defined as:

\begin{align}
    UI = \frac{\sum_{x \sim \text{Uniform}(\mathcal{X})} \mathbb{I}( s(x, \mathcal{S}) - NN(x,\mathcal{S}) )}{|\mathcal{X}|}
\end{align}
where $NN(x, \mathcal{S})$ represents the uncertainty of the datapoint in support set $S$ that is closest to the datapoint $x$. In our experiments, we consider query set inputs for $\mathcal{X}$ and $|\mathcal{X}|$ represents the number of datapoints in the query set.

In our model, the proposed kernel based regularization encourages the model to correct its epistemic uncertainty that is expected to lead to accurate predictive uncertainty and improved performance on Uncertainty increase metric. Similarly, the proposed evidence regularization term encourages the model to have low confidence for wrong predictions. Such regularization are expected to lead improved uncertainty characteristics in our model. 

\subsection{Model Architectures and Setup}
Our ECNP model can capture both the aleatoric and epistemic uncertainty in the few-shot tasks in a single forward pass without the need of sampling. In our experiments, for the context set encoder, we use a 4 layer neural network with 128 dimensional hidden layers that leads to 128 dimensional features. For the decoder, we use the 3 layer neural network with 130 dimensional input (129 dimensional input for function regression experiments), and 128 dimensional hidden layers across all experiments. In all the models with attention mechanism, we use multihead cross-attention with 8 heads in the encoder similar to \cite{kim2018attentive}. To obtain the evidential hyperparameters for ECNP, we transform the output representation using a 2 layer neural network with a 64 dimensional hidden layer. We apply ReLU activation function in the intermediate layers and apply the softplus activation on the final layer to obtain the evidential parameters. In the NP model with latent variable, we sample 5 instances from the latent variable to train (\ie ELBO estimation) and evaluate the model similar to \cite{garnelo2018neural}. In the NP model, the reparameterization trick of variational-auto-encoders \cite{kingma2013auto} is used for the Gaussian distribution. Unless specified, for the CifarFS and CelebA  evidential neural process experiments, we set $\lambda_1 = 0.01$, $\lambda_2 = 0.01$, and for all remaining experiments, we set $\lambda_1 = 0.1$, and $\lambda_2 = 0.1$. For the NIG hyperparameters, we set $\beta_t = f_\theta(.) + 0.2$, and upper bound the $\alpha_t$ and $v_t$ to $20$. Models are trained with the learning rate of $0.001$ and Adam optimizer. For the quantitative results (Table \ref{tab:5shotRegressionBig1}, \ref{tab:overallResultsGen50Shot}, \ref{tab:uncInclusionRepTable}), we average the results over 5 independent runs of the model and report the mean and standard deviation. The experiments use Pytorch, and are carried out on a 8GB GeForce RTX 2070 SUPER GPU-enabled PC and on a cluster with 8GB P4 Nvidia GPU using resources at \cite{https://doi.org/10.34788/0s3g-qd15}. The codes for both the baselines and evidential models are available at https://anonymous.4open.science/r/ENP-DB67/README.md.
\subsection{Complexity Discussion}
Compared to other meta-learning works such as MAML \cite{finn2017model}, the proposed model has rapid inference capabilities, and computationally cheaper training. During training, MAML-based models formulate meta-learning as a bilevel optimization problem that introduces computationally expensive Hessian-gradient products for global model parameter update. Specifically, when training on one task, for each inner loop update over the support set, one additional hessian term needs to be computed for the global parameter update that leads to multiple forward-backward passes over the network. MAML’s bayesian extensions for uncertainty quantification further increase computational cost. For instance, BayesianMAML introduces expensive ensembling of MAML models for uncertainty quantification. In contrast, CNP/ECNP training does not involve any bi-level optimization/hessian gradient products, and learning from a task only involves one forward pass and one backward pass, making it computationally cheap.

Also, during inference, optimization based models are relatively slower/computationally expensive as they need to update the model with the support set information over K gradient steps, whereas CNP/ECNP inference only requires a single forward pass through the network, a highly desirable characteristic in meta-learning algorithms.
A similar model, VERSA \cite{gordon2018meta}, is also computationally cheap to train, achieves rapid inference for prediction, and has uncertainty quantification capabilities. However, to quantify uncertainty, VERSA requires multiple rounds of sampling from the posterior distribution over the class weights of the linear classifier network. In contrast, ECNP leverages evidential learning for uncertainty quantification which avoids posterior sampling, making it even faster than VERSA.

\newpage
\section{Ablation Study}\label{app:ablation}
\begin{figure*}[h] 
\centering
\begin{subfigure}{0.30\textwidth}
  \centering
  \includegraphics[width=0.9\linewidth]{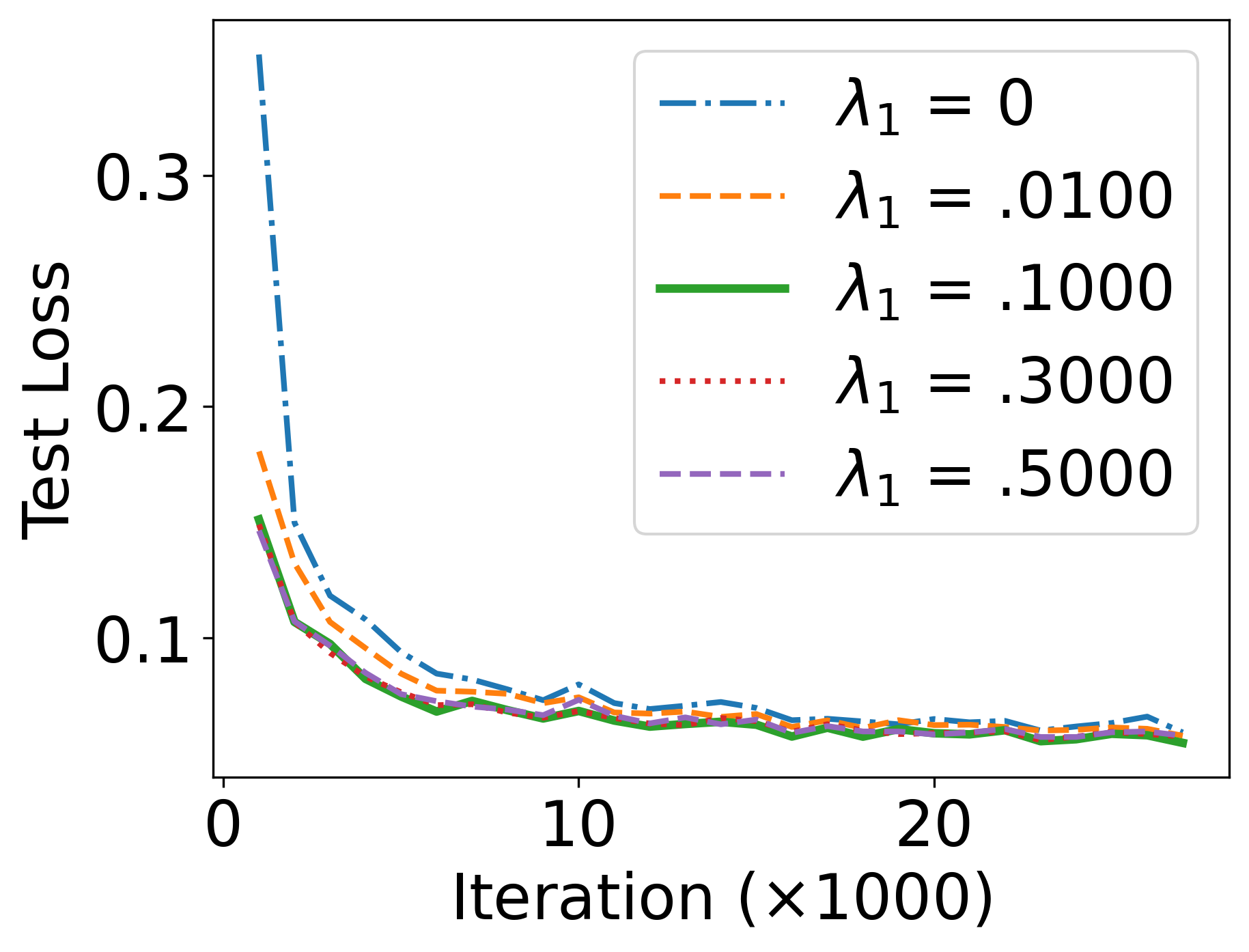}
  \caption{Test Loss Trend}
  \end{subfigure}
 \begin{subfigure}{0.30\textwidth}
  \centering
  \includegraphics[width=0.9\linewidth]{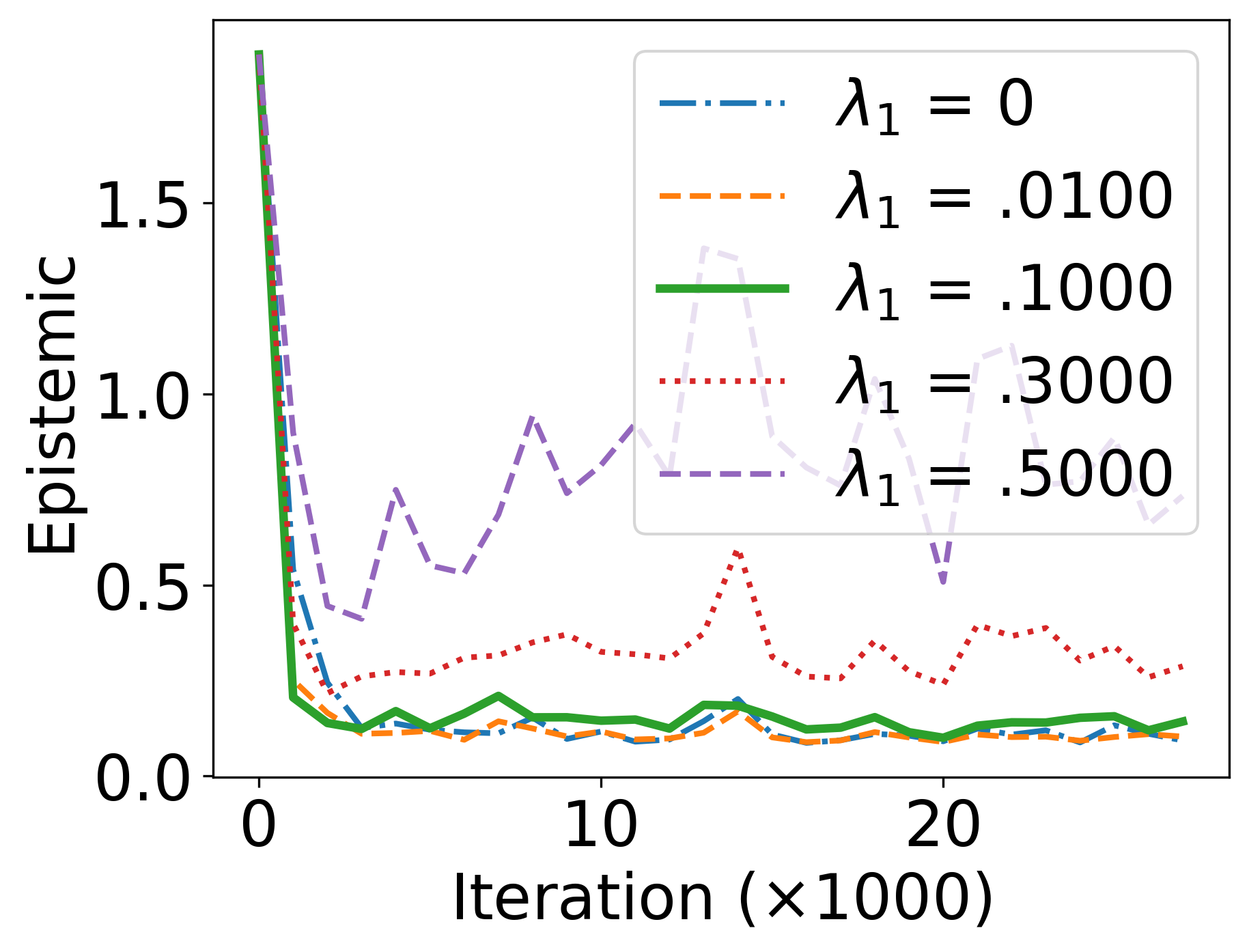}
  \caption{Epistemic Uncertainty Trend}
\end{subfigure}
\begin{subfigure}{0.30\textwidth}
  \centering
  \includegraphics[width=0.9\linewidth]{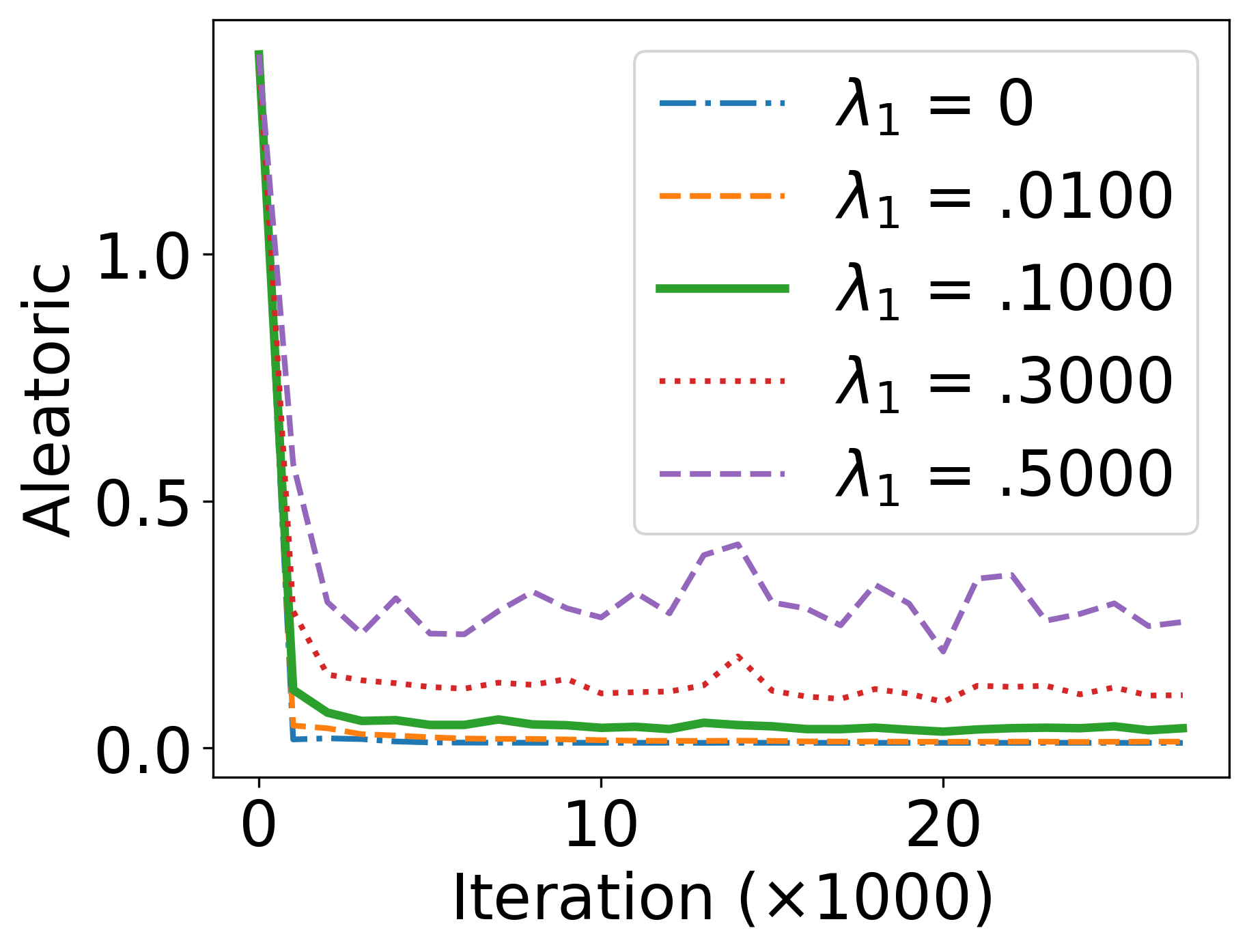}
  \caption{Aleatoric Uncertainty Trend}
\end{subfigure}
\caption{Impact of Regularization parameter $\lambda_1$ in a 10-shot GP function regression task }
\label{fig:lambda1Impact}
\end{figure*}

For the loss function of our ECNP model given by \eqref{eq:cnpLoss}, we introduce two novel regularization terms: incorrect evidence regularizer $L_t^\text{R}$ and epistemic uncertainty regularizer $L_t^\text{KER}$ to guide the model to have accurate uncertainty estimation. The contribution of these terms to model training is controlled by the two parameters $\lambda_1$ and $\lambda_2$, respectively. Here, we study the impact of these hyperparameters in model training and performance. Figure \ref{fig:lambda1Impact} shows the impact of $\lambda_1$ on the test set accuracy, average test set epistemic uncertainty, and aleatoric uncertainty on 10-shot function regression tasks from the GP dataset. The model tends to underestimate the uncertainty values (\ie epistemic and aleatoric uncertainties) when $\lambda_1$ is low whereas a large $\lambda_1$ value causes the model's uncertainty to be large even for accurate model predictions. The model's training is optimal when there is a good balance between minimizing the NLL loss and minimizing the incorrect evidence (\eg $\lambda_1 = 0.1)$. 

\begin{figure*}[h] 
\centering
\begin{subfigure}{0.30\textwidth}
  \centering
  \includegraphics[width=0.9\linewidth]{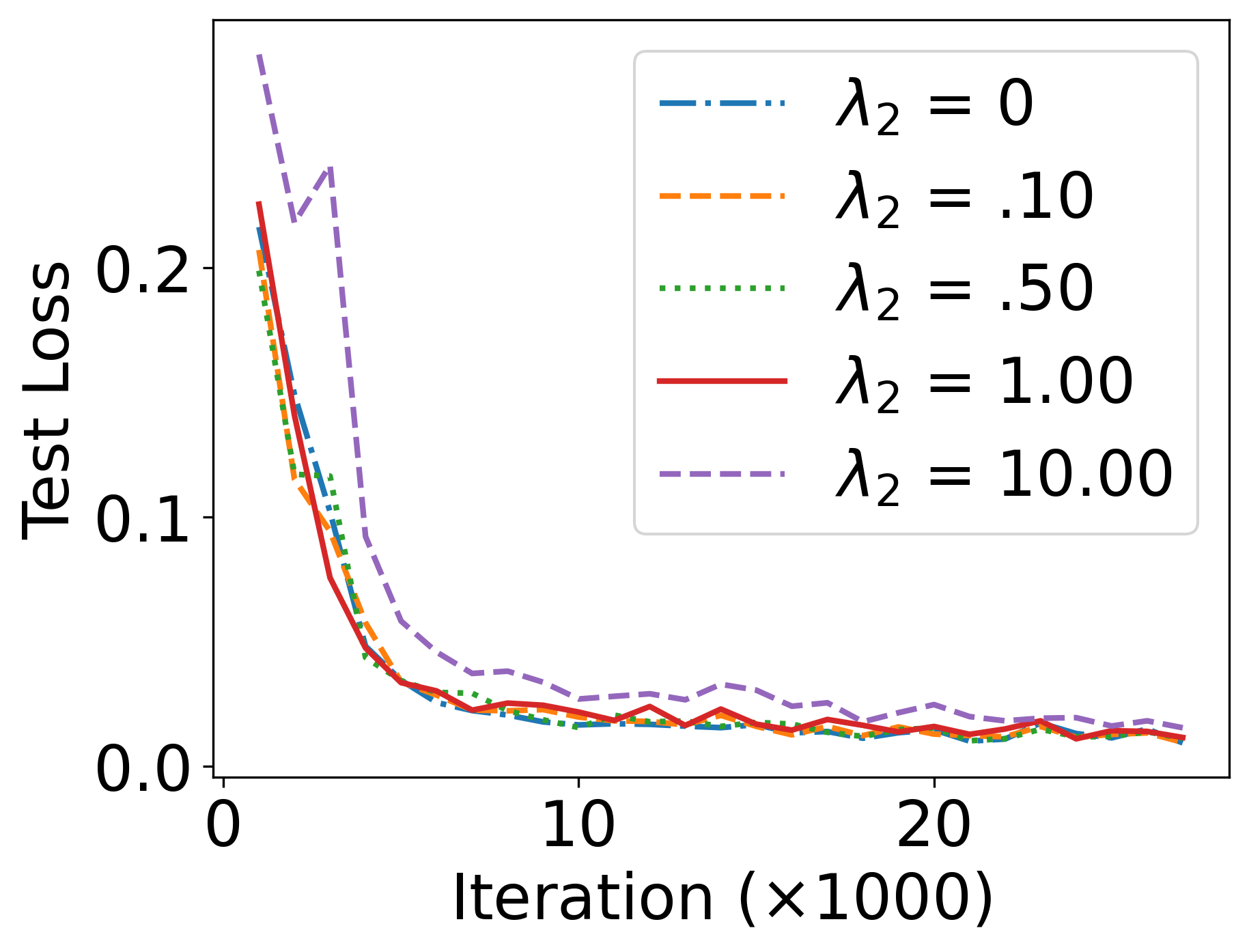}
  \caption{Test Loss Trend}
  \end{subfigure}
 \begin{subfigure}{0.30\textwidth}
  \centering
  \includegraphics[width=0.9\linewidth]{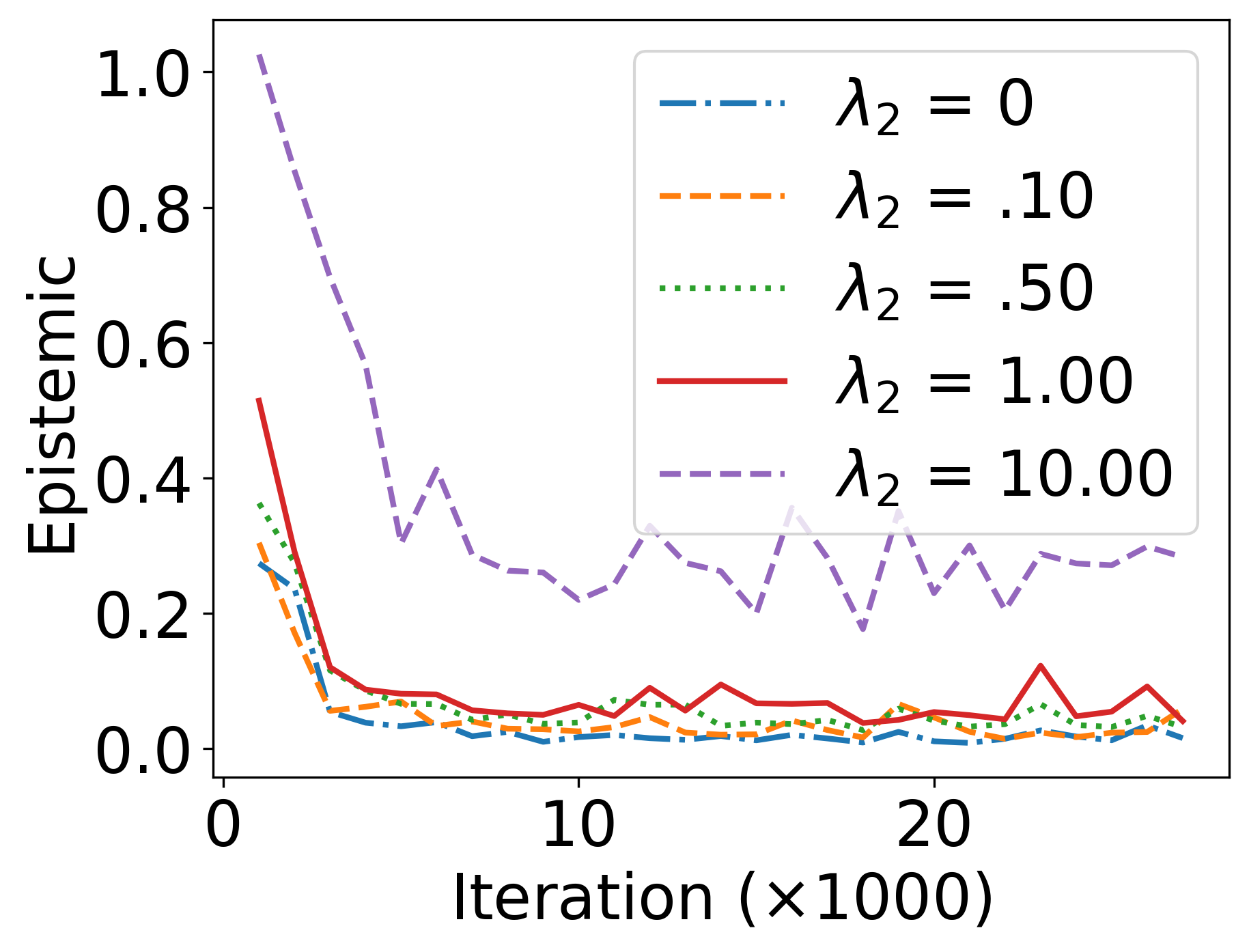}
  \caption{Epistemic Uncertainty Trend}
\end{subfigure}
\begin{subfigure}{0.30\textwidth}
  \centering
  \includegraphics[width=0.9\linewidth]{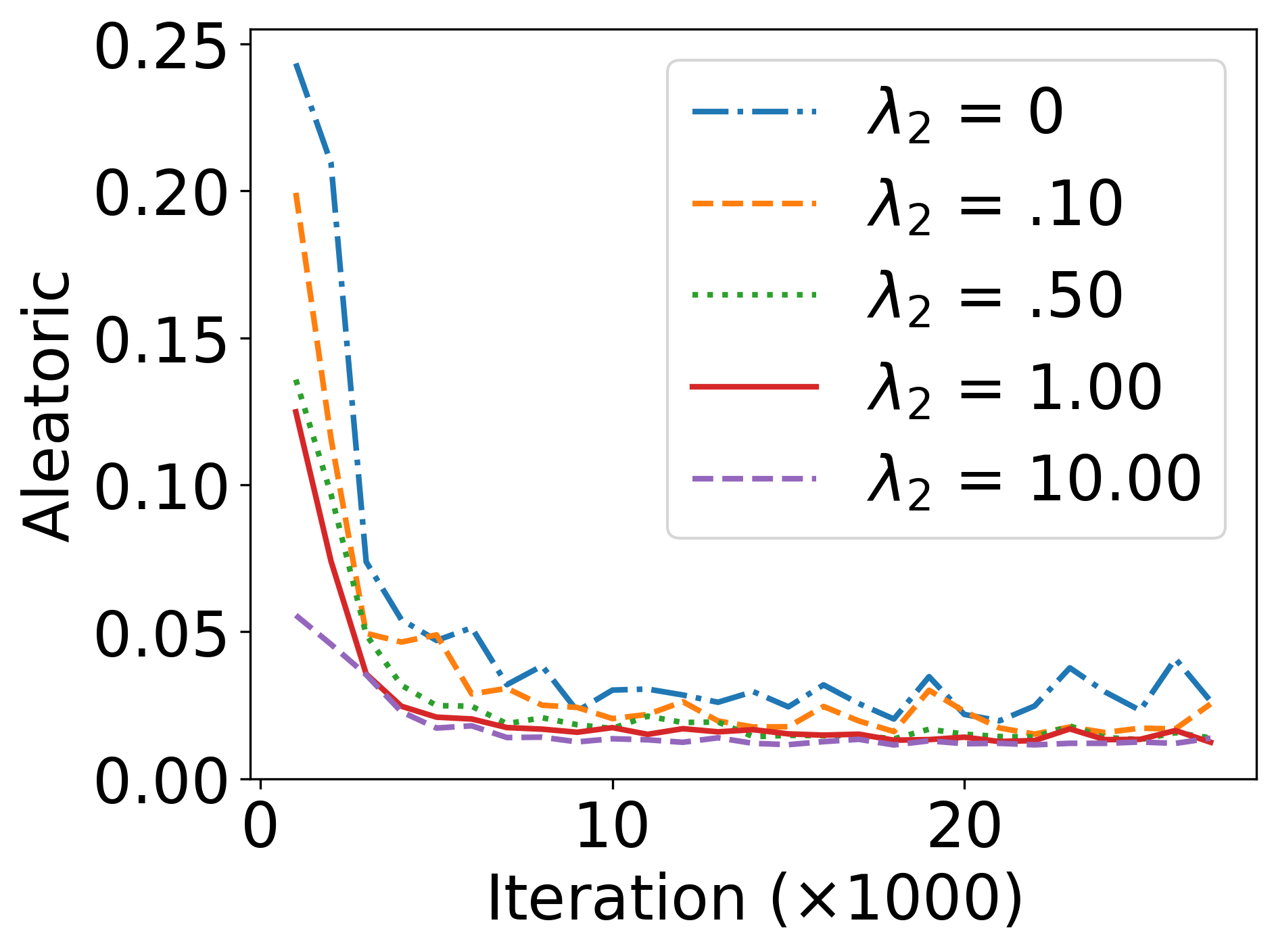}
  \caption{Aleatoric Uncertainty Trend}
\end{subfigure}
\caption{Impact of regularization parameter ($\lambda_2$) in a 5-shot sinusoidal function regression task }
\label{fig:lambda2Impact}
\end{figure*}

\begin{figure*}[h] 
\centering
\begin{subfigure}{0.30\textwidth}
  \centering
  \includegraphics[width=0.9\linewidth]{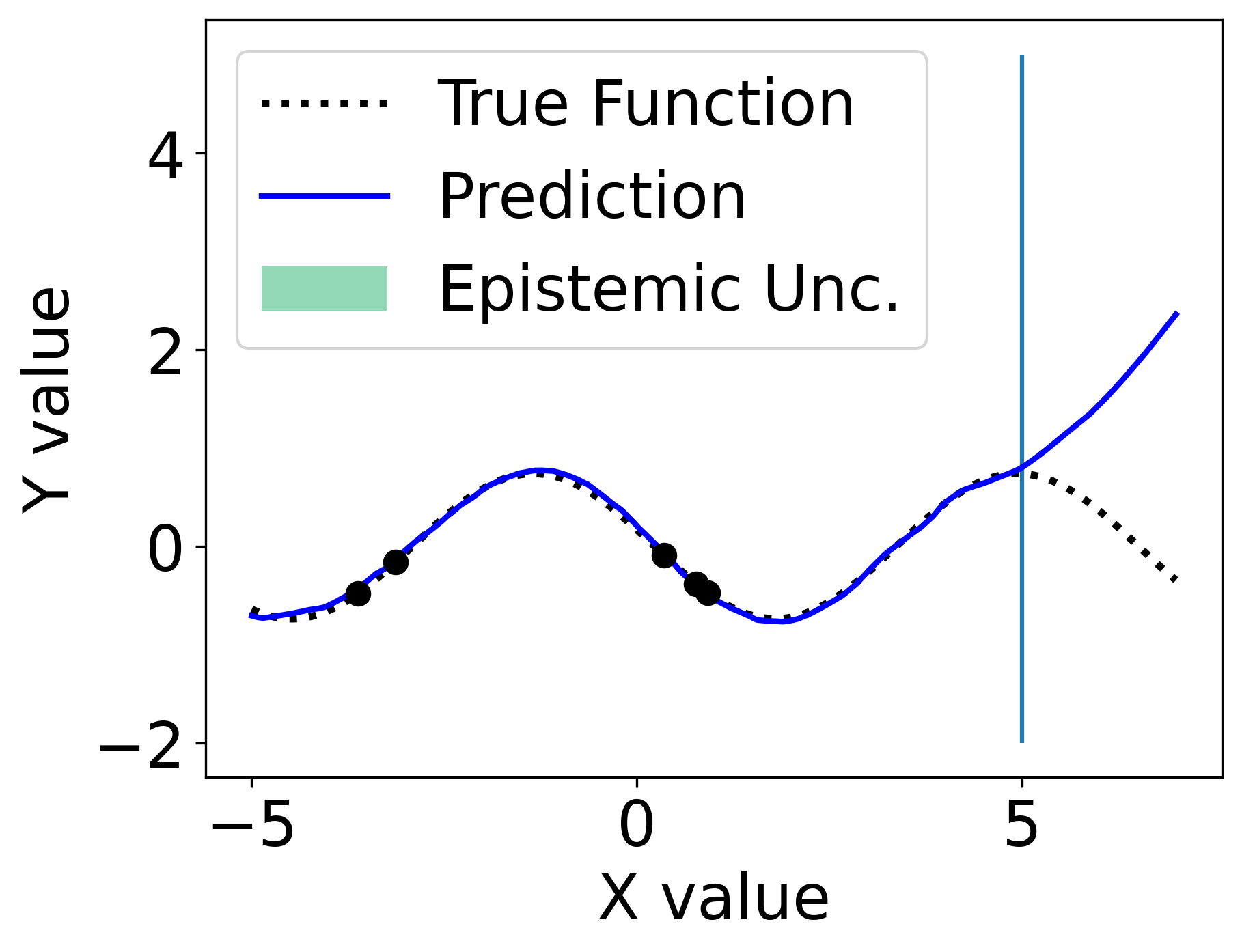}
  \caption{$\lambda_2 = 0$}
  \end{subfigure}
 \begin{subfigure}{0.30\textwidth}
  \centering
  \includegraphics[width=0.9\linewidth]{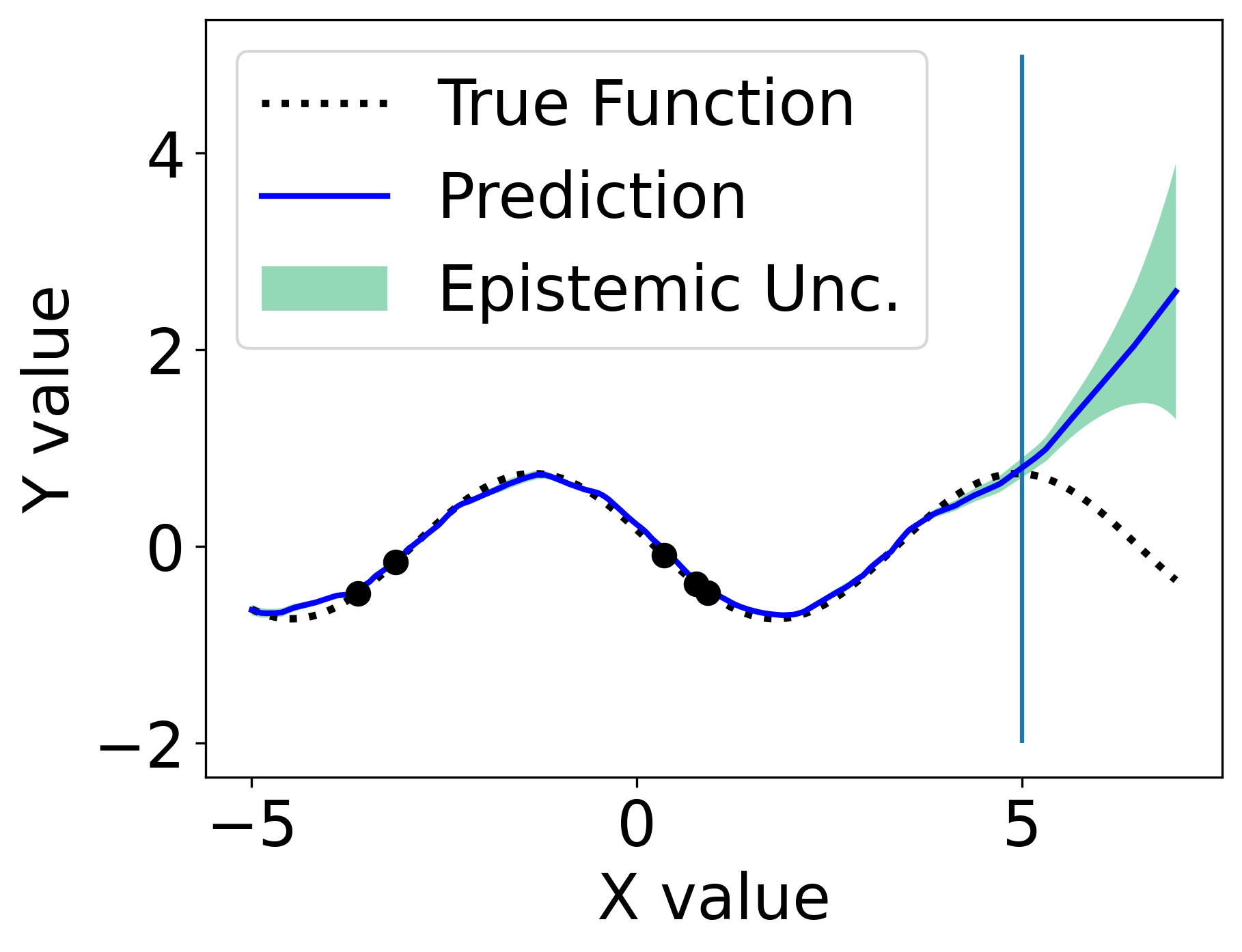}
  \caption{$\lambda_2 = 1.0$}
\end{subfigure}
\begin{subfigure}{0.30\textwidth}
  \centering
  \includegraphics[width=0.9\linewidth]{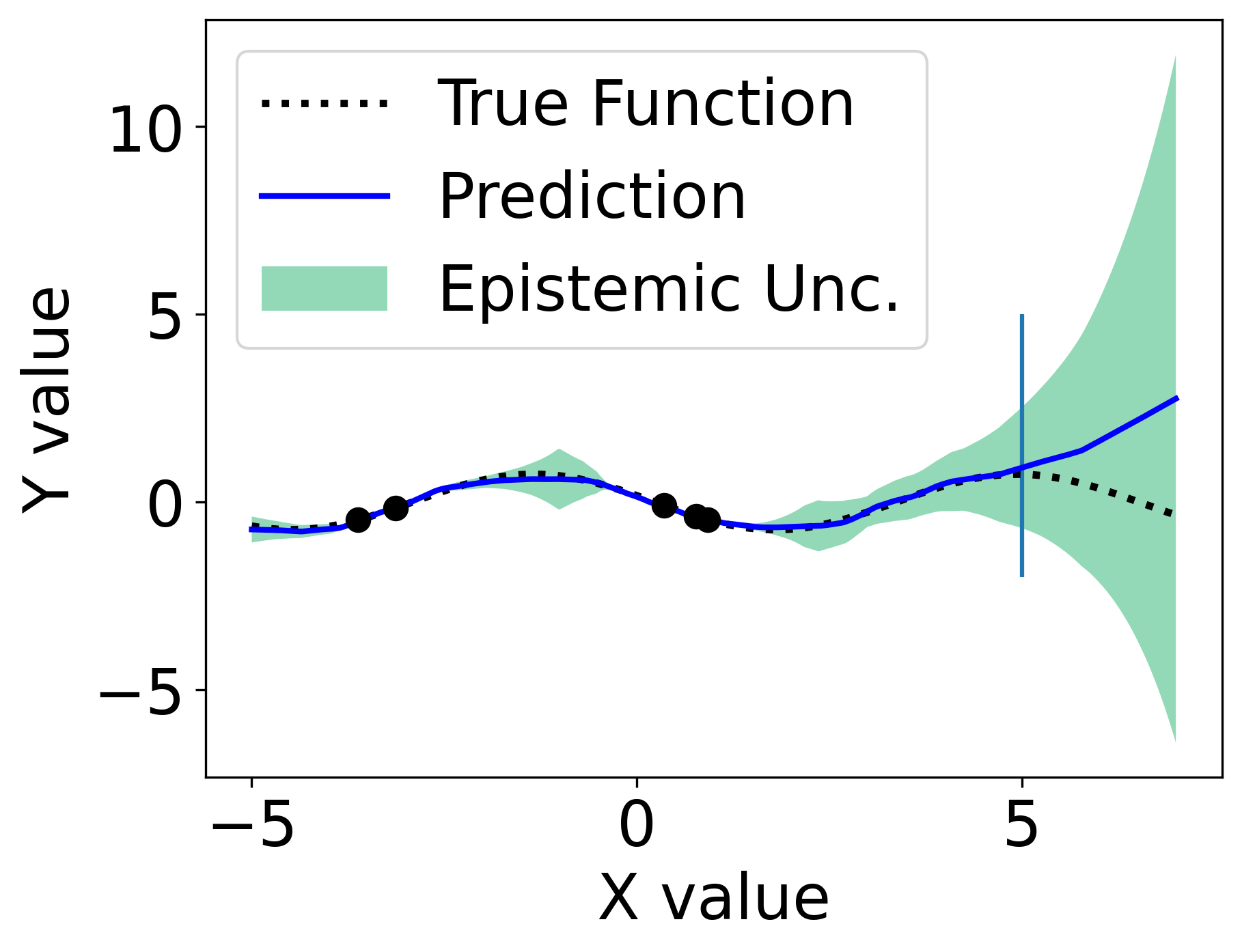}
  \caption{$\lambda_2 = 10.0$}
\end{subfigure}
\caption{Model behavior for different $\lambda_2$ values in a 10-shot sinusoidal function regression task }
\label{fig:lambda2extrapolation}
\vspace{-1mm}
\end{figure*}

The regularization parameter $\lambda_2$ controls the uncertainty estimation in regions far away from the context points. Figure \ref{fig:lambda2Impact} shows the impact of $\lambda_2$ to the model behavior as training progresses in a 5-shot sinusoidal regression problem. A large $\lambda_2$ leads to a sensitive model that outputs very a high epistemic uncertainty as the target data points become far from the context set observations and also hurts the model's generalization performance (\ie average test loss). Conversely, a very low $\lambda_2$ value does not train the model to consider the neighborhood information for uncertainty. The regularization term $\lambda_2$ provides the model with flexibility to consider the neighborhood information (\ie $L_t^\text{KER} = v_t \times D(x_t, \mathcal{C})$ in determining the uncertainty. Next, we carry out extrapolation experiments where the trained model is evaluated outside its training data range (\ie $x > 5.0$). Figure \ref{fig:lambda2extrapolation} visualizes the epistemic uncertainty  for a random task. When the kernel based regularization term is introduced in training (\ie $\lambda_2>0$), the model accurately outputs a high epistemic uncertainty outside the training data range, which is desirable.  

\subsection{Impact of Regularization to Model's Uncertainty Characteristics}
\begin{figure*}[h] 
\centering
\begin{subfigure}{0.31\textwidth}
  \centering
  \includegraphics[width=0.9\linewidth]{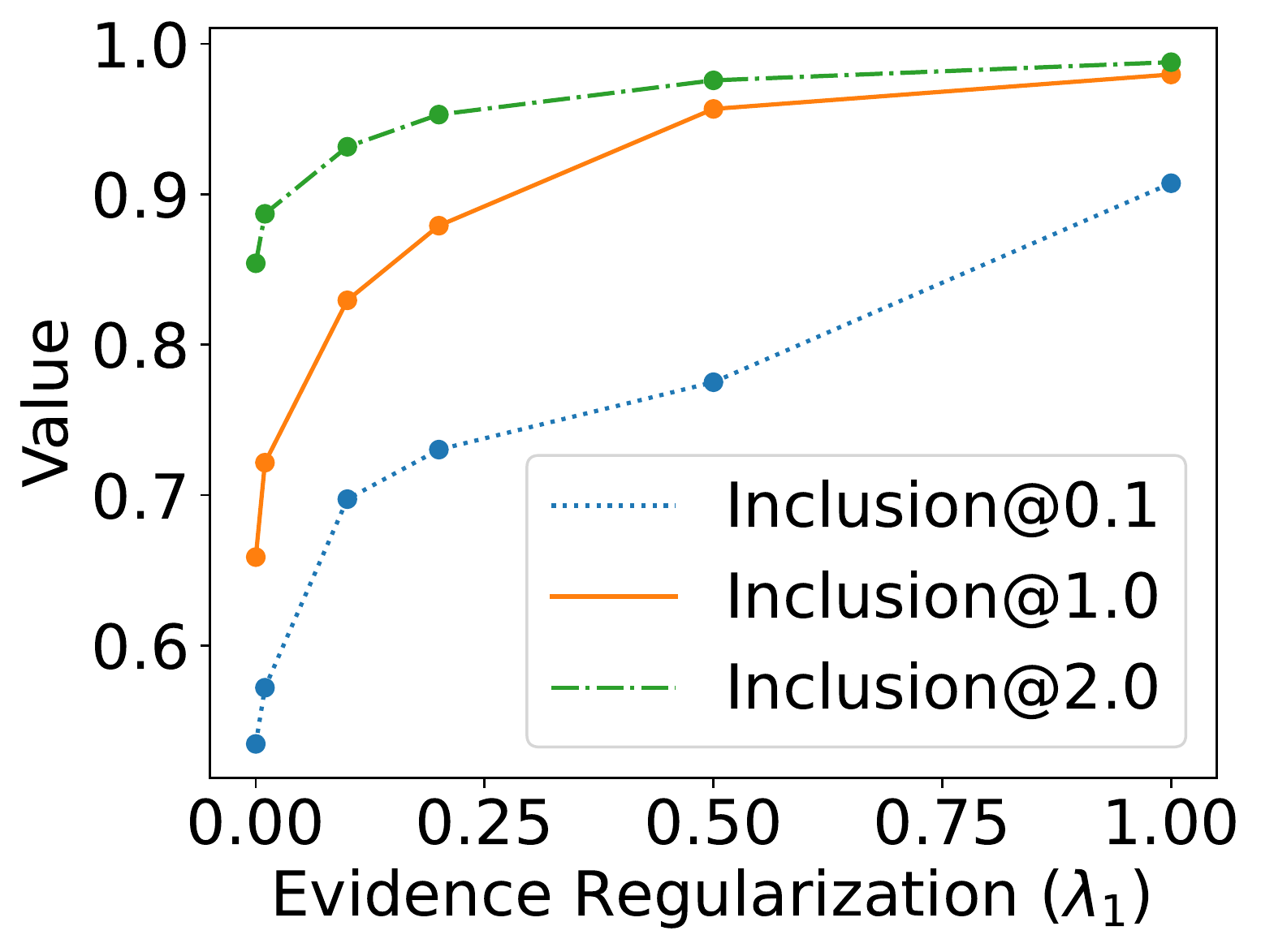}
  \caption{Impact to Inclusion@K}
  \end{subfigure}
\begin{subfigure}{0.31\textwidth}
  \centering
  \includegraphics[width=0.9\linewidth]{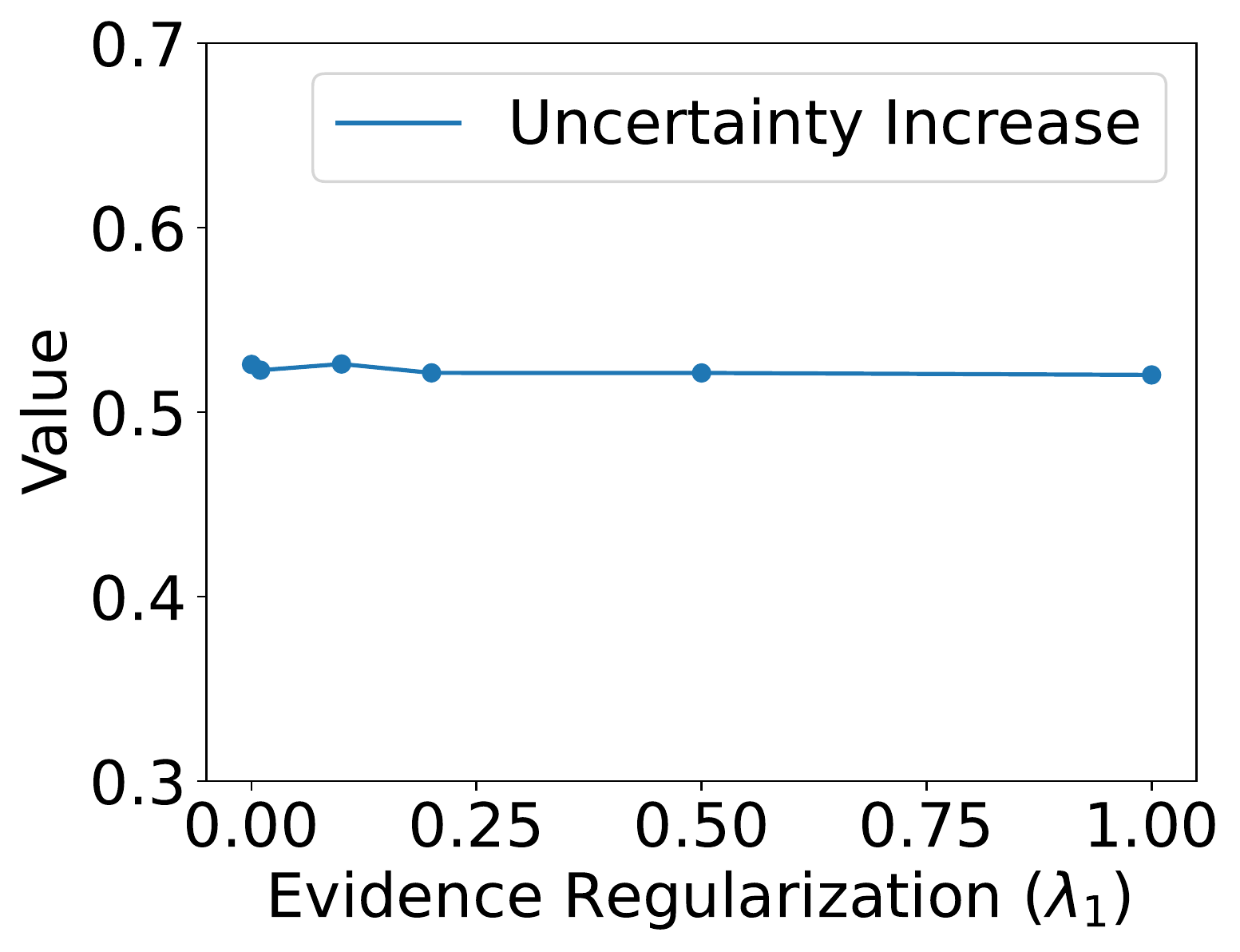}
  \caption{Impact to Uncertainty Increase}
  \end{subfigure}
\begin{subfigure}{0.33\textwidth}
  \centering
  \includegraphics[width=0.9\linewidth]{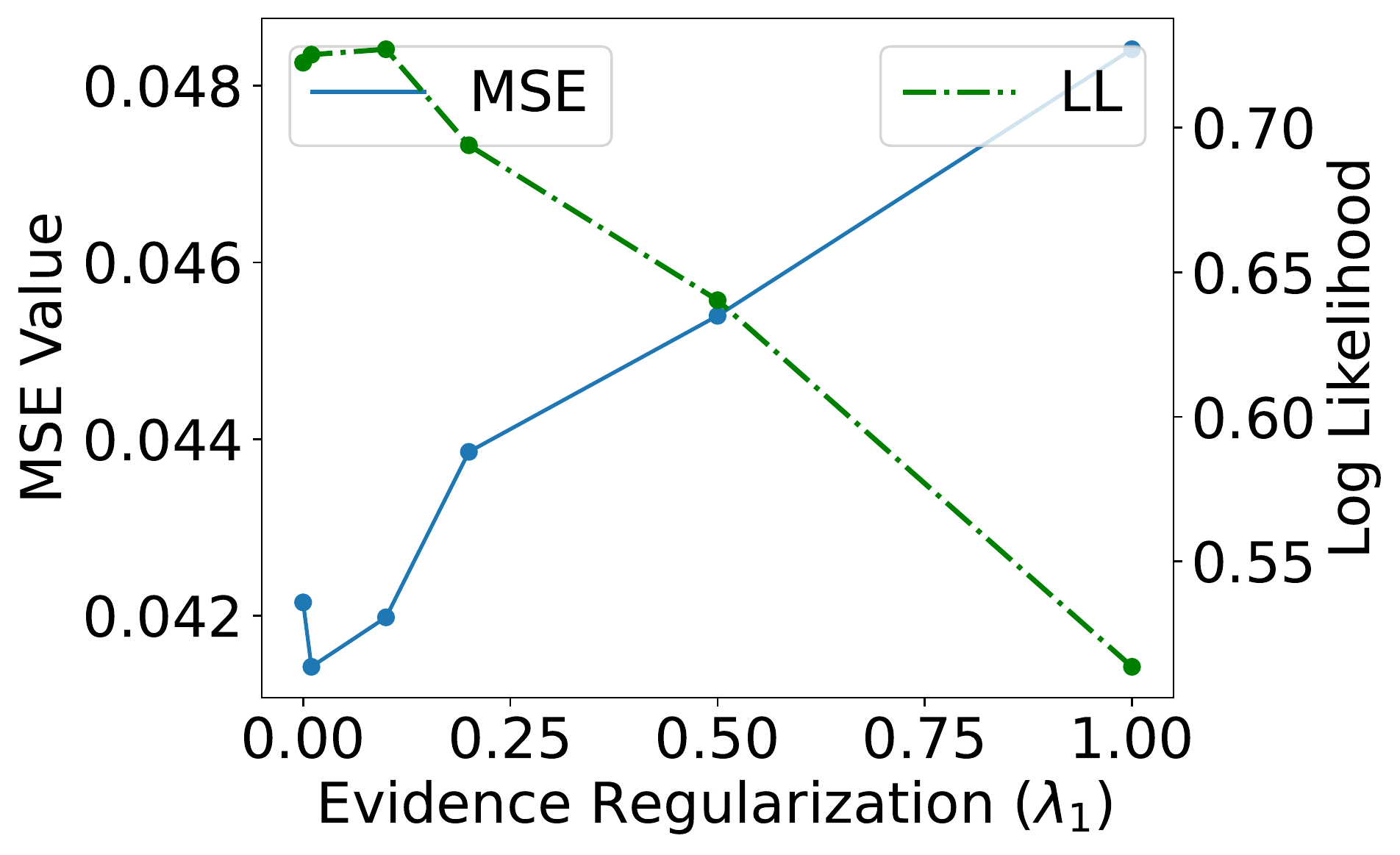}
  \caption{Impact to MSE and LL}
\end{subfigure}
\caption{Impact of Evidence Regularization ($\lambda_1$) to different uncertainty metrics}
\label{fig:ImpactRegUncMetrics}
\end{figure*}

\begin{figure*}[h] 
\centering
\begin{subfigure}{0.31\textwidth}
  \centering
  \includegraphics[width=0.9\linewidth]{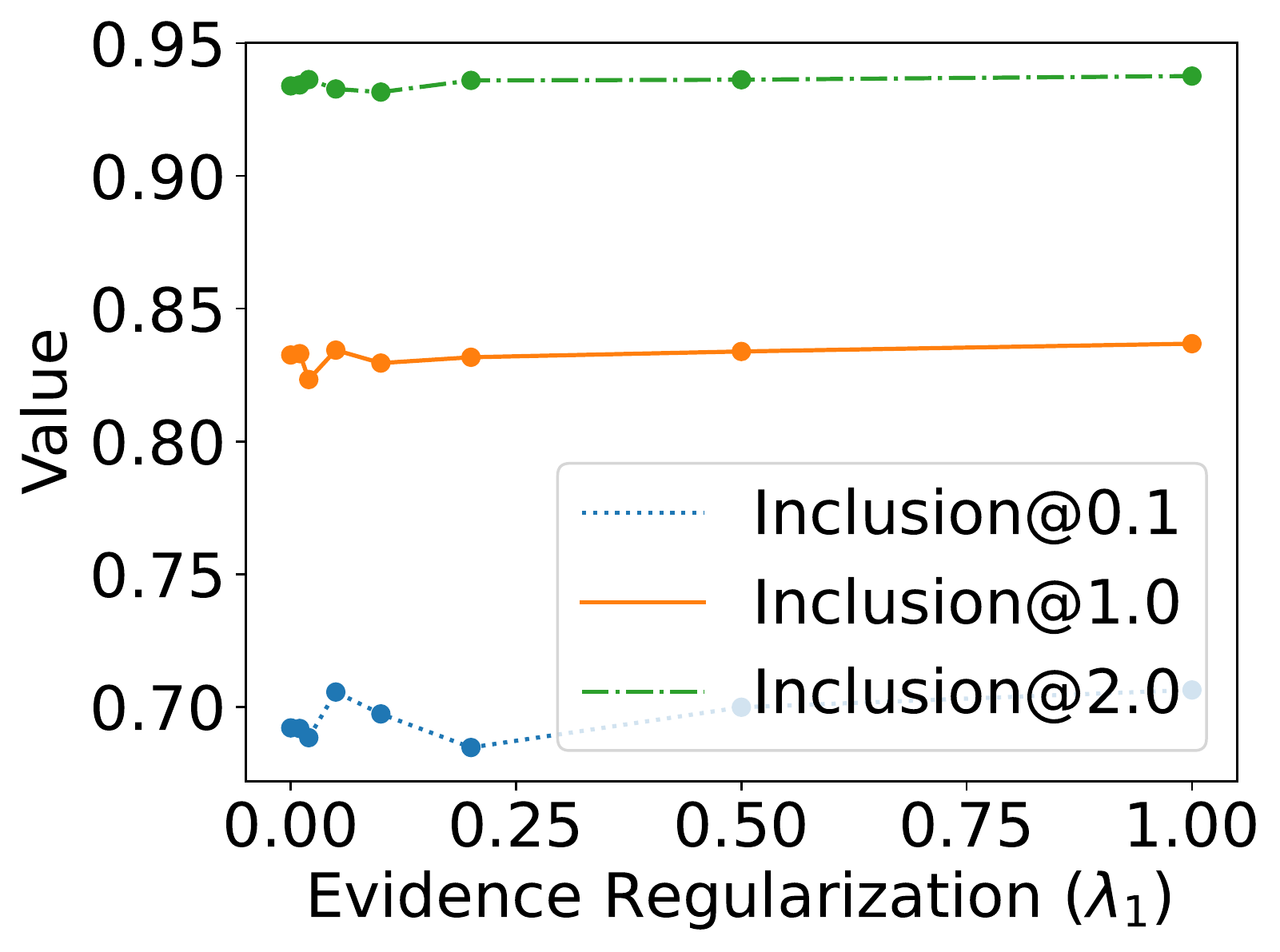}
  \caption{Impact to Inclusion@K}
  \end{subfigure}
\begin{subfigure}{0.31\textwidth}
  \centering
  \includegraphics[width=0.9\linewidth]{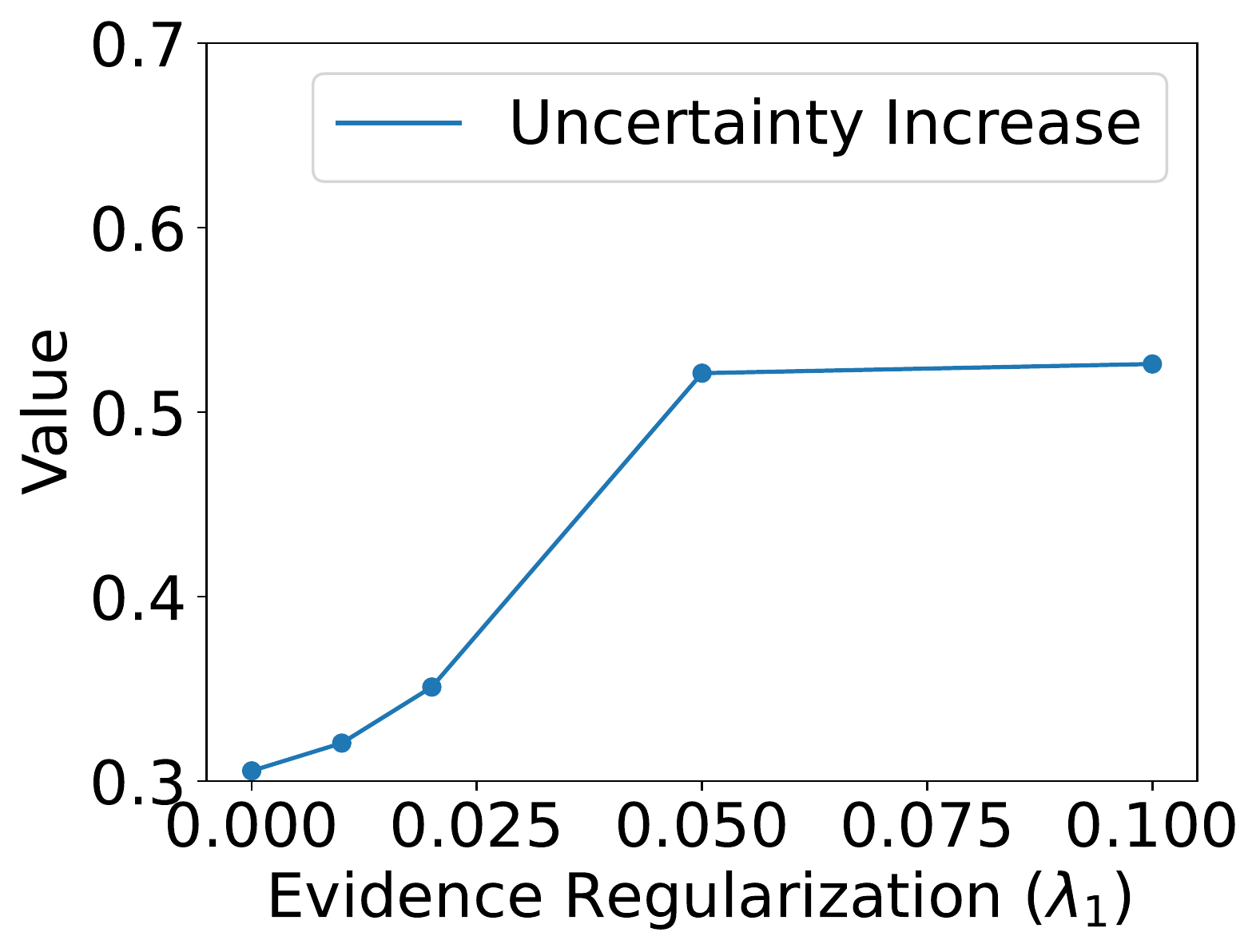}
  \caption{Impact to Uncertainty Increase}
  \end{subfigure}
\begin{subfigure}{0.33\textwidth}
  \centering
  \includegraphics[width=0.9\linewidth]{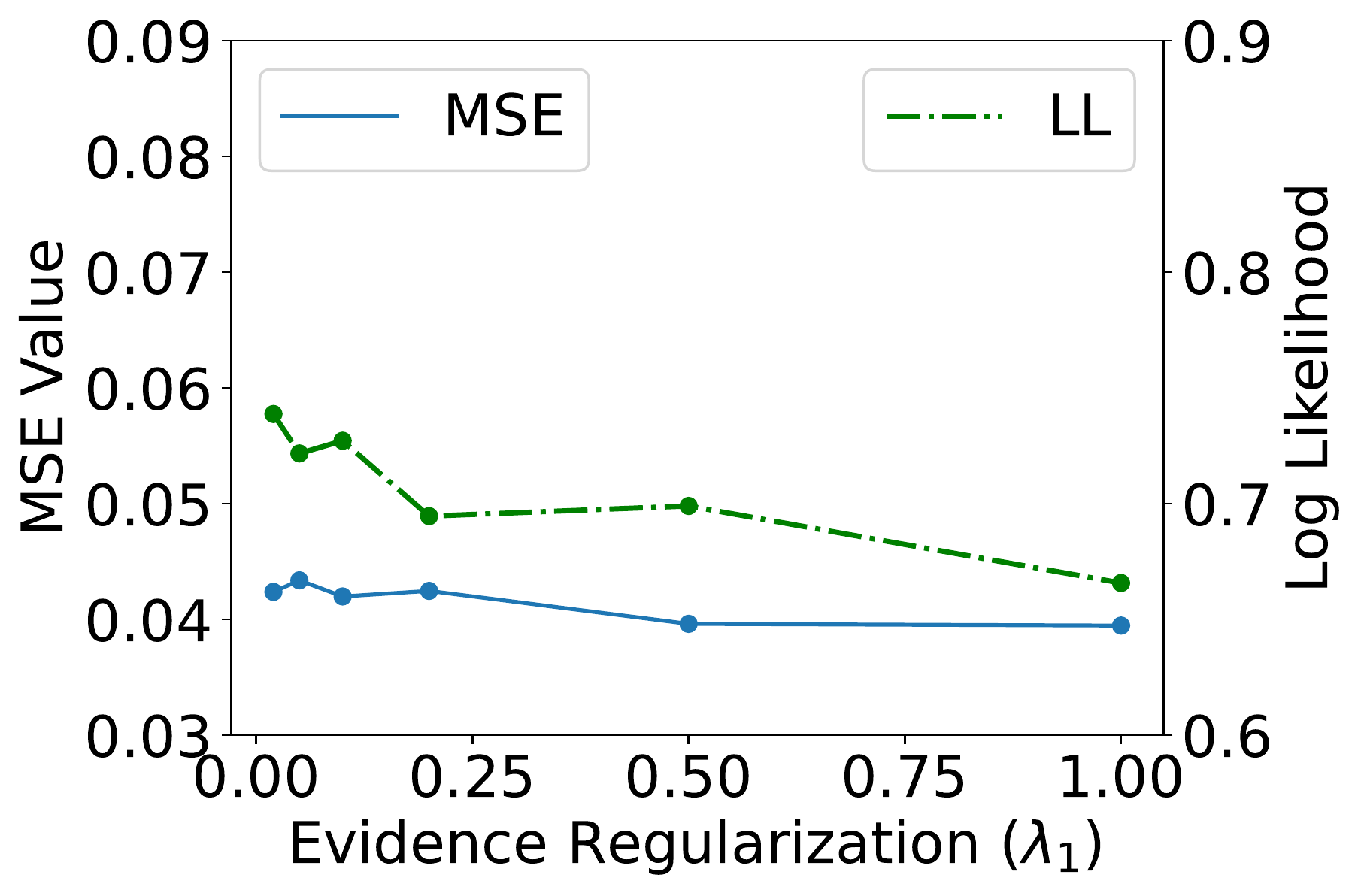}
  \caption{Impact to MSE and LL}
\end{subfigure}
\caption{Impact of Kernel Based Regularization ($\lambda_2$) to different uncertainty metrics}
\label{fig:ImpactKerRegUncMetrics}
\end{figure*}

We also study the impact of regularization terms to uncertainty using MNIST dataset over 50-shot Image Completion experiments. Figure \ref{fig:ImpactRegUncMetrics}
shows the effect of different evidence regularization values $\lambda_1$ when $\lambda_2 = 0.1$. As can be seen, larger regularization leads to improved Inclusion performance without any impact the Uncertainty Increase metrics. Figure \ref{fig:ImpactKerRegUncMetrics} shows the effect of kernel based regularization when evidence regularization term $\lambda_1 = 0.1$. Reasonable value of kernel based regularization helps improve the Uncertainty Increase metric without any impact to the Inclusion. Finally, very large values of the uncertainty regularization terms (both evidence regularization and kernel based regularizations) hurt the model's generalization capabilities as shown in Figure \ref{fig:ImpactRegUncMetrics} (c) and Figure \ref{fig:ImpactKerRegUncMetrics} (c).  

\section{Additional Experimental Results}\label{app:exp}
We present additional qualitative results with GP samples and CelebA that visualize the estimated uncertainty. We also conduct a deeper analysis on the behavior of the evidential parameters that reveal important insights on how the proposed evidential neural processes effectively combine the learning from few-shot samples (\ie context data) and the meta-knowledge from other tasks to achieve accurate prediction performance and fine-grained uncertainty quantification. 

\begin{figure}[h] 
\vspace{-1mm}
\centering
\begin{subfigure}{0.23\textwidth}
  \centering
  \includegraphics[width=0.9\linewidth]{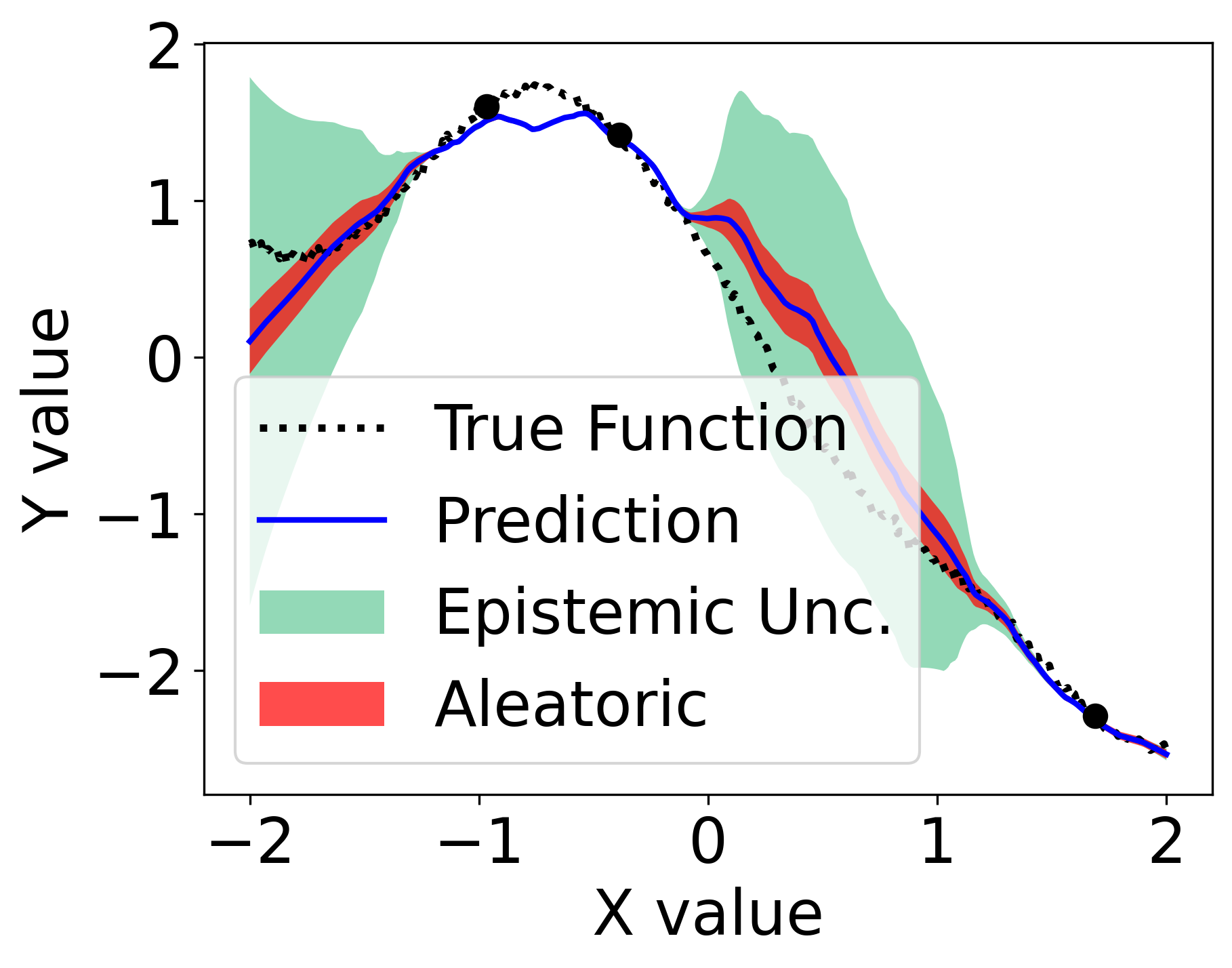}
  \caption{3 Context Points}
  \end{subfigure}
 \begin{subfigure}{0.23\textwidth}
  \centering
  \includegraphics[width=0.9\linewidth]{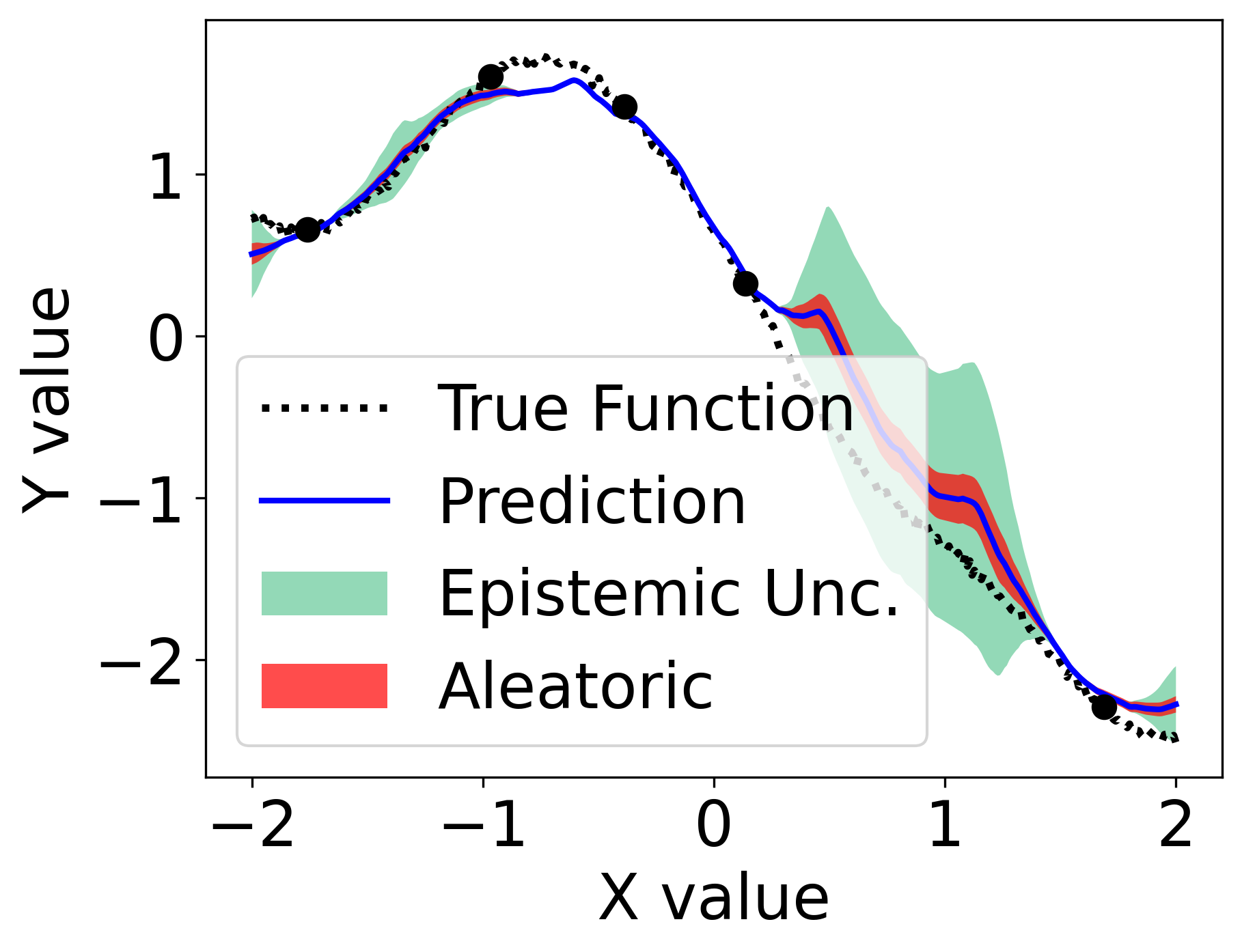}
  \caption{5 Context Points}
\end{subfigure}
\begin{subfigure}{0.23\textwidth}
  \centering
  \includegraphics[width=0.9\linewidth]{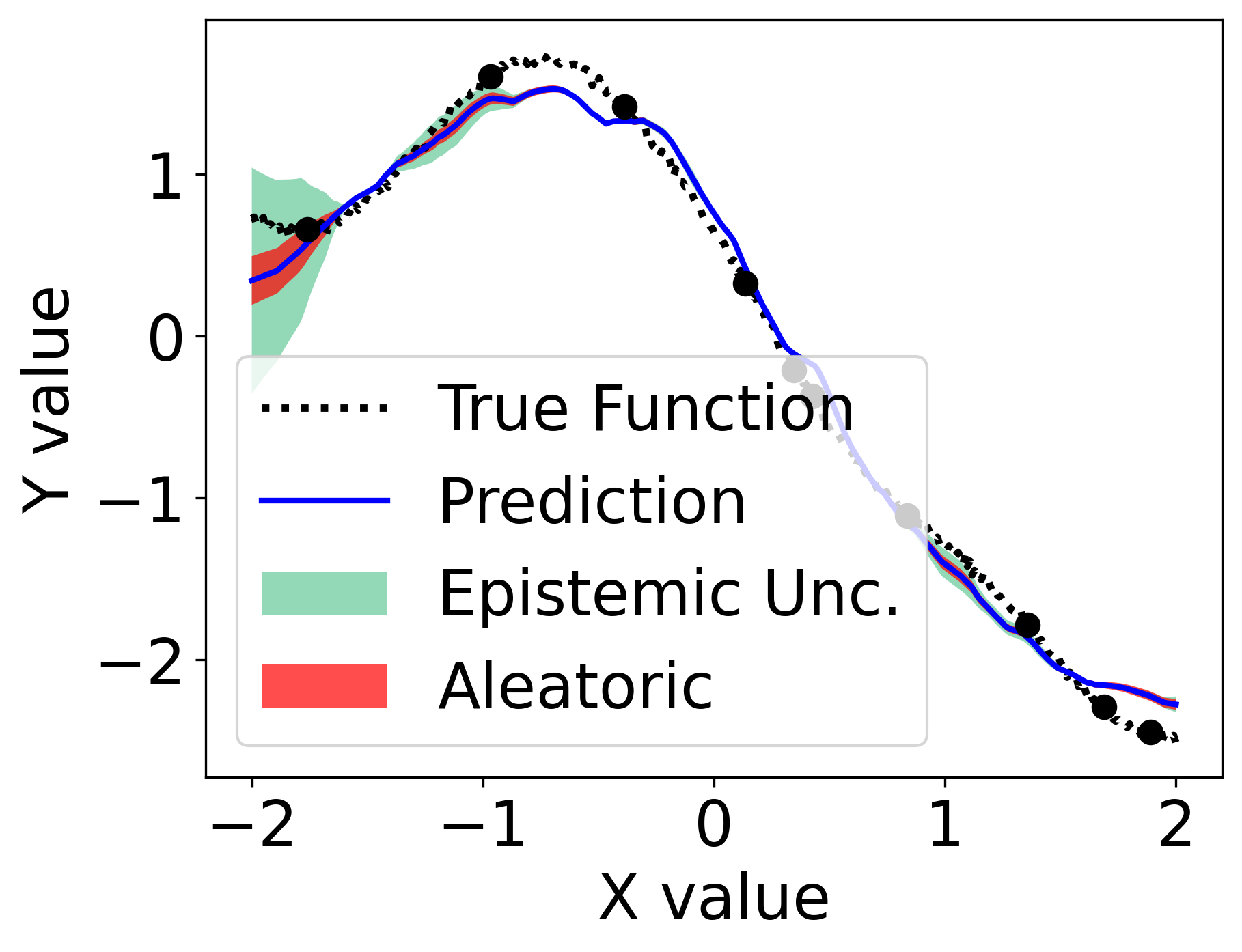}
  \caption{10 Context Points}
\end{subfigure}
\begin{subfigure}{0.23\textwidth}
  \centering
  \includegraphics[width=0.9\linewidth]{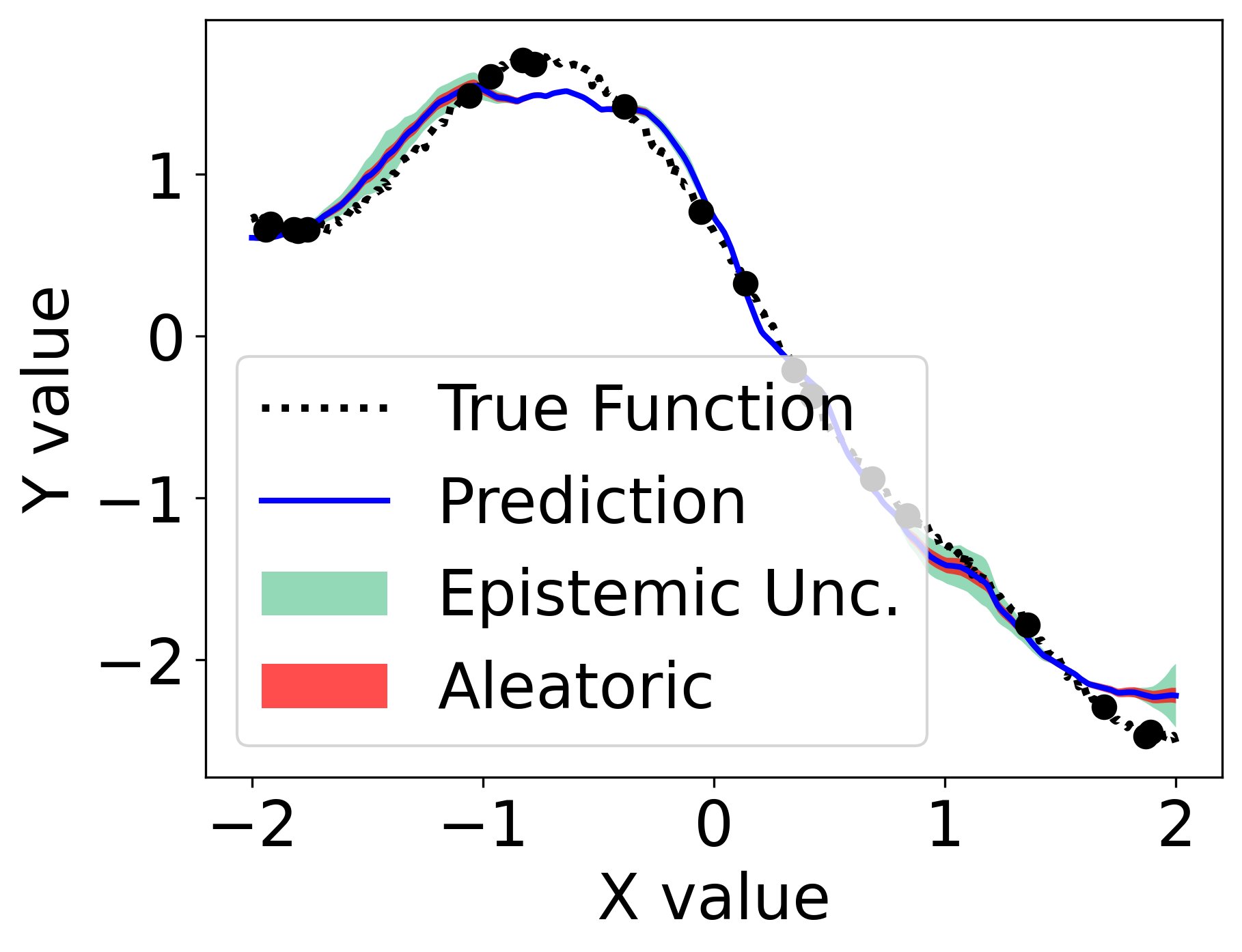}
  \caption{20 Context Points}
\end{subfigure}
\caption{ECNP model performance on a GP function regression task }
\label{fig:gpfunctionREgression}
\vspace{-1mm}
\end{figure}

\paragraph{Regression experiments on GP tasks.}
We trained the ECNP model for 20,000 iterations on 10-shot GP tasks and evaluated the model performance on a random test task. Figure \ref{fig:gpfunctionREgression} shows the model's behavior on random GP tasks for ECNP as we increase the number of data points on the context set. The model outputs a high uncertainty at regions where the model has not observed the context point/s and it believes that the meta knowledge is not sufficient for accurate prediction. A low uncertainty will be predicted otherwise. For instance, in Figure \ref{fig:gpfunctionREgression} (b), the model outputs confident correct predictions in the region between the 3rd and 4th context points even though they are relatively far away from both context points. This may be due to that the meta-knowledge learned from other tasks is rich enough to lower the uncertainty. As we increase the number of observations in the context set, the model's confidence increases along with the predictive accuracy as indicated by Figures \ref{fig:gpfunctionREgression} (c) and \ref{fig:gpfunctionREgression} (d).

\begin{figure}[h] 
\centering
\begin{subfigure}{0.46\textwidth}
  \centering
  \includegraphics[width=0.9\linewidth]{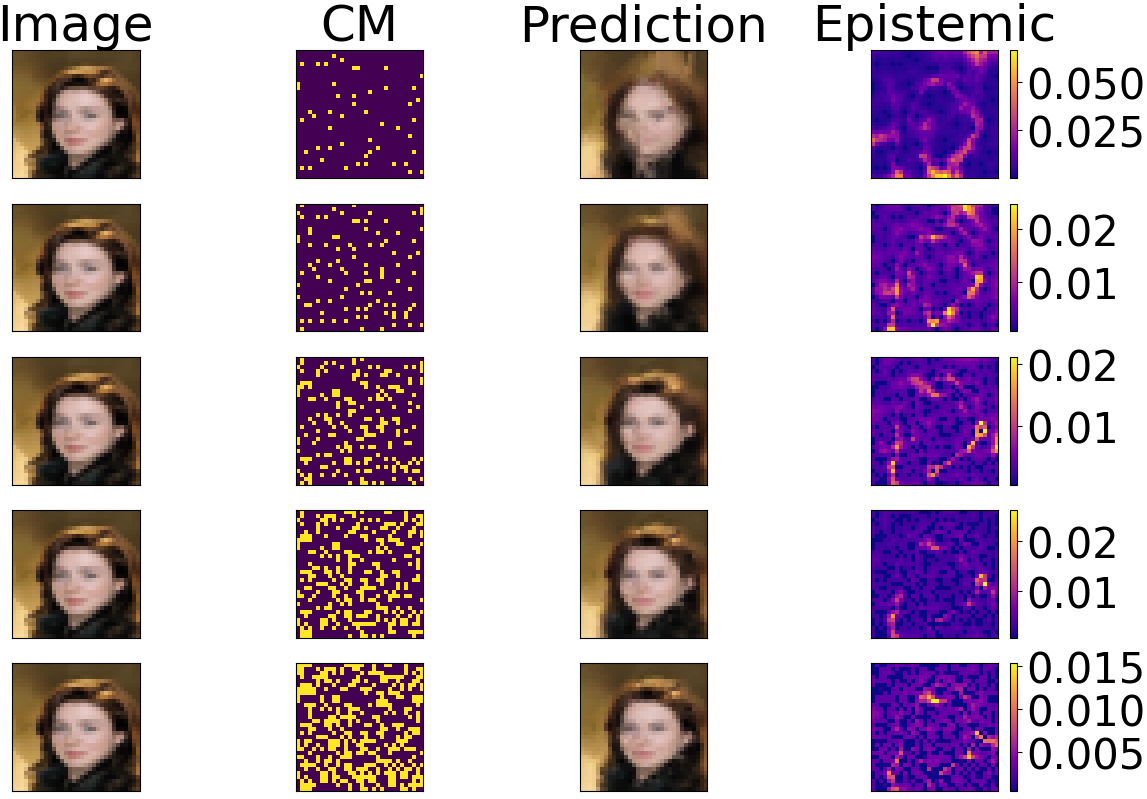}
  \caption{Epistemic Uncertainty Visualization}
  \end{subfigure}
 \begin{subfigure}{0.22\textwidth}
  \centering
  \includegraphics[width=0.9\linewidth]{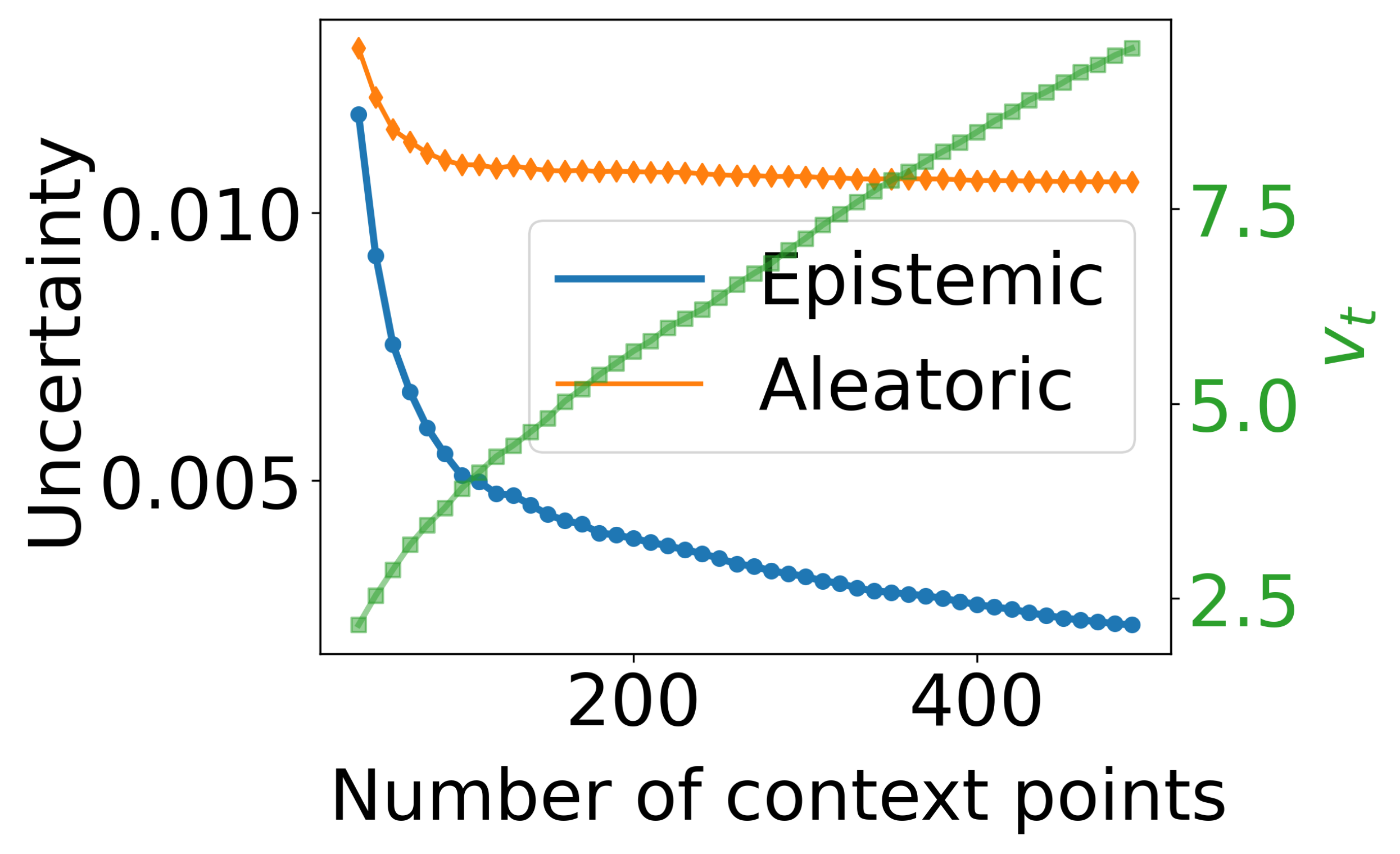}
  \caption{Uncertainty Trends}
\end{subfigure}
\begin{subfigure}{0.22\textwidth}
  \centering
  \includegraphics[width=0.9\linewidth]{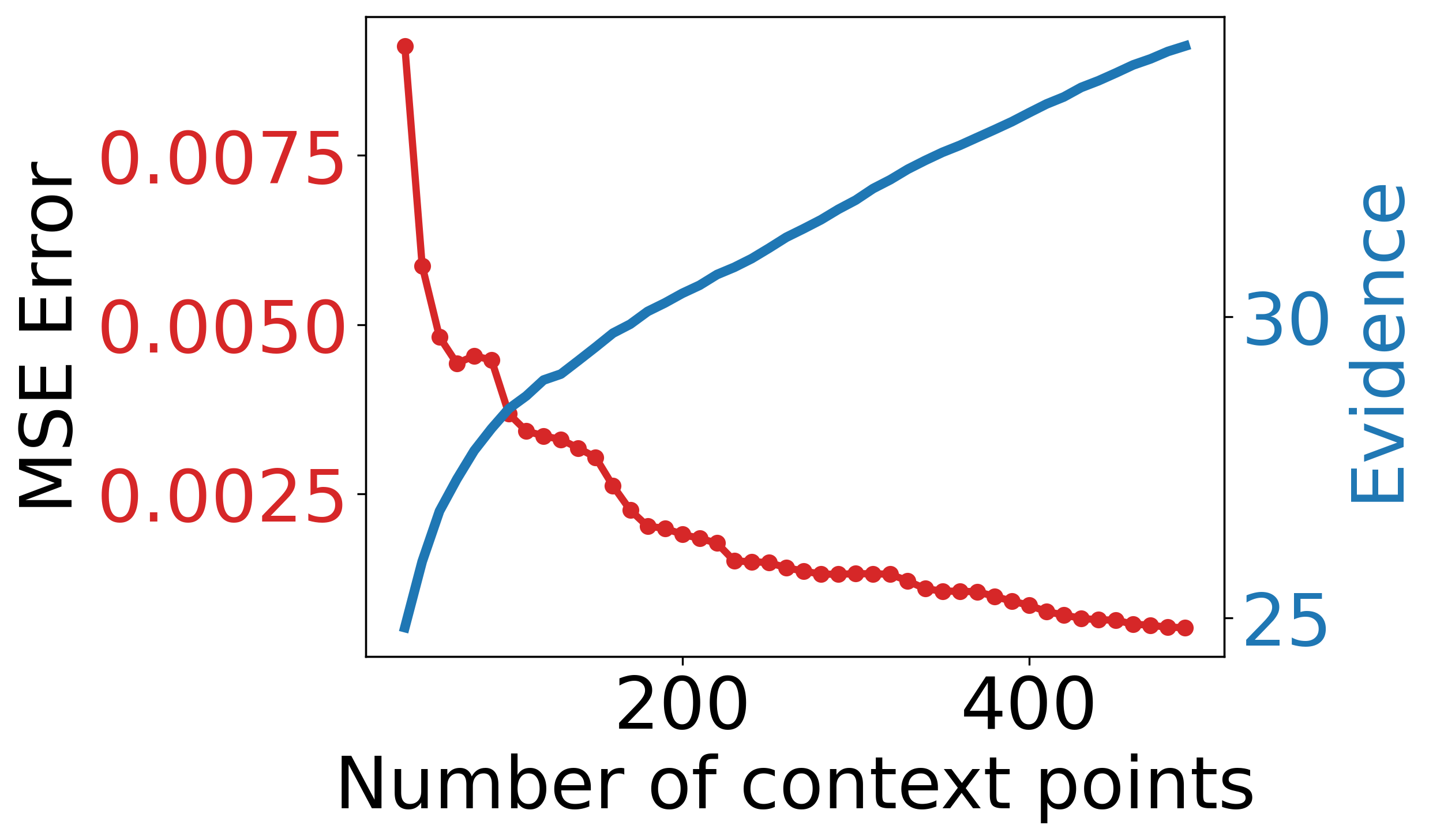}
  \caption{MSE/Evidence Trend}
\end{subfigure}
\caption{Evidential ANP model performance on a CelebA task}
\label{fig:anpcelebaqual}
\end{figure}

\paragraph{Image completion experiments on CelebA.} Figure \ref{fig:anpcelebaqual} shows the qualitative results of the evidential attentive neural process model on a random CelebA test task. The model was trained for 50 epochs using 200-shot CelebA tasks with $\lambda_1 = 0.1$ and $\lambda_2 = 1.0$.  As we increase the number of context points in the task (indicated by Context Mask CM), the average epistemic uncertainty decreases rapidly whereas the aleatoric uncertainty decreases at a slower pace. This could be because there may be inherent noises associated with the few-shot tasks that may not be addressed by newly added context points. Furthermore, from (16) and (17), the epistemic and aleatoric uncertainties vary by a factor of $v_t$, which captures the meta-knowledge according to our previous discussions.  More context points could allow the model to better relate to other meta-training tasks, leading to a larger $v_t$ as shown by \ref{fig:anpcelebaqual} (b). This also contributes to a faster decrease of the epistemic uncertainty. Moreover, with additional data points on the context set, the model has greater evidence for its prediction and the model's predictive accuracy increases as indicated by decrease in the MSE error.

\begin{figure*}[t!] 
\centering
\begin{subfigure}{0.46\textwidth}
  \centering
  \includegraphics[width=0.9\linewidth]{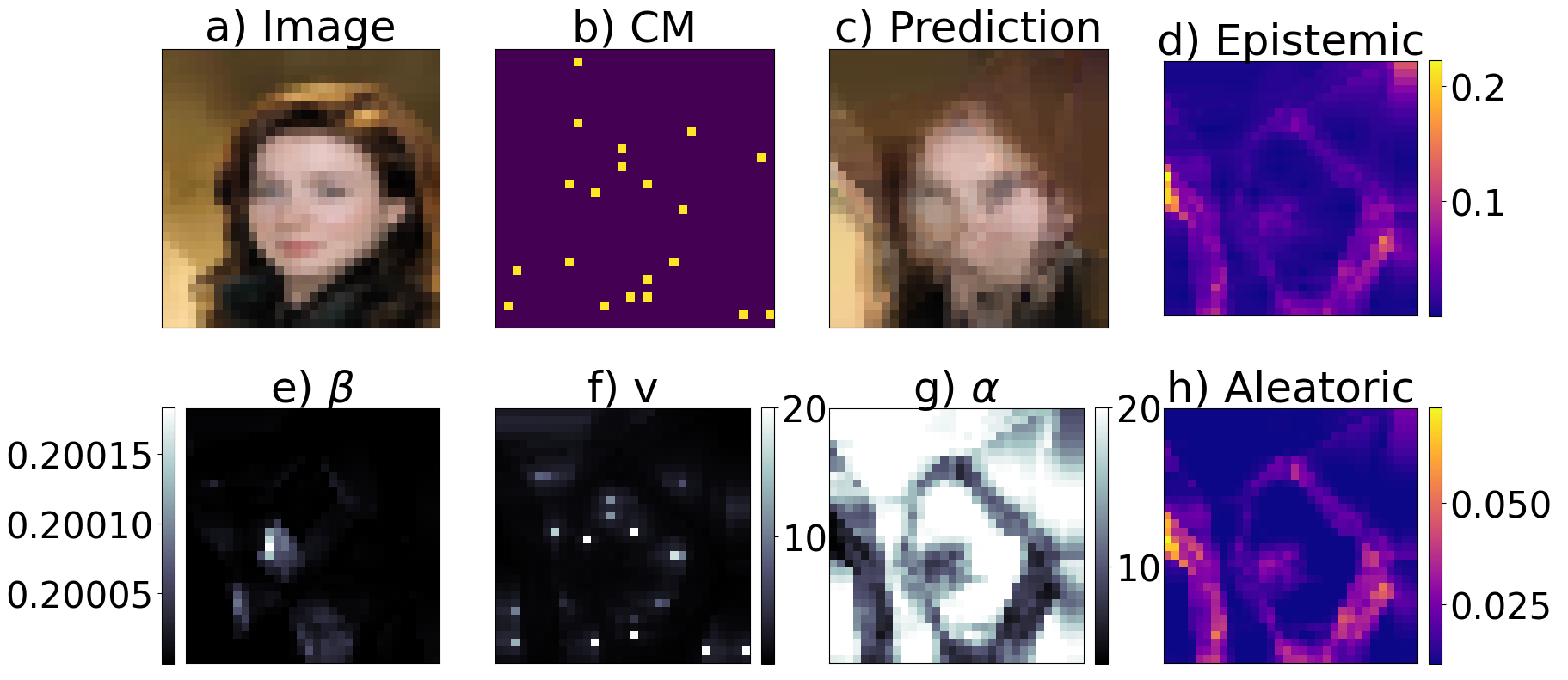}
  \caption{20 Context Points}
  \end{subfigure}
 \begin{subfigure}{0.46\textwidth}
  \centering
  \includegraphics[width=0.9\linewidth]{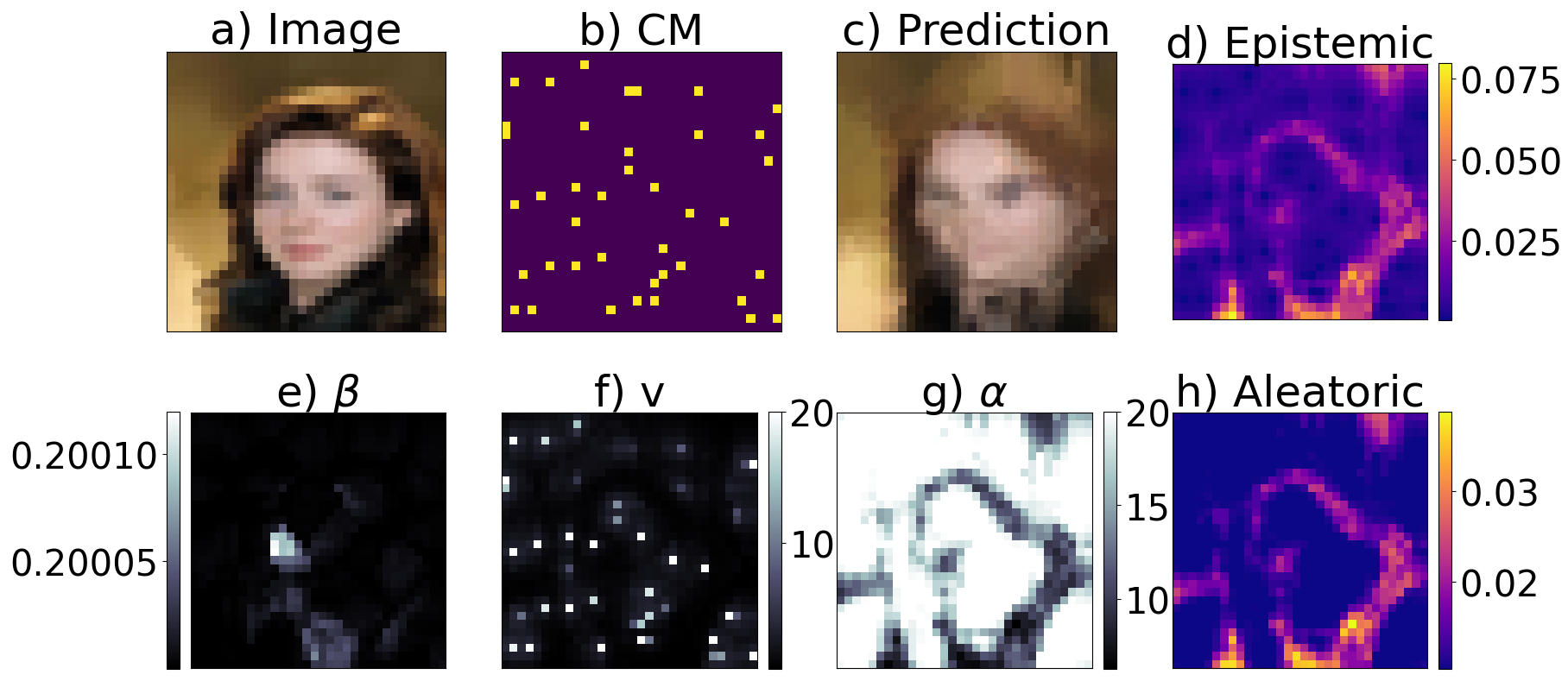}
  \caption{50 Context Points}
\end{subfigure}
\begin{subfigure}{0.46\textwidth}
  \centering
  \includegraphics[width=0.9\linewidth]{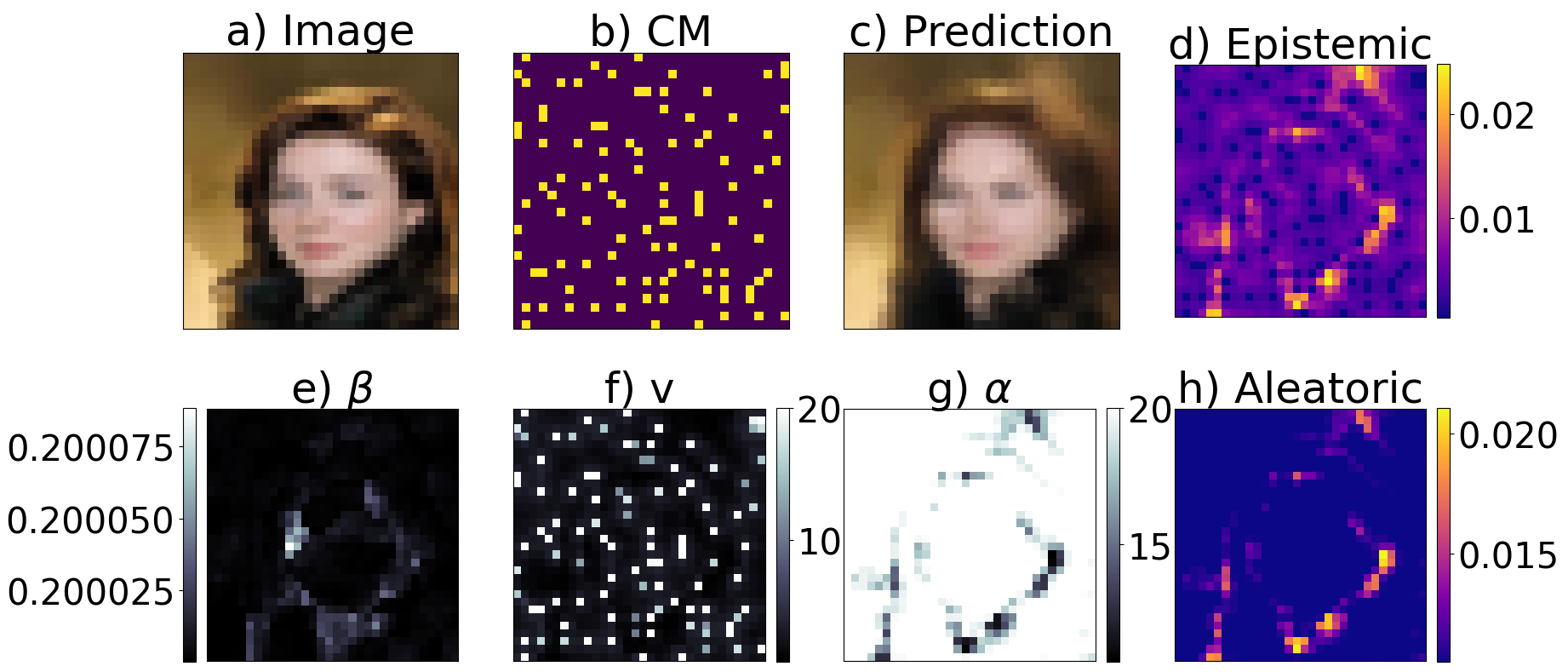}
  \caption{100 Context Points}
\end{subfigure}
\begin{subfigure}{0.46\textwidth}
  \centering
  \includegraphics[width=0.9\linewidth]{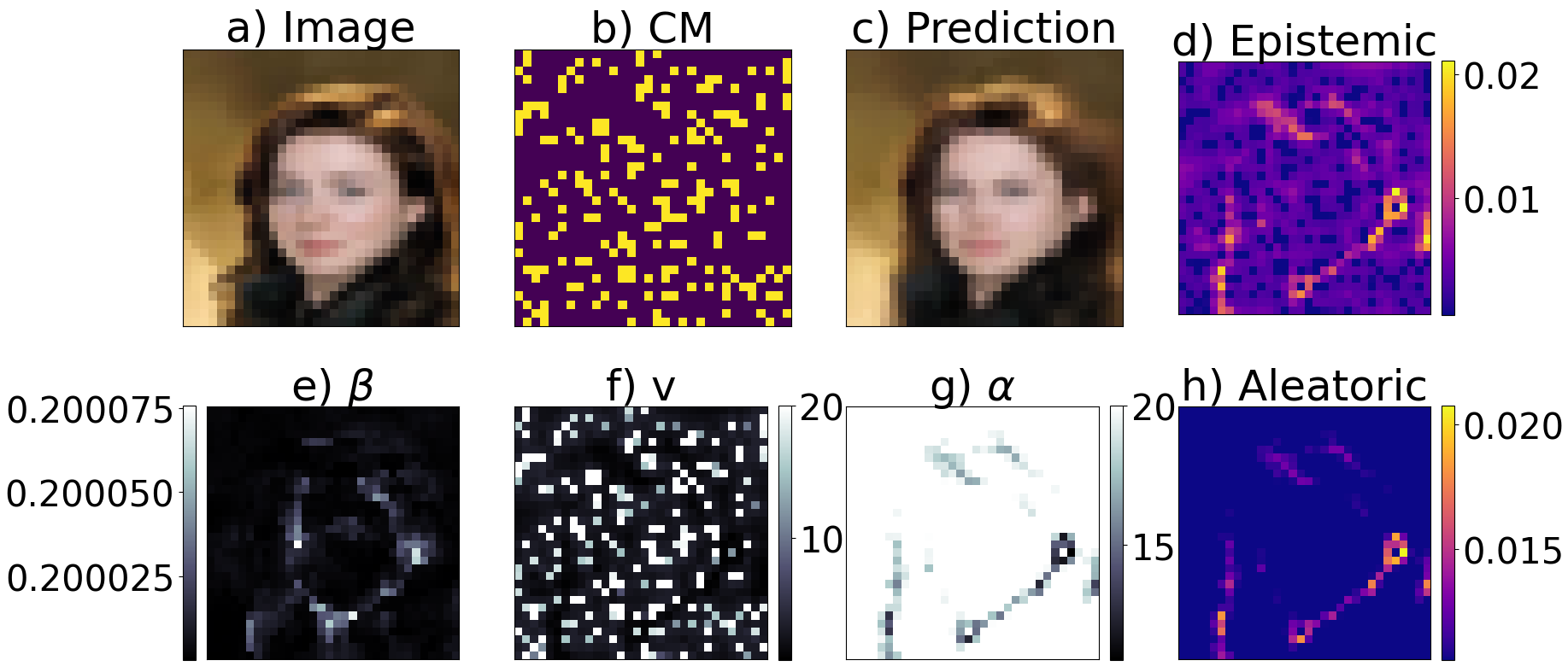}
  \caption{200 Context Points}
\end{subfigure}
\caption{Visualization of the evidential hyperparameters and predicted uncertainty on a CelebA task}
\label{fig:anpevidhyper}
\end{figure*}

Figure \ref{fig:anpevidhyper} visualizes the three higher-order hyperparameters $\alpha$, $\beta$, and $v$ along with the epistemic and aleatoric uncertainties for a CelebA test task with varying number of context points. We average the evidential hyperparameters and uncertainty across the 3 channels for illustration. When the number of context points in the task is low, the epistemic uncertainty and aleatoric uncertainty show similar trends. As we observe more data, the model evidence (\ie $\alpha$, $\beta$, and $v$) gets updated to reflect the model's knowledge. When there are a few context points in the task, these hyperparameter values are guided by the meta-knowledge. As the model observes more context points, the model integrates the meta-knowledge with the information to update the evidential parameters. Specifically, the model considers the evidential hyperparameters $\alpha$ and $\beta$ to estimate the aleatoric uncertainty that seems to be high around the edges and the boundaries. The model considers the evidential hyperparameter $v$ along with $\alpha$ and $\beta$ to estimate the epistemic uncertainty that decreases gradually with more data in the context set. Similar trends are observed across the experiments.

\begin{figure*}[h] 
\centering
\begin{subfigure}{0.22\textwidth}
  \centering
  \includegraphics[width=0.9\linewidth]{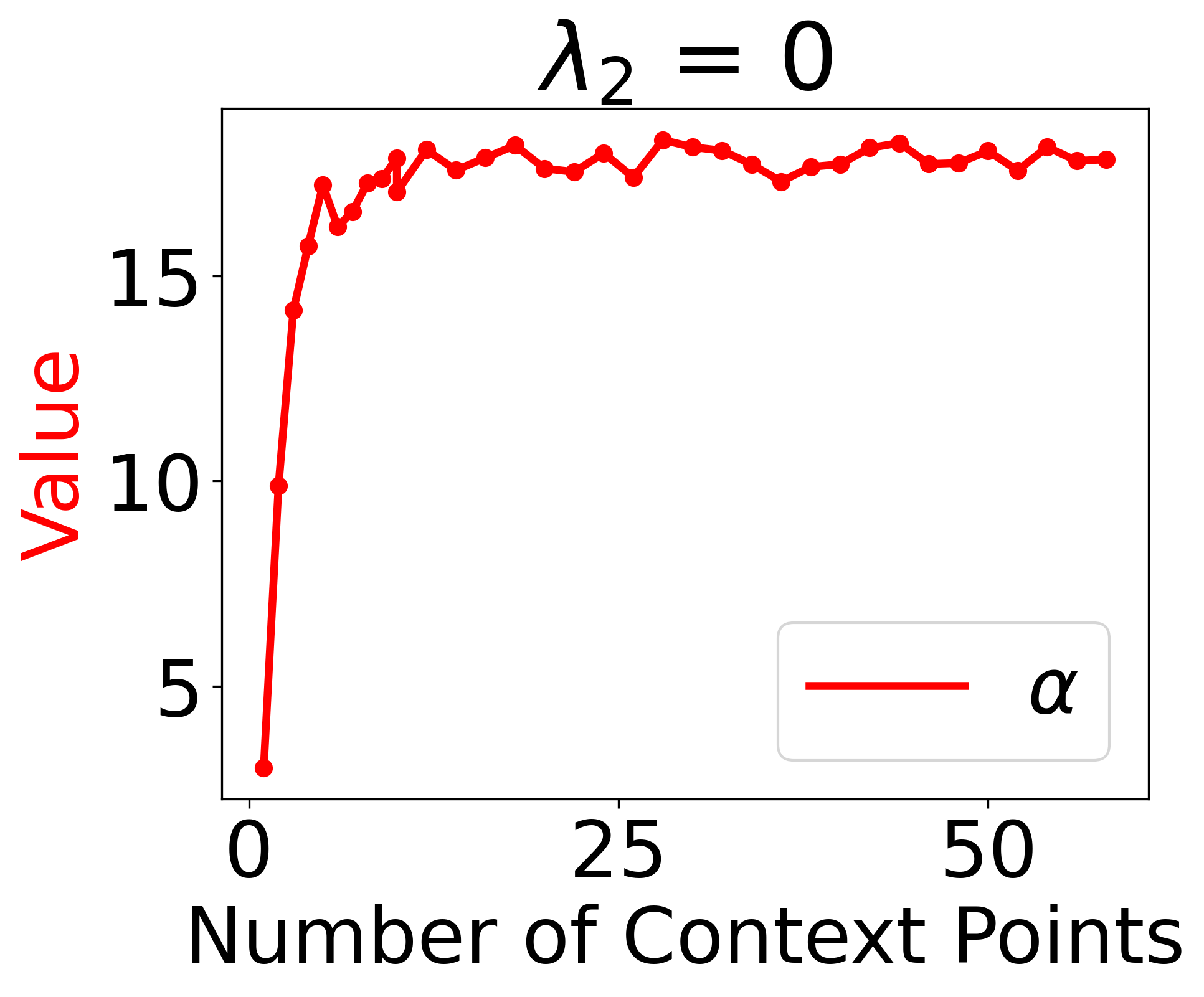}
  \caption{$\alpha$ trend}
  \end{subfigure}
 \begin{subfigure}{0.22\textwidth}
  \centering
  \includegraphics[width=0.9\linewidth]{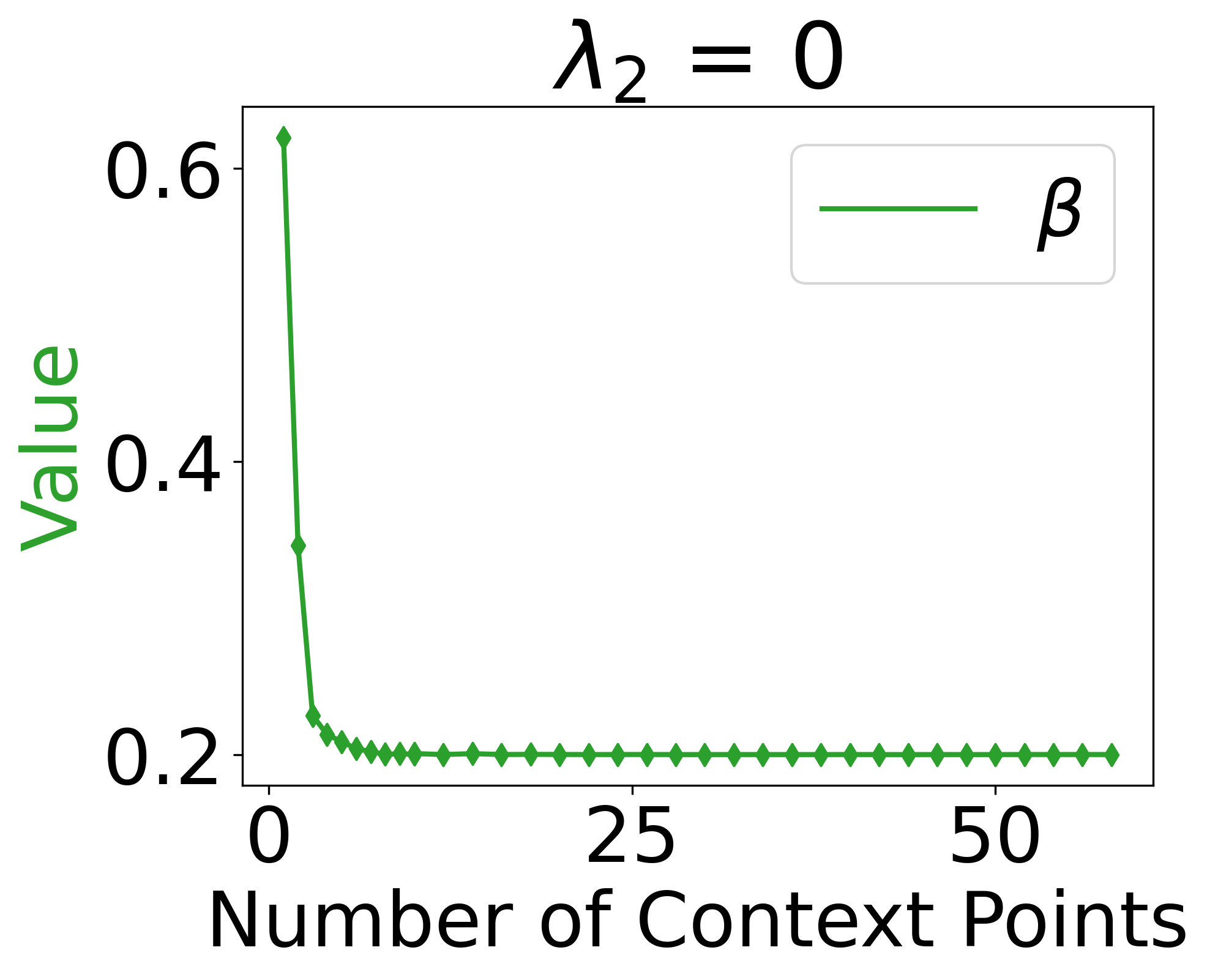}
  \caption{$\beta$ trend}
\end{subfigure}
\begin{subfigure}{0.22\textwidth}
  \centering
  \includegraphics[width=0.9\linewidth]{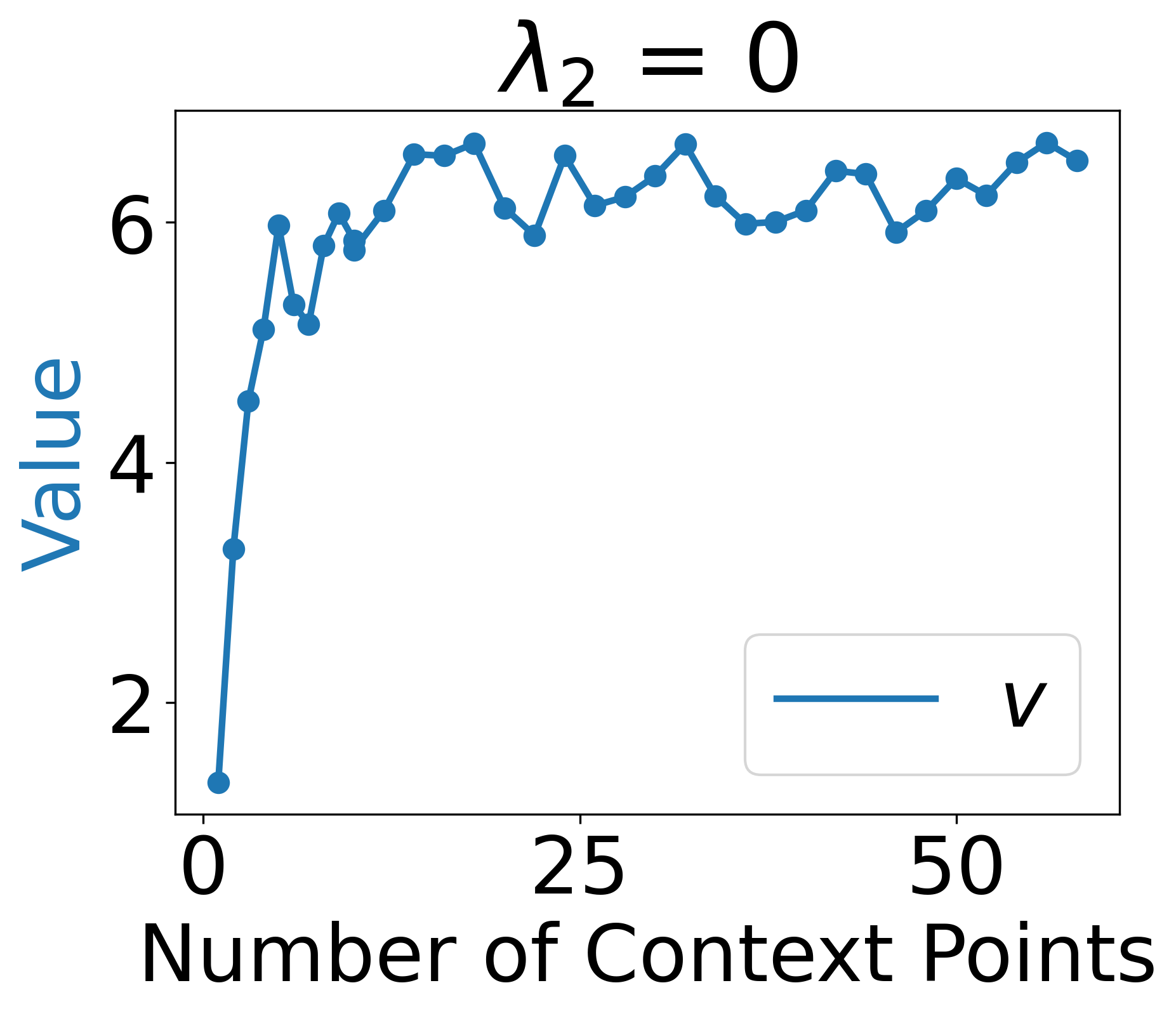}
  \caption{$v$ trend}
\end{subfigure}
\begin{subfigure}{0.22\textwidth}
  \centering
  \includegraphics[width=0.9\linewidth]{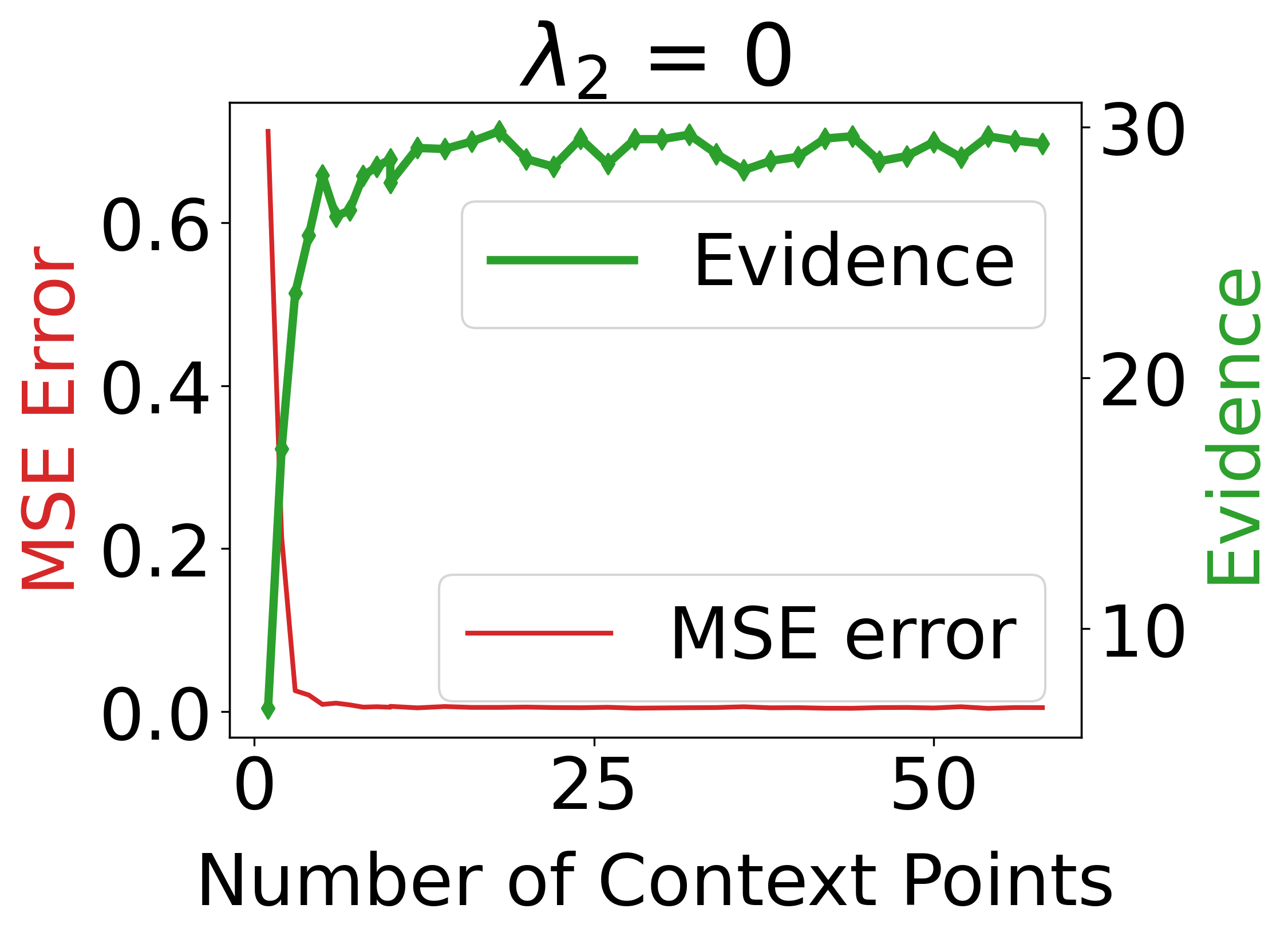}
  \caption{MSE/Evidence trend}
\end{subfigure}
\caption{Evidential CNP model performance on a Sinusoid Regression task for $\lambda_2 = 0$}
\label{fig:cnpSinEvHyp}
\end{figure*}

\begin{figure*}[h] 
\centering
\begin{subfigure}{0.22\textwidth}
  \centering
  \includegraphics[width=0.9\linewidth]{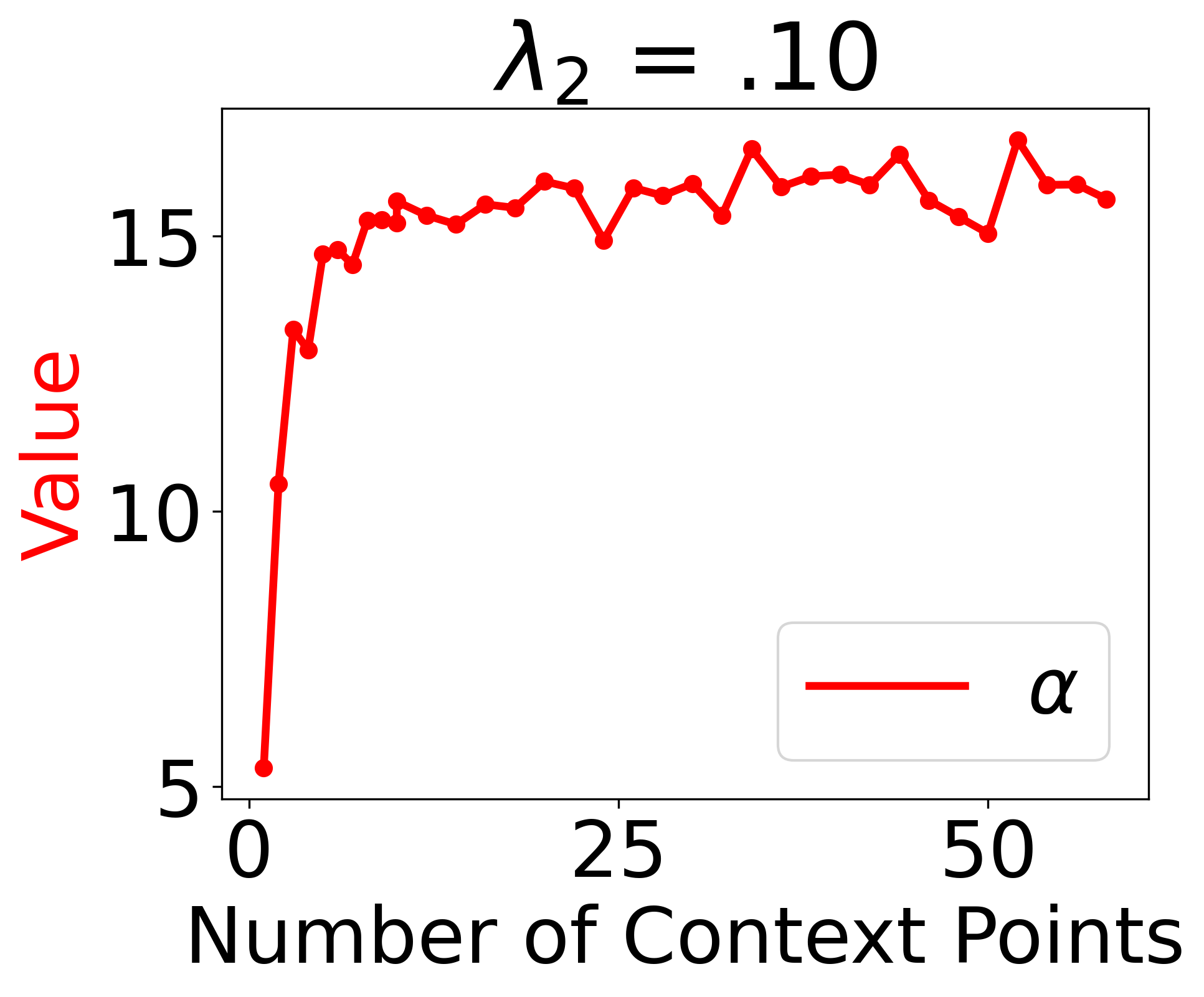}
  \caption{$\alpha$ trend}
  \end{subfigure}
 \begin{subfigure}{0.22\textwidth}
  \centering
  \includegraphics[width=0.9\linewidth]{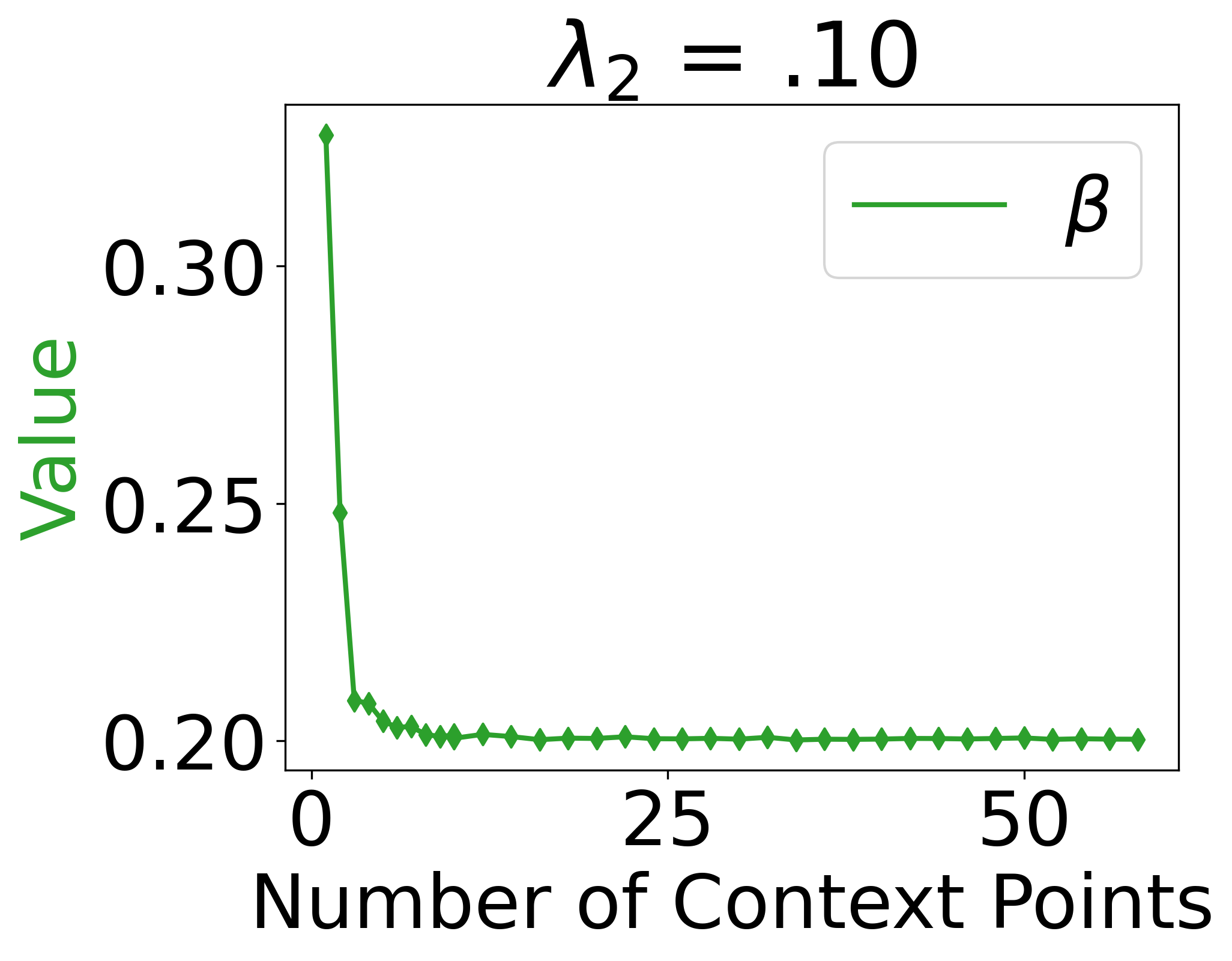}
  \caption{$\beta$ trend}
\end{subfigure}
\begin{subfigure}{0.22\textwidth}
  \centering
  \includegraphics[width=0.9\linewidth]{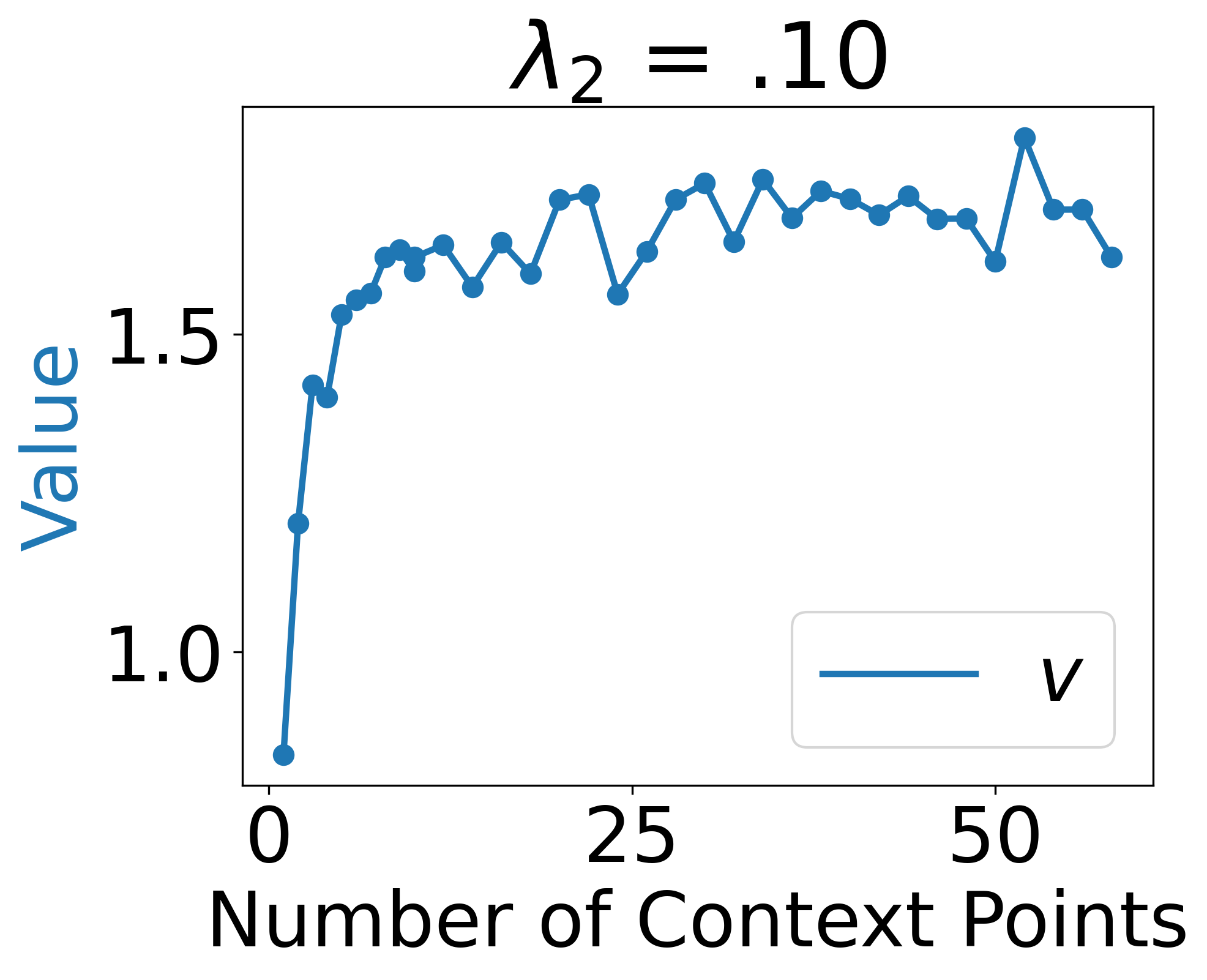}
  \caption{$v$ trend}
\end{subfigure}
\begin{subfigure}{0.22\textwidth}
  \centering
  \includegraphics[width=0.9\linewidth]{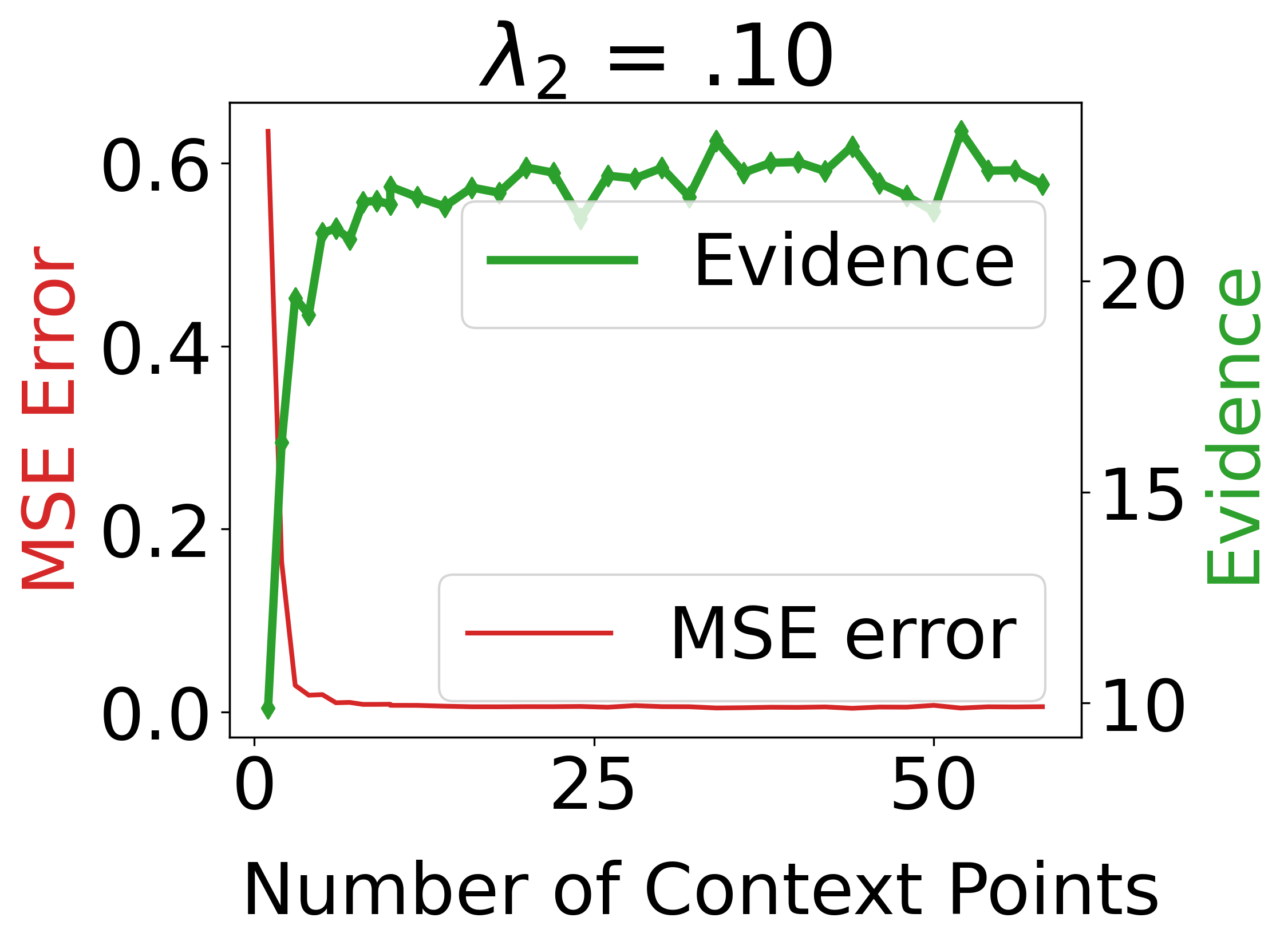}
  \caption{MSE/Evidence trend}
\end{subfigure}
\caption{Evidential CNP model performance on a Sinusoid Regression task for $\lambda_2 = 0.1$}
\label{fig:cnpSinEvHyp2}
\end{figure*}

\paragraph{Evidential parameters guided local vs. meta-learning.} We further inspect the evidential hyperparameters to better understand the model behavior. We first experiment with relatively simple function regression tasks over the sinusoid dataset. We train the evidential conditional neural process model with $\lambda_1 = 0/0.1$ and $\lambda_2 = 1.0$ for 30,000 meta iterations and evaluate on a random test task. The test task consists of 400 target points from which we randomly select $K$-shots in the context set. We track the average MSE error and the evidential hyperparameters  $\alpha$, $\beta$, and $v$ on the target set across all the test tasks. Figures \ref{fig:cnpSinEvHyp} and \ref{fig:cnpSinEvHyp2} show the results of the experiments where we observe an interesting trend. As can be seen from Figure \ref{fig:cnpSinEvHyp} (d) and Figure \ref{fig:cnpSinEvHyp2} (d), the model performance converges after a few context points (\ie $<10$) along with the (average) predicted evidence score. Meanwhile, both  $\alpha$ and $v$ predicted on the testing points also stop increasing. This implies that adding additional context points no longer helps to improve learning from local data (\ie context points), which is captured by $\alpha$ based on the hierarchical structure defined in (1)-(3). Similarly, it does not help to learn from the meta-knowledge, either, which is captured by $v$. This example clearly shows that how the proposed model effectively combines the learning from the local context data points while leveraging meta-knowledge from other similar few-shot task to achieve a fast convergence for relatively simple tasks. 

\begin{figure*}[h] 
\centering
\begin{subfigure}{0.21\textwidth}
  \centering
  \includegraphics[width=0.9\linewidth]{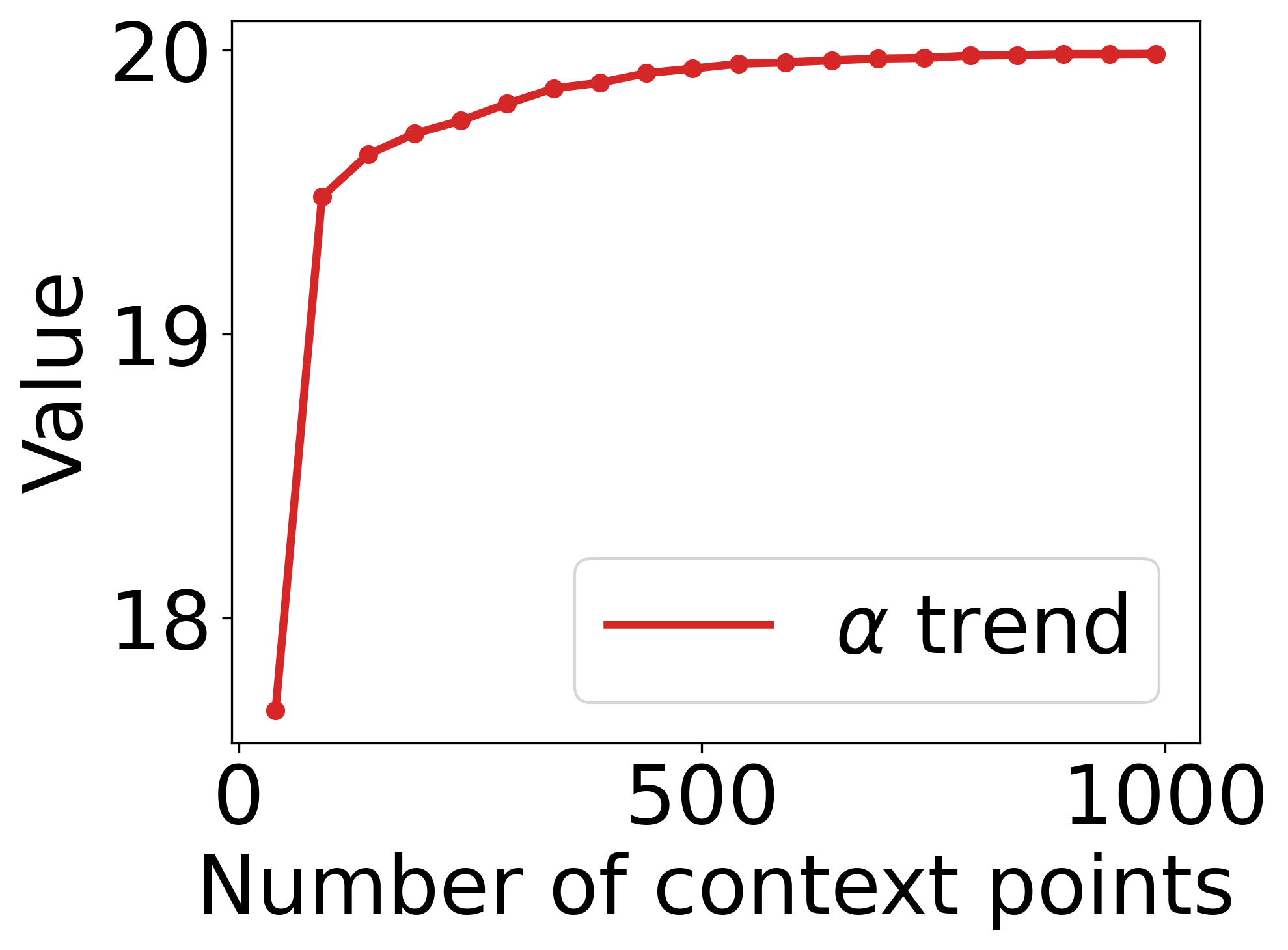}
  \caption{$\alpha$ trend}
  \end{subfigure}
\begin{subfigure}{0.25\textwidth}
  \centering
  \includegraphics[width=0.9\linewidth]{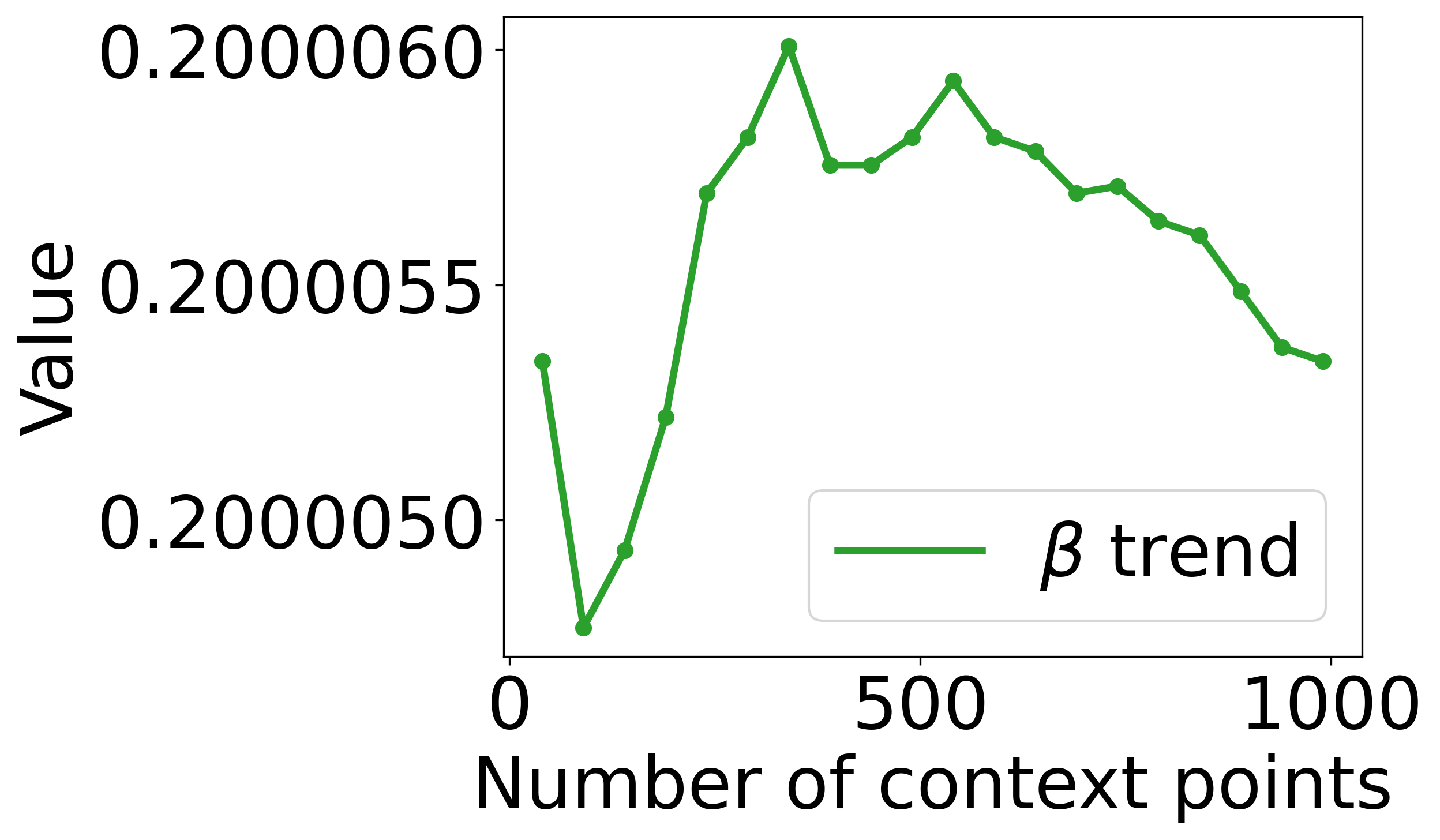}
  \caption{$\beta$ trend}
  \end{subfigure}
\begin{subfigure}{0.21\textwidth}
  \centering
  \includegraphics[width=0.9\linewidth]{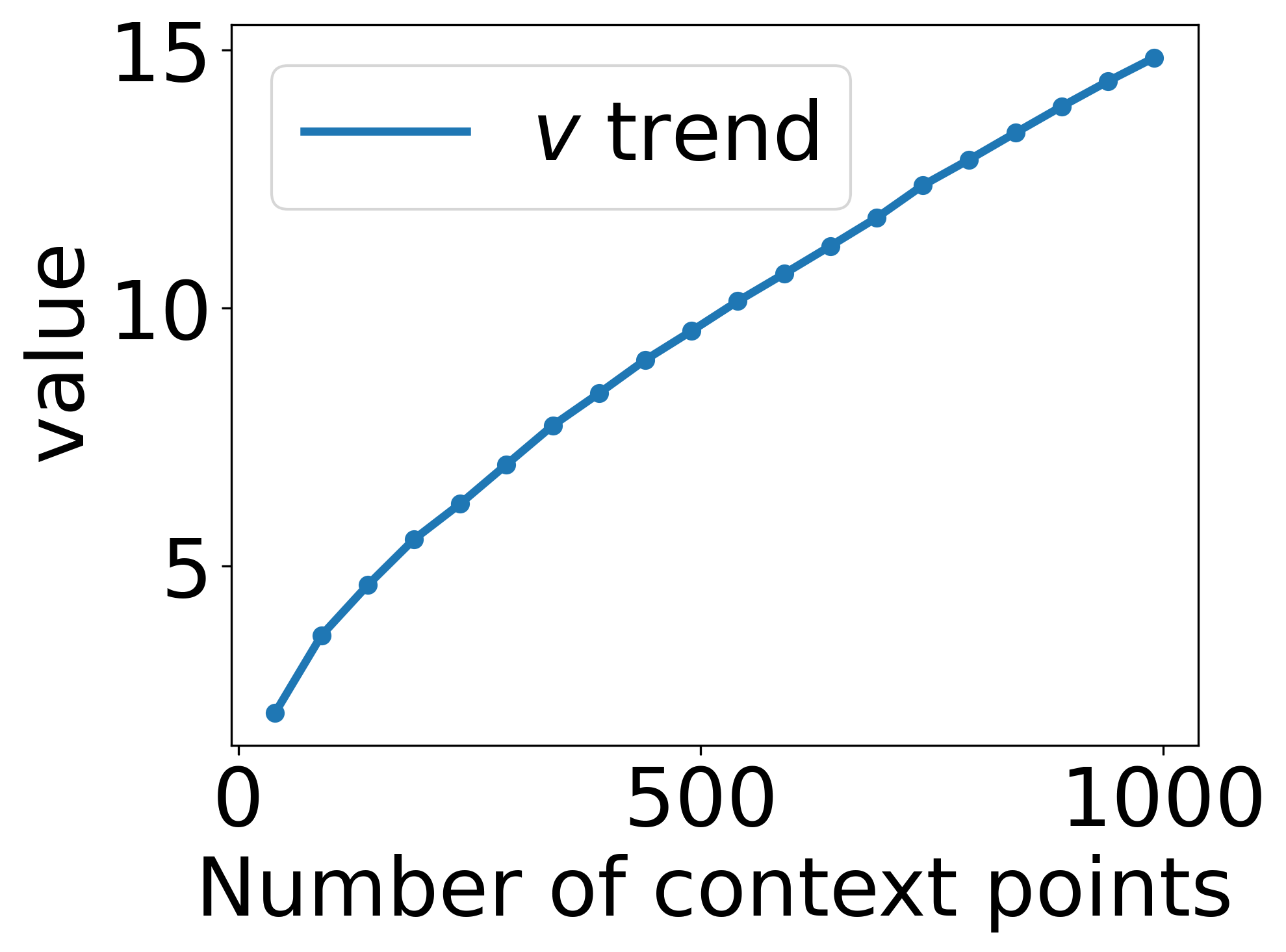}
  \caption{$v$ trend}
\end{subfigure}
\begin{subfigure}{0.22\textwidth}
  \centering
  \includegraphics[width=0.9\linewidth]{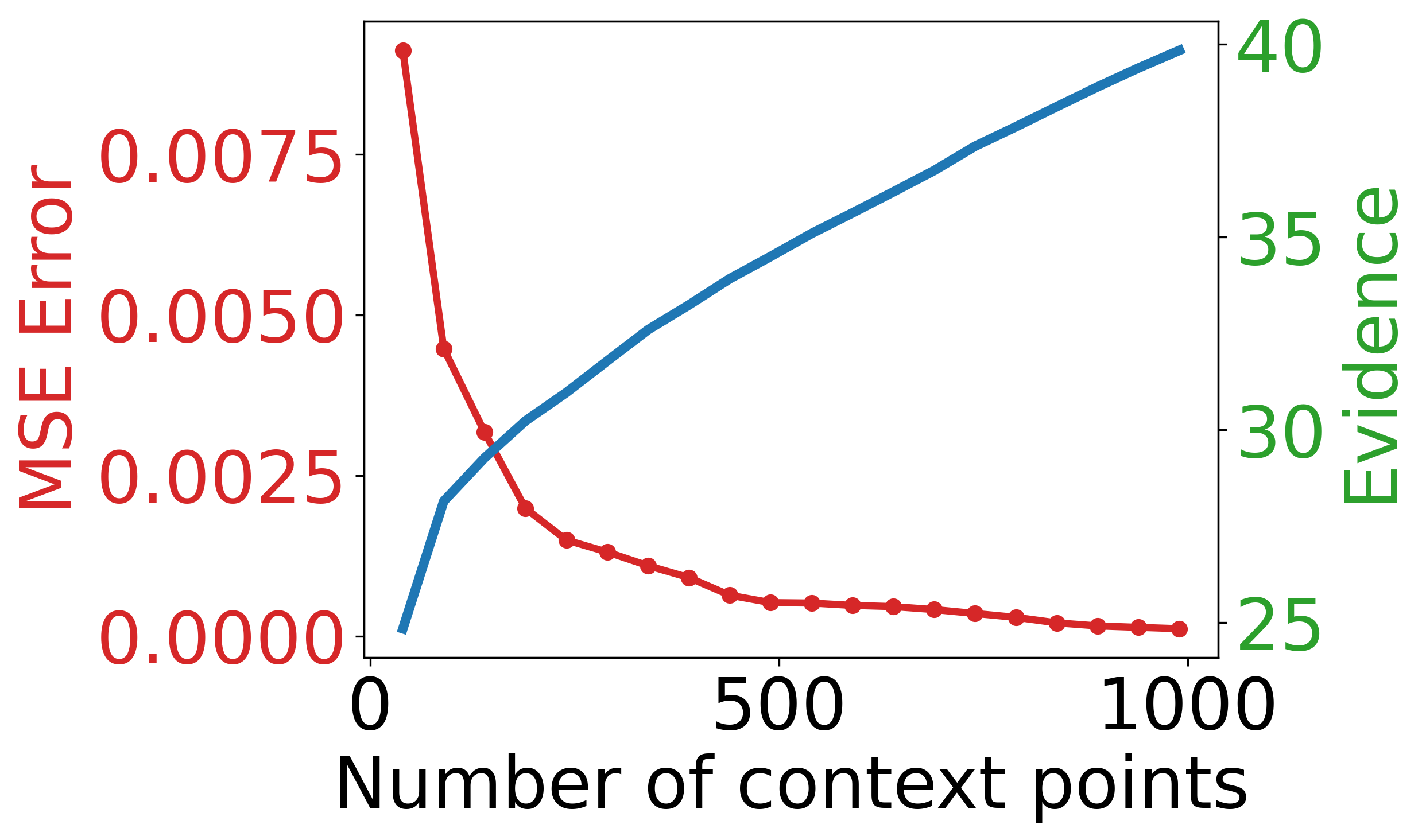}
  \caption{MSE/Evidence trend}
\end{subfigure}
\caption{Evidential ANP model performance on a CelebA task for $\lambda_2 = 1.0$}
\label{fig:ANPEvidHyp}
\end{figure*}

For a more challenging CelebA task as shown in Figure \ref{fig:ANPEvidHyp}, we observe a similar trend for $\alpha$, which increases along with the addition of context points but starts to converge after a number of context points have been included. However, $v$ shows a very different trend that continues to increase along with the addition of context points. The different behavior in these two evidential parameters precisely captures how the proposed model conducts effective learning for more challenging tasks. In particular, for such tasks, the local data (\ie context points) contribute relatively less since they are inherently limited in the few-shot setting for more complex tasks. Meanwhile, meta-knowledge is expected to play a more important role given a potential large number of training tasks available for the model to learn the meta-knowledge (and using the context points related to the learned meta-knowledge). This exactly matches the faster convergence of $\alpha$ and continuous growth of $v$. Furthermore, we also observe a more random trend in $\beta$ due to a higher noise ratio in the image completion tasks. Finally, the MSE/Evidence trend matches the changes on $\alpha$ and $v$: evidence continues to grow due to the contribution from $v$ and MSE converges at a slower pace than the simpler tasks and its decrease at the later stages of the leaning is mainly attribute to the increase of $v$ (\ie the meta-knowledge).

\newpage
\section{Limitations and Future Work}\label{app:limitations} In this work, we focus on the CNP family because of their rapid inference, scalability, and competitive predictive performance. More importantly, they naturally quantify uncertainty by simulating a stochastic process like a GP. We introduce a novel hierarchical Bayesian structure that can be viewed as a general augmentation to the CNP family of models to achieve fine-grained uncertainty decomposition and theoretically guaranteed robustness. Proposed novel structure can be combined with and enhance other models of the CNP family. To this end, we experiment with ConvCNP (Gordan et.al, 2020), a recent improvement of CNP, on 5-shot GP regression. The results are ConvCNP: MSE: 0.268, LL: -0.239, Evidential-ConvCNP: MSE: 0.228, LL:-0.012, which shows the potential of our method to augment recent CNP models. We leave additional exploration of the effectiveness of the proposed structure to other CNP works as a future work.

We developed evidential meta-learning model for fine-grained uncertainty quantification. The proposed ECNP model introduces two additional hyperparameters and requires hyperparameter tuning. Moreover, in this work, we experimented on few-shot regression tasks with 1D regression and 2D image completion. For the considered datasets, a relatively simple distance function such (the Euclidean distance) was effective for kernel based regularization. However, epistemic uncertainty guidance using kernel-based regularization in more challenging datasets such as image and videos may require better and efficient design of the distance function $D(.)$ (e.g. distance/similarity in the embedding space). Also, it can be an interesting future work to extend this work to other meta-learning approaches to equip them with fine-grained uncertainty quantification capabilities in a computationally efficient manner. We now plan to address the issues and experiment on larger datasets such as ImageNet with deeper neural networks as our future work. 

\end{document}